\documentclass{Dissertate}

\usepackage{graphicx}

\newcommand{\chapterseparatorpage}{%
  \clearpage
  \thispagestyle{empty}%
  \begin{center}
    \vspace*{2cm}
    \includegraphics[width=0.6\textwidth]{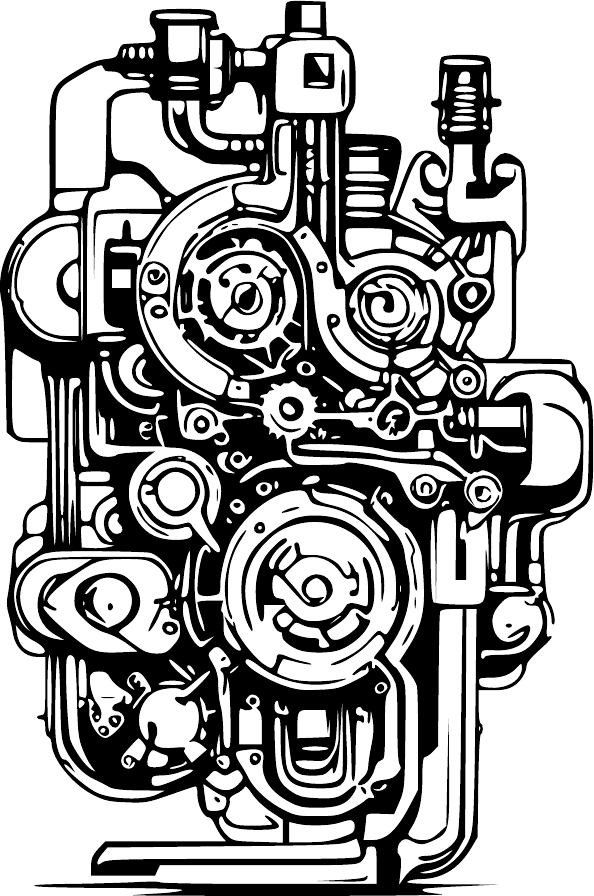}
    \vspace{2cm}
  \end{center}
  \clearpage
}

\renewcommand{\chapterimage}[1]{}

\begin{document}

\frontmatter
\glsdisablehyper
\newacronym{mt}{MT}{machine translation}
\newacronym{smt}{SMT}{statistical machine translation}
\newacronym{qe}{QE}{quality estimation}
\newacronym{nmt}{NMT}{neural machine translation}
\newacronym{llm}{LLM}{large language model}
\newacronym{plm}{PLM}{pre-trained language model}
\newacronym{icl}{ICL}{in-context learning}
\newacronym{ice}{ICE}{in-context example}
\newacronym{da}{DA}{domain adaptation}
\newacronym{dag}{DAG}{data augmentation}
\newacronym{sw}{SW}{subword}
\newacronym{zsl}{ZSL}{zero-shot learning}
\newacronym{zs}{ZS}{zero-shot}
\newacronym{rq}{RQ}{research question}
\newacronym{nlp}{NLP}{natural language processing}

\cleardoublepage

\includepdf[
  pages=1,
  pagecommand={\thispagestyle{empty}},
  fitpaper=true
]{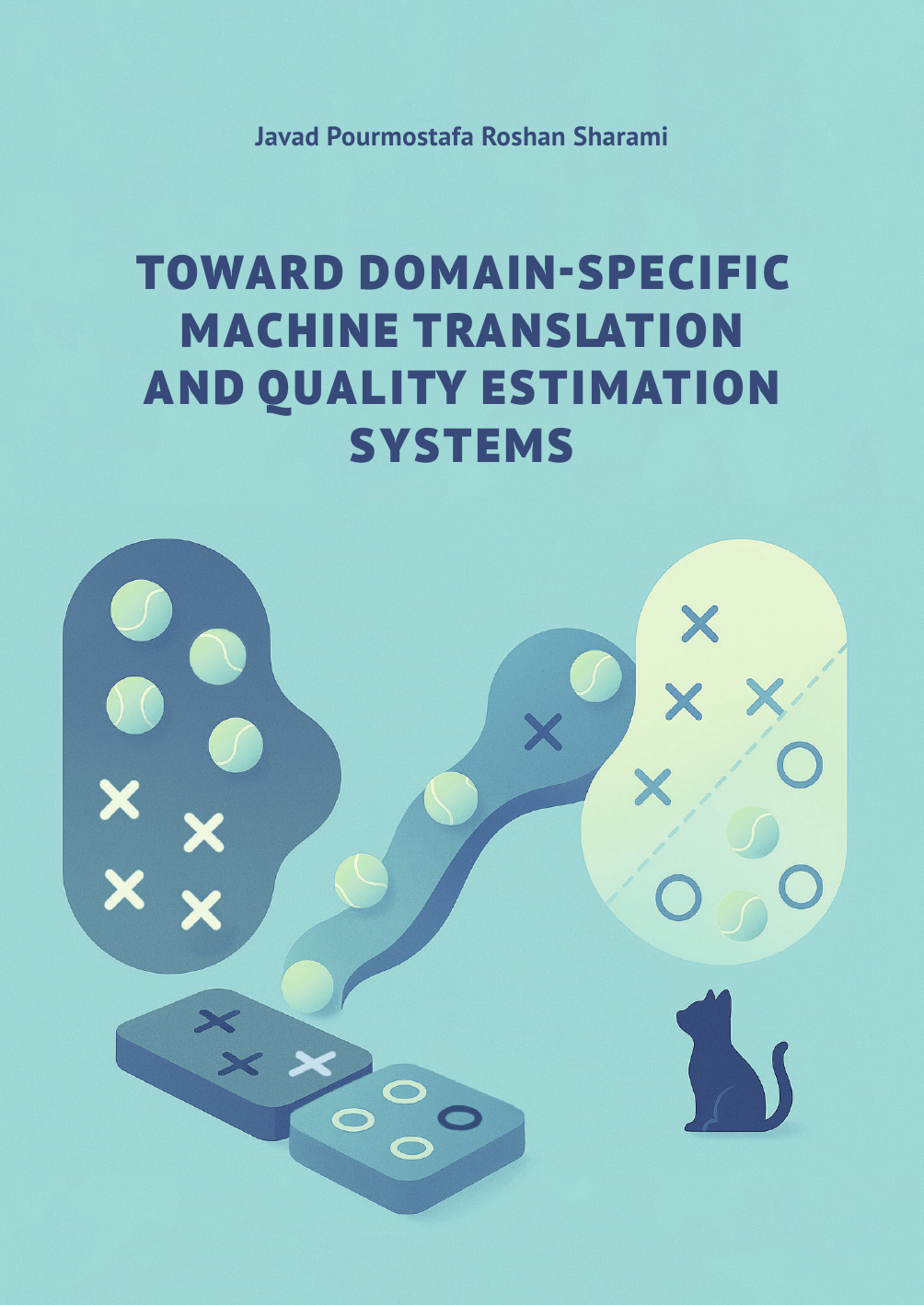}

\setcounter{page}{1}
\thispagestyle{empty}
\begin{center}

\vspace*{3cm}

{\Large\bfseries
Toward Domain-Specific Machine Translation\\[0.2cm]
and Quality Estimation Systems}\\[4cm]

{\small
Proefschrift ter verkrijging van de graad van doctor\\
aan Tilburg University\\[0.3cm]
op gezag van de rector magnificus, prof. dr. W.B.H.J. van de Donk,\\
in het openbaar te verdedigen ten overstaan van een door het college voor promoties aangewezen commissie\\
in de Aula van de Universiteit op woensdag 4 maart 2026 om 16.30 uur
}\\[2cm]

{\normalsize
door \textbf{Javad Pourmostafa Roshan Sharami}\\[0.25cm]
geboren te Someesara, Iran
}

\end{center}

\vfill
\newpage

\thispagestyle{empty}
\begin{flushleft}

\vspace*{1cm}

{\normalsize
Promotor:\\[0.2cm]
\hspace*{0.8cm}prof. dr. ir. P.H.M. Spronck (Tilburg University)\\[0.8cm]

Copromotor:\\[0.2cm]
\hspace*{0.8cm}dr. D. Shterionov (Tilburg University)\\[0.8cm]

Leden promotiecommissie:\\[0.2cm]
\hspace*{0.8cm}prof. dr. D. Kenny (Dublin City University)\\
\hspace*{0.8cm}prof. dr. J. Monti (Università di Napoli L’Orientale)\\
\hspace*{0.8cm}prof. dr. A. Alishahi (Tilburg University)\\
\hspace*{0.8cm}dr. A. Tezcan (Ghent University)\\
\hspace*{0.8cm}dr. A. Karakanta (Leiden University)
}\\[2cm]

{\footnotesize
ISBN 978-94-6537-044-6\\
Printed by Ridderprint (\href{https://ridderprint.nl}{ridderprint.nl})\\
Cover design by Arina van Londen (\href{https://arinavanlonden.com}{arinavanlonden.com})\\[0.4cm]
© 2026 Javad Pourmostafa Roshan Sharami, The Netherlands.\\
All rights reserved. No parts of this thesis may be reproduced, stored in a retrieval system, or transmitted in any form or by any means without the permission of the author.
Alle rechten voorbehouden. Niets uit deze uitgave mag worden vermenigvuldigd, in enige vorm of op enige wijze, zonder voorafgaande schriftelijke toestemming van de auteur.
}

\end{flushleft}
\vfill
\newpage

% --- DEDICATION PAGE ---
\thispagestyle{empty}
\vspace*{5.5cm}

\begin{center}
{\Large\itshape For Mahsa}
\end{center}

\vfill
\newpage

\tableofcontents

\chapter*{Glossary of Acronyms}
\addcontentsline{toc}{chapter}{Glossary of Acronyms}

\noindent\textit{This glossary provides an overview of the acronyms that appear frequently throughout the dissertation.}

\vspace{1.2em}

\renewcommand{\arraystretch}{1.19} 
\begin{table}[h!]
\centering
\begin{tabular}{p{0.20\textwidth} p{0.70\textwidth}}
\hline\hline
\textbf{Acronym} & \textbf{Definition} \\
\hline\hline
MT  & machine translation \\
SMT & statistical machine translation \\
QE  & quality estimation \\
NMT & neural machine translation \\
LLM & large language model \\
PLM & pre-trained language model \\
ICL & in-context learning \\
ICE & in-context example \\
DA  & domain adaptation \\
DAG & data augmentation \\
SW  & subword \\
ZSL & zero-shot learning \\
ZS  & zero-shot \\
RQ  & research question \\
NLP & natural language processing \\
\hline\hline
\end{tabular}
\end{table}

% -------
\chapterseparatorpage
\chapterimage{Assets/separator.pdf}

\chapter[Introduction]{Introduction}
\chaptermark{Introduction}

% MT advancements
\lettrine[lines=4]{M}{achine} translation (\glsentryshort{mt}) has come a long way since its inception, marked by major milestones, breakthroughs, and challenges.\glsunset{mt} One of the earliest significant events was the Georgetown-IBM experiment in 1954, a collaboration between IBM and Georgetown University that successfully translated some Russian sentences into English using the IBM 701 computer and a limited vocabulary. This early (although limited) success showcased the potential of automated translation and significantly boosted public interest in using computers for language translation~\cite{Hutchins2006TheFP}. However, soon after, due to the high costs and slower translation speeds compared to human translators, \gls{mt} faced a notable setback thanks to the ALPAC report in the 1960s, which led to a freeze in U.S. funding for \gls{mt} research~\cite{Nirenburg2003ALPACT}. While this report was disappointing, it also highlighted the danger of over-promising the capabilities of \gls{mt} systems~\cite{Koehn_2020}, a lesson that can be broadly applied across other AI-related fields. 

Despite these setbacks, research persisted, leading to the rise of \gls{smt} in the 1980s. \gls{smt} leveraged bilingual corpora and probabilistic methods, which led to enhanced translation accuracy compared to earlier paradigms, setting the stage for a more transformative breakthrough: \gls{nmt}. While \gls{nmt} is still fundamentally statistical, it replaces phrase-based tables and hand-crafted probability estimates with artificial neural networks that learn to directly model the conditional probabilities of translations. By the 2010s, deep learning techniques, particularly the Transformer model~\cite{NIPS2017_3f5ee243}, revolutionized \gls{mt} by enabling more fluent, context-aware translations. Building on these advancements, large-scale pre-trained models have further advanced \gls{mt}, improving, among others, adaptability for specialized domains and low-resource languages. These models go beyond word-for-word translation, capturing meaning, context, and nuance more effectively. Yet, despite these improvements, \gls{mt} continues to face ongoing challenges and new research directions.

A major driver of recent progress in \gls{mt} has been the availability of large parallel corpora, such as those provided by the Opus platform~\cite{tiedemann-2012-parallel}, which have strengthened data-driven \gls{mt} systems. Also, initiatives like the WMT shared tasks~\cite{wmt-2023} have played a crucial role in fostering advancements in translation quality, evaluation metrics, and \gls{qe}. These annual competitions provide standardized benchmarks, encourage the development of novel \gls{mt} techniques, and facilitate direct comparisons between different approaches. The growing integration of generative \glspl{llm} has also shifted \gls{mt} toward more flexible, in-context translation, allowing models to adapt to various domains with minimal fine-tuning.

Despite these advances, significant challenges and unexplored areas persist. Key issues include but are not limited to domain mismatch, support for rare words, addressing the long sentences issue, lacking the volume of training data––specifically in low-resource languages, word alignment, and decoding strategies~\cite{koehn-knowles-2017-six}. While some of these challenges have seen notable improvements––such as word alignment and handling rare words using \gls{nmt}%and subword tokenization techniques 
––they continue to be actively researched, with ongoing advancements aimed at further enhancing translation accuracy and adaptability.

Among these challenges, \textit{domain mismatch} persists despite ongoing research. This issue occurs when an \gls{mt} system is trained on data from one domain but used to translate text from a different domain, leading to vocabulary discrepancies and context mistranslations. For example, an \gls{mt} model trained solely on the chemical domain is unlikely to perform well on the IT domain, let alone reach human parity~\cite{saunders2021domain}. Addressing domain mismatch has been a central focus of research since the early days of \gls{mt}. It is safe to say that the study of domain mismatch and methods to mitigate this phenomenon is one of the most extensively researched areas in the field due to its widespread relevance and significant impact on various \gls{mt} applications. Beyond degrading \gls{mt} performance, domain mismatch also affects evaluation quality, both in reference-based and reference-free evaluation methods~\cite{rei-etal-2020-comet,sharami-etal-2023-tailoring,zouhar2024finetunedmachinetranslationmetrics}.

\section{Domains and how machine translation adapts to them}
To address domain mismatch effectively, we must first understand what constitutes a ``domain.'' As explored in the work of \citeA{van-der-wees-etal-2017-dynamic} and later \citeA{saunders2021domain}, the term ``domain'' has been defined in multiple ways, each depending on the perspective or application. One common definition involves the concept of \textit{provenance} or \textit{origin}—essentially, where the text comes from. This could refer to the source of the dataset (e.g., the EN-DE WMT \gls{qe} shared task 2023) or any distinguishing label that sets it apart from other types of data. 

Domains can also be understood in terms of \textit{topics} (e.g., IT or medical), which imply a concentration of field-specific vocabulary and terminology, or in terms of \textit{genre}, which includes not only subject matter but also stylistic and structural characteristics—factors that can vary significantly even within a single dataset, particularly in user-generated content~\cite{van-der-wees-etal-2017-dynamic}. While these perspectives are valuable, they often introduce ambiguities that make domain boundaries difficult to define consistently. 

In this dissertation, we therefore adopt a narrower view and focus primarily on \textit{provenance}, i.e., the origin of the training data. Provenance offers an observable and reproducible way of distinguishing between domains, since \gls{mt} datasets are typically collected, labeled, and released with reference to their source or origin. This makes it particularly suitable for experimental work in \gls{da}, where comparability and reproducibility across studies are essential.

The preference for a provenance-based definition of domain is mainly methodological. In \gls{mt}, adaptation depends more on linguistic form––vocabulary, phrasing, and style––than on conceptual topic. Provenance, which reflects where and how data were produced, provides a concrete and reproducible proxy for these characteristics. Topic-based labels, by contrast, are often subjective and vary even within a field; for example, ``medical'' datasets may range from clinical reports to public health texts. Thus, provenance offers a stable linguistic basis on which topic-based distinctions can be interpreted.

By grounding domain in provenance, this dissertation establishes a reproducible and linguistically informed framework for studying and improving \gls{da}. This framing also clarifies how topic-based interpretations relate to the empirical design of this work: provenance defines the data used for experimentation, while topic-based perspectives guide the interpretation of results across different applied areas such as healthcare, law, and IT.

In this work, \gls{da} of an \gls{mt} model refers to enabling the model to recognize and adapt to the \textit{origin or provenance} of training data, specifically its domain-specific characteristics as defined by source corpus, collection setting, or dataset label. The goal is to ensure that the \gls{mt} model can accurately translate within a specific domain, implying that it sufficiently covers the domain regardless of its type.

An \gls{mt} model that is adapted to a domain is expected to deliver high-quality translations that are faithful not only to the source text but also to the expectations of the domain it serves. For example, a model trained on medical datasets should accurately translate specialized terms, use the appropriate register\footnote{Throughout this dissertation, references to domains such as ``medical,'' ``legal,'' or ``IT'' are used as conceptual abstractions to help illustrate ideas, while the underlying analysis follows a provenance-based view. In other words, these topic labels serve as intuitive references for readers, but the research itself treats domains in terms of their linguistic and data provenance characteristics.} and avoid ambiguity that could lead to misunderstandings in critical contexts.

To address domain mismatch, researchers have proposed a broad range of \gls{da} techniques, which are generally grouped into two main categories: data-centric and model-centric, as summarized in the surveys by \citeA{chu-wang-2018-survey} and \citeA{saunders2022domain}. \textit{Data-centric approaches} emphasize curating or selecting training data that are representative of the target domain. \textit{Model-centric approaches}, on the other hand, focus on modifying the model’s learning process, architecture, or decoding strategies to make it more responsive to domain-specific input. The studies in this dissertation follow both of these categories, with a majority being data-centric. Specifically, Chapters~\ref{chap:CLIN}, \ref{chap:EAMT}, and \ref{chap:BPE} follow a data-centric approach, while Chapter~\ref{chap:AMTA} is model-centric.

% This dissertation draws from both types of approaches. Chapters~\ref{chap:CLIN}, \ref{chap:EAMT}, and \ref{chap:BPE} employ data-centric strategies to tackle DA, while Chapter~\ref{chap:AMTA} explores a model-centric solution. Together, these chapters offer a comprehensive view of how MT systems can be made more sensitive and responsive to domain-specific requirements.

\vspace{0.5em}

\noindent This dissertation presents a series of interconnected studies aimed at understanding and addressing the problem of domain mismatch in \gls{mt}. %To set the stage, we begin by clarifying two foundational concepts—\textit{domain} and \textit{domain adaptation}—which underpin the rest of this work. Once these concepts are established, 
In the remainder of this chapter, we introduce the problem statement addressed in this dissertation, followed by the specific \glspl{rq} that guide our investigation. We then conclude the chapter with an overview of the dissertation structure, outlining the rationale and relevance of each chapter in relation to the overarching research objectives.

\vspace{0.5em}

\section{Problem statement}
\label{Intro:problem-statement}
As language technology becomes more central to our daily lives, the role of \gls{mt} systems in supporting specialized domains––such as healthcare, law, and IT––has never been more critical. Yet, the very success of \gls{mt} hinges on its ability to adapt to the nuanced needs of these domains. A generic translation that misinterprets a medical dosage or a legal clause is not just a technical failure—it could be dangerous and/or costly. This underlines a key insight: not all data is created equal. That is, the quantity alone is not enough; the data's relevance and specificity to the domain truly determine \gls{mt} performance. As the volume of multilingual data continues to grow, it becomes clear that domain relevance matters more than ever~\cite{wang-etal-2017-sentence,d84596fb74844fd7a414e2f25ad81e3e}.

This insight has important consequences for how we build and deploy \gls{mt} systems. General-purpose models, including powerful \glspl{llm}, often struggle when confronted with the specialized terminology, style, and context required in professional settings~\cite{wassie2024domainspecific,Jiawei_Zheng_2024}. Even when fine-tuned, these models typically demand large volumes of domain-specific data to approach the quality of smaller, domain-specific systems~\cite{eschbach-dymanus-etal-2024-exploring}. And the challenge does not stop at translation. \gls{qe} systems, which assess the reliability of translations, are equally vulnerable to domain mismatch, leading to inaccurate quality scores, unnecessary human intervention, and increased costs.

This dissertation identifies several intertwined challenges to building \gls{mt} and \gls{qe} systems that truly support domain-specific needs: (1) Large-scale \gls{mt} systems––including \glspl{llm} such as GPT-4, used for instance in legal settings~\cite{wan2024reformulatingdomainadaptationlarge}––trained on general-domain data often fail to perform well in specialized domains due to a lack of domain-specific data. (2) Unlike \gls{mt}, which can often generalize across domains given sufficient training data, \gls{qe} models are especially vulnerable to domain mismatch. They depend on scarce, high-quality annotated datasets, making it challenging to ensure accurate and reliable automated translation evaluation in real-world applications~\cite{c-de-souza-etal-2014-machine}. (3) When adapting \gls{mt} models to new domains, simply introducing domain-specific data is not enough––without careful control over vocabulary and \gls{sw} tokenization, the model can misinterpret new terminology, leading to degraded translation quality and wasted training effort.  While the issue of unknown or out-of-vocabulary (OOV) words has been largely mitigated by advanced \gls{sw} tokenization algorithms, improper or inconsistent \gls{sw} splits can still lead to suboptimal translations~\cite{li2024evaluatinglargelanguagemodels}. (4) \gls{icl} allows \glspl{llm} to translate without fine-tuning, but their success depends entirely on finding the right examples—ones that match the domain, context, and complexity of the source text—yet identifying such examples without human input or costly reference translations remains a major challenge.
These challenges define the central problem addressed in this dissertation:

\vspace{0.5em}
\noindent
\textbf{Problem Statement:} \textit{How can we design \gls{mt} and \gls{qe} systems that are accurate, adaptable, and efficient across specialized domains?}
\vspace{0.5em}

To address this problem, in Chapter~\ref{chap:CLIN}, we begin by tackling one of the most fundamental issues: the scarcity of high-quality, domain-specific data. Moreover, we introduce a lightweight, data-centric methodology for selecting in-domain examples from generic corpora to improve translation quality while minimizing computational overhead. This approach is supported by an easy-to-use Python tool designed to help the broader research community replicate and scale the method efficiently. We also investigate how much in-domain data is needed to reach strong performance without incurring large computational costs.

Next, recognizing that domain mismatch also affects \gls{qe}, Chapter~\ref{chap:EAMT} explores the adaptation of \gls{qe} models using both \gls{da} and \gls{dag}.\footnote{``\gls{dag}'' denotes data augmentation, to avoid confusion with \gls{da}.} We demonstrate that just like \gls{mt} models, \gls{qe} systems benefit significantly from domain-specific training, especially when reference data are limited or unavailable.

Chapter~\ref{chap:BPE} investigates how the choice of vocabulary and \gls{sw} tokenization strategies impacts the added value of \gls{da}. Then, in Chapter~\ref{chap:AMTA}, we introduce an \gls{icl} approach that uses domain-adapted \gls{qe} models to select better\footnote{``better'' means \glspl{ice} that lead to better translation or faster inference.} \glspl{ice} for \glspl{llm}. This not only improves translation quality but also offers a more sustainable alternative to fine-tuning, with reduced carbon emissions and computational demands. 

In summary, these contributions offer practical, efficient, and scalable solutions for enhancing the adaptability of \gls{mt} and \gls{qe} systems across various domains.

\section{Research questions}
\label{Intro:research-questions}
To guide our investigation, this dissertation is structured around four core challenges. Each is framed as a main \gls{rq}––accompanied by sub-questions––that drives the research, development, literature review, and empirical efforts presented in the subsequent chapters.

\begin{enumerate}[label=\textbf{RQ\arabic*}, leftmargin=3em]

  \item \textit{What is the optimal amount of in-domain data required to achieve state-of-the-art \gls{mt} quality at low computational costs?}
  \begin{description}
    \item[RQ1a:] How does the quality of selected in-domain data affect translation performance?
    \item[RQ1b:] What trade-offs arise between translation performance and computational cost when using in-domain versus generic-domain data?
  \end{description}
\end{enumerate}

In Chapter~\ref{chap:CLIN}, we explore whether it is possible to find and extract in-domain data from publicly available bilingual corpora to support high-quality, domain-adapted \gls{mt}. To this end, we introduce a data selection method tailored to retrieve similar\footnote{The specific type of similarity is discussed in the relevant chapter.} in-domain examples from generic datasets and run experiments to see how they compare to generic-domain data––especially in terms of how much performance gain they offer for the computational effort required.

\begin{enumerate}[label=\textbf{RQ\arabic*}, leftmargin=3em, resume]
  \item \textit{What are the effects of combining \gls{da} and \gls{dag} on the performance and generalizability of \gls{qe} models across different domains and languages?}
  \begin{description}
    \item[RQ2a:] How does \gls{da} impact \gls{qe} performance in low-resource scenarios?
    \item[RQ2b:] What challenges arise when applying \gls{dag} in \gls{qe} across diverse languages?
  \end{description}
\end{enumerate}

In Chapter~\ref{chap:EAMT}, we explore whether \gls{qe} models, like \gls{mt} systems, would benefit from being tailored to specific domains. Moreover, we introduce a \gls{qe} methodology that combines \gls{da} and \gls{dag}, and show how it can make \gls{qe} systems more robust––especially in multilingual and domain-diverse scenarios, including \gls{zsl} and cross-lingual settings.

\begin{enumerate}[label=\textbf{RQ\arabic*}, leftmargin=3em, resume]
  \item \textit{What is the optimal combination of \gls{sw} tokenization model source and vocabulary source to maximize translation quality and efficiency during fine-tuning?}
  \begin{description}
    \item[RQ3a:] How do different sources for training the \gls{sw} tokenization model affect translation quality when fine-tuning out-of-domain models on in-domain data?
    \item[RQ3b:] How do the choice of sources for the \gls{sw} model and vocabulary impact the trade-off between computational efficiency (e.g., training time, resource consumption) and model accuracy in \gls{da}?
  \end{description}
\end{enumerate}

In Chapter~\ref{chap:BPE}, we investigate how the choice of data sources for training the vocabulary and \gls{sw} tokenization model affects the fine-tuning of \gls{mt} models for in-domain translation, starting from a model trained on out-of-domain data. We investigate which sources are most effective for learning the tokenization model and extracting vocabularies, and how these choices impact both translation quality and fine-tuning efficiency.

\begin{enumerate}[label=\textbf{RQ\arabic*}, leftmargin=3em, resume]
  \item \textit{How effective are domain-specific \gls{qe} models in determining effective \glspl{ice} for improving \gls{mt} quality in generative \glspl{llm}?}
  \begin{description}
    \item[RQ4a:] What criteria should be used to select effective \glspl{ice} for improving \gls{mt} quality?
    \item[RQ4b:] How does the integration of \gls{qe} affect the success of \gls{ice} selection?
  \end{description}
\end{enumerate}

In Chapter~\ref{chap:AMTA}, we propose an \gls{icl} methodology that leverages domain-specific \gls{qe} to guide the selection of \glspl{ice} for \glspl{llm}. The goal is to identify empirically effective numbers and combinations of \glspl{ice} that improve translation quality without relying on reference translations. %We further examine whether this \gls{icl} approach offers advantages over fine-tuning a pre-trained \gls{mt} model.

% In Chapter~\ref{chap:AMTA}, we investigate whether \gls{icl} offers advantages over fine-tuning a pre-trained \gls{mt} model. To this end, we propose an \gls{icl} methodology that uses domain-specific \gls{qe} to guide the selection of \glspl{ice}. The goal is to identify empirically effective numbers and combinations of \glspl{ice} that improve translation quality, without relying on reference translations.

\section{Dissertation overview}
This section outlines the motivations, significance, and main contributions of this dissertation, while highlighting how each contribution connects to the chapters that follow. Building on the RQs introduced earlier, we begin by examining the importance of in-domain data and its influence on both \gls{mt} and \gls{qe}, with a focus on the challenges posed by domain mismatch. We then turn to the role of \gls{sw} tokenization and vocabulary design in enabling effective \gls{da} for \gls{mt}. Finally, we discuss \gls{icl} for \gls{mt}, emphasizing how the selection of \gls{icl} can significantly impact translation quality.

\subsection{Why in-domain data matters}
As the volume of data to develop \gls{mt} grows, it becomes clear that some data are more relevant to specific domains than others. This highlights the need for adaptation of models tailored to specific use cases, such as medical or patent translations. This requirement extends to nearly all \gls{nlp} downstream tasks, including \gls{mt} and \gls{qe}. Generic models, including \glspl{llm}, often struggle in these contexts, failing to handle domain-specific style, terminology, and levels of formality. Even when fine-tuned, \glspl{llm} require large amounts of domain-specific data to match the performance of smaller, dedicated translation systems~\cite{eschbach-dymanus-etal-2024-exploring}. Training \gls{mt} models with datasets that cover domains similar to the target domain can effectively yield high-quality results with practical usefulness in domain-specific applications~\cite{wang-etal-2017-sentence,d84596fb74844fd7a414e2f25ad81e3e}. This highlights the limitations of data-driven \gls{mt} when relying on general-domain data, regardless of its size (as demonstrated in Chapter~\ref{chap:CLIN}), and emphasizes the importance of using domain-specific datasets for building effective \gls{mt} systems.

However, acquiring sufficient authentic domain-specific data with human annotators, though ideal from a quality and reliability perspective, is labor-intensive and costly. Human-annotated data ensures high accuracy, domain expertise, and adherence to stylistic and terminological conventions, making it the gold standard for training domain-specific \gls{mt} models. However, due to its high cost and limited availability, researchers have explored alternative approaches for generating synthetic or pseudo-parallel datasets through data-centric approaches, primarily selecting in-domain data from generic parallel corpora. Chapter~\ref{chap:CLIN} of this dissertation presents a methodology for selecting in-domain data for \gls{mt} through a data-centric method.
After pre-processing steps to control computational costs, we identify and select examples with sentences similar to the target domain. This approach aims to operate with minimal computational costs and data capacities, thereby reducing the overhead associated with the data selection process. 

Our experiments highlight the trade-off between translation performance and efficiency, emphasizing the value of smart data selection in low-resource, domain-specific settings. Additionally, we have developed a Python tool that simplifies the data selection process through easy command-line operation and customizable configurations, making it more accessible to the community. This tool, which supports the data selection methods proposed in our work, is discussed in Section~\ref{CLIN:tool} of Chapter~\ref{chap:CLIN} and can be found at~\url{https://github.com/JoyeBright/domain-adapt-mt}.

Beyond \gls{mt}, \gls{qe} plays a crucial role in evaluating \gls{mt} output without requiring human references. \gls{qe} models predict the quality of translations, offering insights into whether a machine-translated sentence requires post-editing or is suitable for direct use. Unlike \gls{mt}, which focuses on generating translations, \gls{qe} aims to assess their reliability, making it a valuable tool for automated quality control in real-world applications. 

However, \gls{qe} introduces additional complexities, particularly in domain-specific scenarios, where training data scarcity and domain mismatches pose significant challenges. \gls{qe} systems require post-edited text alongside source and machine-translated text for training, but such datasets are scarce. Compared to the abundance of \gls{mt} datasets, publicly available \gls{qe} datasets are significantly fewer, making the development of reliable \gls{qe} models more challenging. While both \gls{mt} and \gls{qe} face challenges with generalizability, \gls{qe} is significantly more affected. \gls{qe} models are highly sensitive to domain shifts, leading to unreliable predictions when trained on data that does not match the \gls{mt} system's domain. Therefore, ensuring accurate \gls{qe} across different domains––both general and specialized––remains a critical yet challenging task, further exacerbated by the scarcity of high-quality training data.

In response to this challenge, Chapter~\ref{chap:EAMT} is guided by the view that \textit{\gls{qe} models, like \gls{mt} models, require \gls{da} to accurately assess the quality of \gls{mt} outputs.} In this context, the comparison implies that just as \gls{mt} models depend on \gls{da} to improve translation quality, \gls{qe} models similarly require \gls{da} to enhance their ability to assess \gls{mt} outputs effectively. This idea is particularly important because \gls{mt} practitioners and translation experts depend on the precision of \gls{qe} models. Domain mismatch issues can lead to inaccurate quality assessments, potentially increasing costs for essential tasks such as dissemination.

To investigate this perspective, we first demonstrate the necessity of \gls{da}. Building on this, Chapter~\ref{chap:EAMT} introduces a \gls{qe} methodology that integrates both \gls{da} and \gls{dag}, supported by a novel \gls{qe} training pipeline. We evaluated this methodology using several publicly available language pairs, including English to German, Chinese, Italian, Czech, Japanese, Romanian, and Russian to English and showed significant improvements across all language pairs under consideration, indicating that our proposed solution has a beneficial impact in addressing the aforementioned challenges.

The chapter also investigates how the proposed approach performs in \gls{zs} and cross-lingual settings, where no training data is available for certain language pairs or domains. The analysis aims to evaluate how our approach affects \gls{qe} model performance in situations where there is no prior training data for specific language pairs or domains, and to identify the factors that contribute to its effectiveness or limitations. This is especially important given the significant lack of \gls{qe} data across many languages and domains, which makes robust, adaptable solutions all the more necessary.

\subsection{\gls{sw} choices and vocabulary in domain adaptation}

In Chapter~\ref{chap:EAMT} of this dissertation, we present experiments showing that pre-trained language models, when used without any adaptation (i.e., \gls{zs} inference), perform poorly. To improve this, the standard practice is to fine-tune the models. This involves training a generic model initially and then refining it with a domain-specific dataset different from the original training data, tailored to the specific downstream task, such as \gls{mt} in our case \cite{chu-wang-2018-survey,saunders2022domain}. Despite being more resource-intensive compared to parameter-efficient fine-tuning (PEFT) techniques like adapter-based methods~\cite{https://doi.org/10.48550/arxiv.1902.00751}, fine-tuning generally achieves better performance, particularly with smaller, memory-efficient pre-trained \glspl{llm}. These models are easier to manage in terms of memory, as opposed to larger \glspl{llm}, which require significant memory just to load and update. Therefore, fine-tuning still remains a viable option for \gls{da} in \gls{mt}.

Introducing new domain-specific data during fine-tuning brings additional information, such as previously unseen words. However, this can also lead to suboptimal word tokenization if not handled properly, potentially degrading system performance \cite{lim2018exploring,essay80128,sato-etal-2020-vocabulary}. Therefore, identifying the optimal setup for incorporating new data is vital for maintaining and enhancing \gls{mt} performance. This includes ensuring that both the pre-trained model's existing knowledge and the new information are effectively covered.

To address this, Chapter~\ref{chap:BPE} investigates the impact of vocabulary and \gls{sw} tokenization on fine-tuning \gls{mt} models for in-domain translation. Our research aims to provide insights into the ideal conditions for fine-tuning, with the goals of improving translation quality and reducing training time. In Chapter~\ref{chap:BPE}, we seek to answer the primary \gls{rq}: \textit{Given an \gls{mt} model and a fine-tuning dataset, what is the optimal combination of \gls{sw} model source and vocabulary?} We systematically compare different configurations of \gls{sw} tokenization models and vocabulary sources to identify which combinations yield the most effective fine-tuning of \gls{mt} models for specific domains. Through this evaluation, we provide practical guidance for selecting and aligning \gls{sw} tokenization and vocabulary when adapting \gls{mt} models to new domains.

\subsection{\glspl{ice} for \glspl{llm} in \gls{mt}}
Recently, the increasing popularity of generative \glspl{llm} has introduced a new paradigm known as \gls{icl}. Unlike traditional fine-tuning methods, \gls{icl} typically generates output directly, without modifying the model parameters~\cite{Radford2019LanguageMA,NEURIPS2020_1457c0d6}. This is achieved by providing the model with a few examples, known as \glspl{ice}, which prime the \gls{llm} to enhance its performance for the given task~\cite{10.1162/tacl_a_00324}. In our case, this task is \gls{mt}.
This approach reduces the computational costs associated with fully fine-tuning generative \glspl{llm}, which typically have billions of parameters. However, as demonstrated by~\cite{vilar-etal-2023-prompting}, a key challenge with generative \glspl{llm} is that the quality of translation is directly proportional to the quality of \glspl{ice}, where quality refers to \glspl{ice} being relevant, clear, accurate, and domain-specific. Therefore, \textit{finding \glspl{ice}} that are most relevant and domain-specific to the text to be translated is another crucial challenge for improving \gls{mt} performance.

To tackle this issue, a key question arises: \textit{How can one determine if the selected \glspl{ice} are effective, i.e., relevant to the domain and context of the source text to be translated, and thus improve \gls{mt} performance?} Although this evaluation typically relies on translated references or human judgments, the former can introduce bias, and the latter is costly~\cite{dinh-niehues-2023-perturbation}. In Chapter~\ref{chap:AMTA}, we explore whether integrating \gls{qe} could potentially mitigate reference bias and offer a better approach for selecting more relevant \glspl{ice} (compared to relying on development sets and other existing approaches~\cite{sia-duh-2023-context,agrawal-etal-2023-context}), thereby improving translation quality and efficiency in LLM-based translation systems.

Previous findings from Chapter~\ref{chap:EAMT} and \cite{sharami-etal-2023-tailoring} highlight that \gls{qe} must be domain-specific to accurately estimate \gls{mt} quality across various domains. If a generic \gls{qe} model––one that is not tailored to specific domains––is used to select \glspl{ice}, it may fail to capture domain-specific nuances. Building on this, Chapter~\ref{chap:AMTA} explores whether such models may lead to the selection of ineffective \glspl{ice}, ultimately leading to suboptimal translations. This highlights the importance of \gls{da} in \gls{qe} for guiding effective \gls{ice} selection.

To explore this question, in Chapter~\ref{chap:AMTA}, we answer the following \gls{rq}: \textit{How effective are domain-specific \gls{qe} models in guiding the selection of effective \glspl{ice} for translation tasks in \glspl{llm}?} To answer this question, we propose a novel \gls{icl} methodology that leverages domain-specific \gls{qe} to assist in the selection of \glspl{ice}, with the goal of identifying empirically effective (suboptimal)\footnote{Here, ``suboptimal'' means empirically effective \gls{ice} numbers and combinations identified under experimental constraints, without a claim of theoretical optimality.} numbers and combinations of \glspl{ice} that improve \gls{mt} quality, all without reference translations. We also assessed the impact of using our proposed methodology in comparison to fine-tuning a pre-trained multilingual \gls{mt} model, namely mBART-50~\cite{tang2020multilingual}. By considering all computational factors, this comparison allows us to determine whether \gls{icl} presents a more advantageous approach compared to fine-tuning a pre-trained \gls{mt} model. Our experiments highlight the value of \gls{icl} over fine-tuning, showing that our method not only achieves better translation performance but also reduces carbon emissions, making it a more sustainable and efficient alternative. 

\section{Evaluation metrics}
\label{Intro:eval-metrics}

Throughout this dissertation, we evaluate translation models and \gls{qe} systems using both traditional and modern metrics. Below is a brief overview of each:

\begin{itemize}
  \item \textbf{BLEU} measures n-gram overlap between the \gls{mt} output and a reference. Scores range from 0 to 100; scores above 30 are considered reasonable, and above 40 strong. BLEU tends to favor surface-level similarity and may not fully capture adequacy~\cite{papineni-etal-2002-bleu}.

  \item \textbf{COMET} is a neural metric trained on human ratings. Scores typically range from -1 to +1. A score above 0.3 is decent; 0.5+ is strong. COMET correlates well with human judgment~\cite{rei-etal-2020-comet}.

  \item \textbf{TER} counts the edits needed to match a reference. Lower is better. TER below 50 is acceptable; below 40 is strong. TER penalizes word order and omission errors more explicitly~\cite{snover-etal-2006-study}.

  \item \textbf{chrF2} is a character-level F-score. Scores range from 0 to 100. chrF2 above 60 is typical for good systems and is useful for morphologically rich languages~\cite{popovic-2015-chrf}.
\end{itemize}

\paragraph{Interpreting metric scores.}
While thresholds like BLEU >30 or COMET >0.3 are often used to indicate translation quality, their interpretation depends on dataset, language pair, references, and preprocessing~\cite{zouhar-etal-2024-pitfalls,tutorialsdojo_bleu}. We therefore focus on relative comparisons with strong baselines and support findings with statistical significance tests. Both absolute scores and relative gains are reported and interpreted in context.

\section{Computational considerations and efficiency}
\noindent 
Computational efficiency is a key consideration in modern \gls{mt} research, where model training, fine-tuning, and inference often demand substantial GPU resources, energy, and time.
Optimizing these computational costs not only reduces training overhead but also plays a vital role in ensuring the scalability and sustainability of \gls{mt} systems––especially when deployed in real-world settings.

In this dissertation, we systematically examine the computational demands of the experiments conducted throughout our work. We highlight the trade-offs associated with each approach in terms of efficiency, training time, inference speed, and environmental impact. Specifically: Chapter~\ref{chap:CLIN} quantifies the efficiency gains from data selection in reducing training costs; Chapter~\ref{chap:EAMT} examines the trade-offs of \gls{da} pipelines; Chapter~\ref{chap:BPE} investigates the impact of \gls{sw} tokenization on computational efficiency; and Chapter~\ref{chap:AMTA} evaluates the computational performance of \gls{icl}, including inference time, energy consumption, and sustainability. A more detailed, chapter-by-chapter breakdown of these computational trade-offs––including training times, energy usage, and carbon impact––is provided in the appendix. %~\ref{Appendix:ComCosts}.

\section*{Carbon footprint acknowledgment}
\noindent
Throughout this research, we tried to reduce computational costs and develop alternative methods to improve efficiency. However, the investigation, model training, and inference phases still required significant processing power and runtime. The total carbon footprint of the experiments is estimated at approximately 1.5 metric tons of CO\textsubscript{2}.

To help mitigate this impact, I chose to support a certified reforestation initiative through \href{https://treesforall.nl/}{\textit{Trees for All}}.\footnote{Trees for All is a Dutch non-profit organization dedicated to sustainable forestry and climate compensation through certified tree-planting projects.} A donation equivalent to the estimated emissions was made to their program, and a certificate confirming the offset is available.\footnote{\url{https://javad.pourmostafa.me/assets/files/C02Offset.pdf}}

\section{List of publications}

This dissertation builds on and extends the following peer-reviewed publications:

\begin{itemize}
    \item Javad Pourmostafa Roshan Sharami, Dimitar Shterionov, and Pieter Spronck. \href{https://www.clinjournal.org/clinj/article/view/137}{Selecting Parallel In-Domain Sentences for Neural Machine Translation Using Monolingual Texts}. \textit{Computational Linguistics in the Netherlands Journal}, vol. 11, Dec. 2021, pp. 213–230.

    \item Javad Pourmostafa Roshan Sharami, Dimitar Shterionov, and Pieter Spronck. 2023. \href{https://aclanthology.org/2023.crowdmt-1.4/}{A Python Tool for Selecting Domain-Specific Data in Machine Translation}. In \textit{Proceedings of the 1st Workshop on Open Community-Driven Machine Translation}, pp. 29–30, Tampere, Finland. European Association for Machine Translation.

    \item Javad Pourmostafa Roshan Sharami, Dimitar Shterionov, Frédéric Blain, Eva Vanmassenhove, Mirella De Sisto, Chris Emmery, and Pieter Spronck. 2023. \href{https://aclanthology.org/2023.eamt-1.2/}{Tailoring Domain Adaptation for Machine Translation Quality Estimation}. In \textit{Proceedings of the 24th Annual Conference of the European Association for Machine Translation}, pp. 9–20, Tampere, Finland. European Association for Machine Translation.

    \item Javad Pourmostafa Roshan Sharami, Dimitar Shterionov, and Pieter Spronck. 2024. \href{https://amtaweb.org/amta-2024-program/}{Guiding In-Context Learning of LLMs through Quality Estimation for Machine Translation}. In \textit{Proceedings of the 16th Biennial Conference of the Association for Machine Translation in the Americas}, (Volume 1: Research Track), pp. 88–101, Chicago, USA. Association for Machine Translation in the Americas.

    \item Javad Pourmostafa Roshan Sharami, Dimitar Shterionov, and Pieter Spronck. \href{https://acl-bg.org/proceedings/2025/RANLP\%202025/pdf/2025.ranlp-1.111.pdf}{Analysis of Vocabulary and Subword Tokenization Settings for Optimal Fine-tuning of MT: A Case Study of In-domain Translation}. In \textit{Proceedings of the 15th International Conference on Recent Advances in Natural Language Processing}, pp. 970–979, Varna, Sep 8–10, 2025.
\end{itemize}

\vspace{1em}
\noindent
I also contributed to several other publications during my PhD. Although they are not included in this dissertation, I list them below to provide additional context for my research trajectory.

\begin{itemize}
    \item \textbf{Industry Collaboration:}  
    Elena Murgolo, Javad Pourmostafa Roshan Sharami, Dimitar Shterionov. 2022. \href{https://aclanthology.org/2022.eamt-1.43/}{A Quality Estimation and Quality Evaluation Tool for the Translation Industry}. In \textit{Proceedings of the 23rd Annual Conference of the European Association for Machine Translation}, pp. 307–308, Ghent, Belgium.  
    \\\\
    \textit{This work was conducted in collaboration with Aglatech14 (Orbital14), a language service provider. We developed and deployed a tailored machine translation and quality assessment tool for industry use.}
    
    \item \textbf{Supervision Output:}
    \begin{itemize}[leftmargin=1.5em]
        \item Ali Boluki, Javad Pourmostafa Roshan Sharami, Dimitar Shterionov.
        \href{https://doi.org/10.1007/978-3-031-47718-8_2}{\textit{Evaluating the Effectiveness of Pre-trained Language Models in Predicting the Helpfulness of Online Product Reviews}}.
        In \textit{Intelligent Systems and Applications (IntelliSys 2023)}, Lecture Notes in Networks and Systems, vol. 825. Springer, Cham.
        
        \item Adil Derrazi, Javad Pourmostafa Roshan Sharami.  
        \href{https://link.springer.com/chapter/10.1007/978-3-031-99958-1_27}{\textit{Integrating SAINT with Tree-Based Models}}.
       In \textit{Intelligent Systems and Applications (IntelliSys 2025)}, Lecture Notes in Networks and Systems, vol 1553. Springer, Cham.
    \end{itemize}

    \textit{These papers resulted from supervising MSc thesis students within the Data Science and Society program at Tilburg University.}
    
\end{itemize}

%-----------Main Work--–––––––––––––
\chapterseparatorpage
\chapterimage{Assets/separator.pdf}

\prechaptertext{This chapter is based on the following published paper:
\\\\
\hspace*{0.5em}Javad Pourmostafa Roshan Sharami, Dimitar Shterionov, and Pieter Spronck. \href{https://www.clinjournal.org/clinj/article/view/137}{Selecting Parallel In-Domain Sentences for Neural Machine Translation Using Monolingual Texts}. Computational Linguistics in the Netherlands Journal, vol. 11, Dec. 2021, pp. 213-230.
\\\\
The Python tool developed from this research was also published:
\\\\
\hspace*{0.5em}Javad Pourmostafa Roshan Sharami, Dimitar Shterionov, and Pieter Spronck. 2023. \href{https://aclanthology.org/2023.crowdmt-1.4/}{A Python Tool for Selecting Domain-Specific Data in Machine Translation}. In Proceedings of the 1st Workshop on Open Community-Driven Machine Translation, pp. 29–30, Tampere, Finland. European Association for Machine Translation.
}
\\\\
{Improvements have been made to the title, figures, and certain sections to align the content with the broader context of this dissertation.}

\chapter[Generating In-domain data for Machine Translation]{Generating In-domain data for Machine Translation}
\chaptermark{CLIN}
\label{chap:CLIN}

\lettrine[lines=4]{C}{ontinuously-growing} data volumes lead to larger generic \gls{mt} models. Specific use cases are usually left out since generic models tend to perform poorly in domain-specific cases. In this chapter, we address this gap with a method for selecting in-domain data from generic-domain (parallel text) corpora for the task of \gls{mt}. This method provides a solution to the problem by efficiently extracting relevant data that aligns with the specific domain, ensuring better performance in specialized translation tasks than generic models. The proposed method ranks sentences in parallel general-domain data according to their cosine similarity with a monolingual domain-specific dataset. We then select the top $K$ sentences with the highest similarity score, which can be used to train a new machine translation system tuned to the specific domain. There, we define a domain as a collection of data that shares a common provenance—texts originating from similar sources or contexts that exhibit consistent linguistic characteristics. Our experimental results show that models trained on this in-domain data outperform models trained on generic or a mixture of generic and in-domain data. Also, our method selects high-quality domain-specific training instances at low computational cost and data size.

\newpage

\section{Introduction}

A widely accepted belief among \gls{mt} researchers and practitioners is that more training data is better. That is, the larger the training corpus is, the more robust and accurate the model can be. However, substantial amounts of parallel data are not available for all language pairs or domains of interest \cite{currey-etal-2017-copied,van-der-wees-etal-2017-dynamic,stergiadis2021multidomain}. Furthermore, data-driven machine translation systems' performance depends not only on the quantity but also on the quality of available training data \cite{fadaee-etal-2017-data}. Despite the fact that more and more training data for \gls{mt} are becoming accessible every day, only those that cover the same or similar domains of interest are commonly able to boost translation quality \cite{wang-etal-2017-sentence,d84596fb74844fd7a414e2f25ad81e3e}. Hence, for domain-specific use-cases, data-driven paradigms may perform poorly when trained on general-domain data, regardless of the size of the corpus. Training \gls{mt} systems on large amounts of data in many cases uses substantial amounts of resources such as memory and time which is the undesired effect of striving to boost the performance of \gls{mt} systems. As such, it is paramount to be able to train systems on high-quality domain-specific data. We, therefore, face a two-sided challenge: (i) what is high-quality, in-domain data and (ii) what amounts of parallel, in-domain data are necessary to achieve state-of-the-art \gls{mt} quality at low computational and data capacities.

To address these challenges, the research community has made many efforts to improve \gls{mt} performance through \gls{da} techniques.
\gls{da} for \gls{mt} has versatile definitions, but we primarily follow \citeA{chu-wang-2018-survey}, who state that \emph{\gls{da} would be employing out-of-domain parallel corpora and in-domain monolingual corpora to improve in-domain translation}. Among other definitions, \citeA{saunders2021domain} defined \gls{da} as any scheme that aims to level up translation's performance from an existing system for a specific topic or genre language. Studies in this area are mainly divided into two categories, namely (i) data-centric and (ii) model-centric~\cite{chu-wang-2018-survey}. The data-centric category includes methods that operate at the corpus / data level by selecting, generating, joining, or weighting sentences or datasets for training purposes. This category selects or generates the domain-related sentences from the general domain using existing in-domain / domain-specific data. In contrast, in the model-centric category, studies mostly fall into the areas aiming to alter the usual function of models. This is usually fulfilled through mixing, fine-tuning and reordering models or weighting objective functions, or applying regularization techniques.

Our proposed methodology falls into a data-centric category and is specifically considered a data selection method. However, a few previous studies have investigated the generation of parallel in-domain sentences for \gls{mt}. Their limitations motivate us to expand this area of research by proposing a novel data selection algorithm for collecting in-domain training data. %and each has its limitations which results in motivation for us to expand this area of research. i.e., selection of in-domain training data employing a data selection approach. 
In this regard, we aim to improve in-domain translation in low-resource scenarios by selecting in-domain sentences from out-of-domain corpora, then possibly employing \gls{da} for \gls{nmt} leveraging both out-of-domain parallel corpora and in-domain monolingual data. In essence, our proposed approach leads to one main contribution: \emph{a language-agnostic data selection method for generating a parallel in-domain corpus using monolingual (untranslated) domain-specific corpora}. Monolingual corpora are often abundant and can easily be collected without requiring any translations or further sentence alignments. This has two consequences. The first one is related to the proportion of high-quality data to the number of in-domain sentences. In particular, our method generates fewer but of higher quality sentences with the same or at least competitive performance on \gls{nmt} systems. The higher quality is quantified by the efficiency of using fewer sentences that are more impactful, allowing the model to perform just as well, if not better. The second one is the reduction of training time. This is a consequence of the fact that less data is used for training an \gls{nmt} system. 

%In this work, we created a large EN$\Rightarrow$FR corpus, using several smaller corpora. This is mainly because a general-domain corpus should be sufficiently broad in terms of sentence diversity such that it increases the number of in-domain data. Likewise, a monolingual in-domain corpus containing in-domain sentences (either EN or FR) was utilized. Both corpora were then transformed into word embeddings for further analysis. %A dimensionality reduction technique, i.e. Principal Component Analysis (PCA), was applied to them to mitigate the computational costs. Dimensionality reduction will be discussed in detail in Section~\ref{CLIN:sec:dimensionality_reduction}.%
%The in-domain vectors were compared to out-of-domain sentences, then similar embedded vectors were ranked in descending order to generate an in-domain parallel corpus. The ranked sentences were also mixed to increase the amount of training data. Eventually, each one was fed into the MT systems to be trained and subsequently, the best translation indicates the best quality of an in-domain parallel corpus. This abstract perception is depicted in Figure \ref{CLIN:fig:method_overview}.% revised based on the comment

\begin{figure}[ht]
	\centering 
	\includegraphics[width=3.2in]{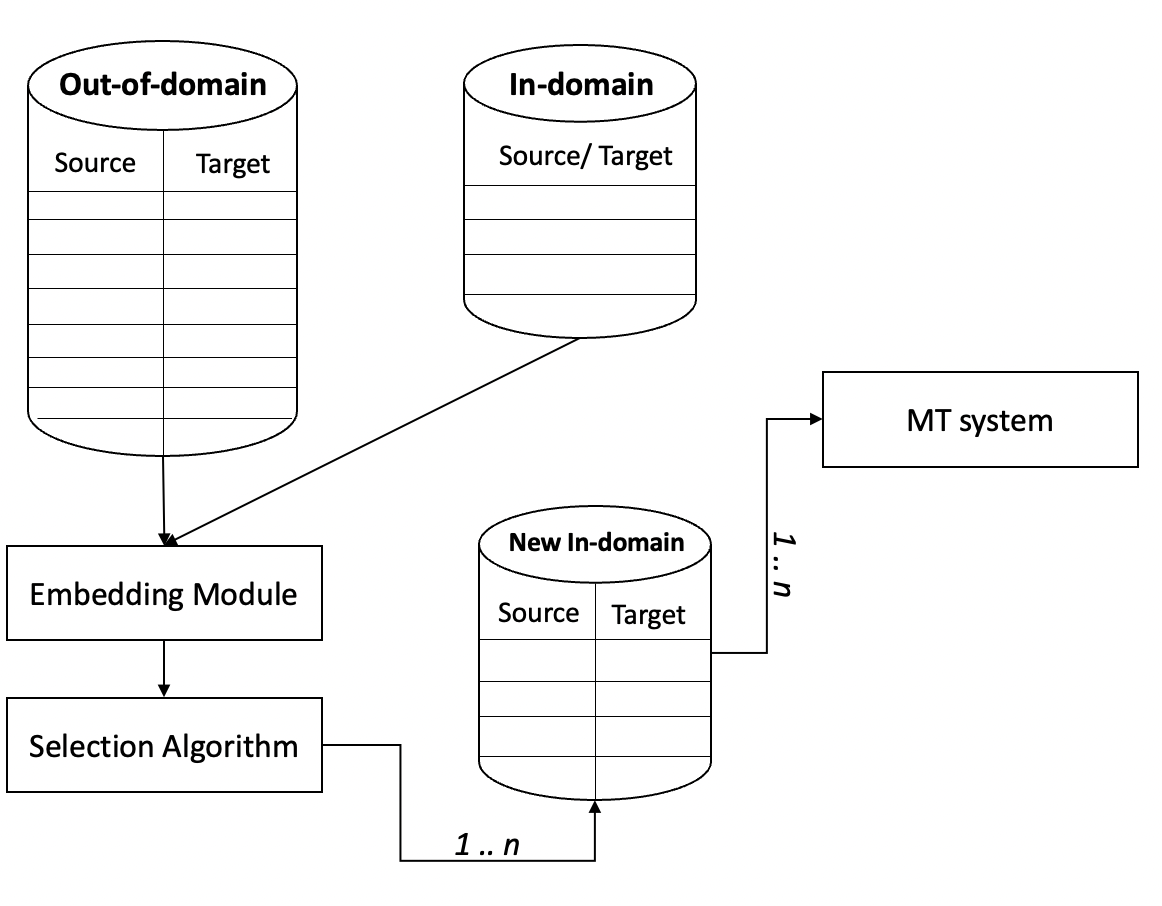} 
	\caption{\textbf{An overview of the proposed methodology.} The notation $1 .. n$ in the figure indicates that the algorithm can select between $1$ and $n$ sentences, where $n$ is an arbitrary number. For example, if $n = 5$, the algorithm selects up to $5$ parallel instances from the out-of-domain corpus for each in-domain sentence.}
	\label{CLIN:fig:method_overview}
\end{figure}

In a nutshell, we started with two corpora: a parallel general-domain EN$\rightarrow$FR corpus and a monolingual in-domain corpus (either EN or FR). The general-domain corpus provides broad sentence diversity, which helps increase the number of in-domain data. Meanwhile, the in-domain corpus ensures the inclusion of domain-specific sentences. Both corpora were converted into word embeddings for analysis. To compare the corpora, we measured the similarity between in-domain and general-domain sentence embeddings using cosine similarity. Based on this measure, we ranked the general-domain sentences in descending order of similarity to the in-domain corpus, creating an in-domain parallel corpus. The resulting ranked sentences were also mixed to augment the training data. Each dataset was then used to train \gls{mt} systems, and the best translations determined the quality of the in-domain parallel corpus. An overview of this methodology is presented in Figure~\ref{CLIN:fig:method_overview}.

The chapter is organized as follows. We first cover the related work and define specialized terminology in Section~\ref{CLIN:sec:related_work}. In Section~\ref{CLIN:sec:data_selection}, our data selection strategy is presented. Next, Section~\ref{CLIN:sec:experiments} covers the evaluation of our approach, detailing the datasets, system specifications, baselines, and results. Section~\ref{CLIN:sec:discussion} presents a discussion of our analysis. Section~\ref{CLIN:sec:conclusions} concludes the work.

\section{Related work}
\label{CLIN:sec:related_work}
There is a significant volume of research in DA for \gls{mt} paradigms. However, to the best of our knowledge, few prior studies have been conducted particularly on selecting in-domain sentences efficiently and then exploiting them to improve the in-domain translation. We reviewed data-centric and, more specifically, data-selection papers that were closely related to our research. \citeA{Luong2015StanfordNM} did major work in this area when they adapted an existing English-German deep Long Short-Term Memory (LSTM) model by training it for additional 12 epochs on new domain data in the same languages; the original training data is general-domain, while the one used for adaptation is from the conversational domain. This \gls{da} approach led to an increase by 3.8 BLEU points compared to the original model (25.6 to 29.4) without further training. Similarly, \citeA{zoph-etal-2016-transfer} proposed a transfer learning method for low-resource language pairs.

Among previous works, the research presented in \cite{wang-etal-2017-sentence} is very similar to our work in terms of the intuition behind the selection methodology. Their method selects in-domain data based on similarity scores computed over embeddings drawn from an \gls{nmt} system trained on in and out of domain data. This approach has several limitations related to the fact that it relies on a particular \gls{nmt} system that needs to be trained on both in- and out-of-domain data. That is, the complexity of their approach makes it rather difficult to employ in practice. Furthermore, as it relies on the embeddings of a particular \gls{nmt}, it implies that a (language-specific) \gls{nmt} system needs to be available or trained which may add computational or economic overhead.

\citeA{axelrod-etal-2011-domain} proposed a \gls{da} approach using data selection which is a common baseline for many more recent works. This is mainly because their work for the first time introduced the concept of \gls{da} in \gls{mt}. They selected and ranked sentences with three cross-entropy based methods for the task of \gls{smt}. They also showed that all three methods presented in their paper outperformed the general-domain model. In the same direction, \citeA{Chen2016BilingualMF} presented another data selection technique employing Semi-Supervised Convolutional Neural Networks based on bitokens (Bi-SSCNN). The method they proposed only requires a small amount of in-domain data to train the selection model. Suggested methods were tested on two translation tasks (Chinese-to-English and Arabic-to-English) and showed that the Bi-SSCNN is more efficient than other approaches in terms of averting noisy sentence pairs. We compare our approach to the aforementioned models (among others). In Section~\ref{CLIN:sec:experiments}, we outline those models and present further details and comparisons. 

With respect to data selection studies, \citeA{van-der-wees-etal-2017-dynamic} also investigated a method, called dynamic data selection, to discern whether it is feasible to improve \gls{nmt} performance. Their method sifts through all training data between training epochs subsequently and reduced the training data size by selecting sentence pairs most relevant to the translation task. By doing so, unlike fixed training data, the training becomes a gradual fine-tuning process, which iterates over different training subsets made. \citeA{chu-etal-2017-empirical} proposed a novel \gls{da} method, called mixed fine-tuning, incorporating fine-tuning into multi-domain \cite{sennrich-etal-2016-controlling,Kobus_2017} for \gls{nmt}. In the context of the corpora they experimented with, their fine-tuning method on a mix of in-domain and out-of-domain solves the problem of overfitting. 

It is also worth mentioning the work of \citeA{aharoni-goldberg-2020-unsupervised} which proposes two methods for data selection based on unsupervised language models. The first one computes the centroid of the in-domain data and selects the samples that are nearest to the centroid, according to their cosine similarity. The second method is based on a binary classifier trained on in-domain sentences and a random negative sample (i.e., general-domain) of sentences. They evaluate the proposed approach on 5 domains and compare it to the work of \citeA{moore-lewis-2010-intelligent}. Among other findings, they show that data selection is a valuable technique. Furthermore, they show that in multi-domain scenarios, more data does not necessarily lead to better performance, emphasizing the importance of data quality over quantity.

\section{Data selection method}
\label{CLIN:sec:data_selection}
%We present a method to select and rank in-domain subsets of an out-of-domain corpus, in an attempt to boost in-domain translation performance. 
Our method ranks sentences in a general-domain (or out-of-domain) dataset according to their similarity with an in-domain dataset. This in-domain dataset is monolingual; if a parallel corpus is provided only the source or target side is to be used. Once the sentences are ranked, we can then extract the top K sentences with the highest score, i.e., ranked the highest and use those for training a new \gls{mt} system. According to our architecture, initially the input data, both in-domain and out-of-domain is first converted into embedding vectors. Our method then computes the similarity between these vectors and uses the similarity score for ranking and consecutively, for selection. The embedding space we use, Sentence BERT~\cite{reimers-gurevych-2019-sentence}, is of high dimensionality. As such, these vectors become quite large to be processed effectively. To address this, we apply Principal Component Analysis (PCA) to reduce the dimensionality, enabling more efficient semantic search, ranking, and selection.
 
\subsection{Sentence embedding and dimensionality reduction}
\label{CLIN:sec:dimensionality_reduction}
%Word representation is a rich resource for gaining information for downstream tasks such as classification, entailment, translation, etc. In the field of NLP, including MT, models are highly dependent on the quality and efficacy of input representations \cite{10.1145/3434237}. While the myriad of categorical and fixed methods such as Bag-of-Words (BOW), Continuous BOW model, Skip-Gram Model \cite{mikolov2013efficient}, FastText \cite{bojanowski2017enriching} have been employed in this regard, nowadays researchers tend to benefit more from unsupervised contextual word representations architectures and in particular transformer-based language models \cite{vaswani2017attention}. The main reasons are that (i) these models keep the full context of the input and (ii) they reduce the computational time for most NLP tasks. These directly align with our research objectives: to be able to select semantically similar in-domain sentences and to reduce MT training time as well as similarity computational time. Hence, efficient sentence embedding plays a major role in our work.

Word representation is a rich resource for gaining information for downstream tasks such as classification, entailment, translation, etc. In the field of \gls{nlp}, including \gls{mt}, models are highly dependent on the quality and efficacy of input representations \cite{10.1145/3434237}. Models greatly depend on how the input is represented because it influences their ability to capture semantic meaning and context effectively, which directly impacts their performance in tasks such as \gls{mt} and classification. While the myriad of categorical and fixed methods such as Bag-of-Words (BOW), Continuous BOW model, Skip-Gram Model \cite{mikolov2013efficient}, FastText \cite{bojanowski2017enriching} have been employed in this regard, nowadays researchers tend to benefit more from unsupervised contextual word representation architectures and, in particular, transformer-based language models \cite{vaswani2017attention}. The main reasons are that (i) these models keep the full context of the input and (ii) they reduce the computational time for most \gls{nlp} tasks. These directly align with our research objectives: to be able to select semantically similar in-domain sentences and to reduce \gls{mt} training time as well as similarity computational time. For example, these models provide representations with consistent dimensions, making comparisons straightforward, and reorient words or sentences within the context provided by the data, such as the same domain. Hence, efficient sentence representations play a major role in our work.

%There exist several transformer-based language models, such as, BERT \cite{devlin-etal-2019-bert} and RoBERTa \cite{liu2019roberta} that have set a state-of-the-art baseline \cite{cer-etal-2017-semeval}. However, for tasks like Semantic Textual Similarity (STS), Sentence-BERT (SBERT) \cite{reimers-gurevych-2019-sentence} recently showed a better performance. It is a modification of the pre-trained BERT network that employs Siamese \cite{10.5555/2987189.2987282} and triplet network structures \cite{Schroff_2015}, capable of capturing meaning-related relationships which allow assessing the degree to which two sentences are semantically similar at reduced computational costs. We note that SBERT is only used in our research for embedding words and not as input for MT.

There exist several transformer-based language models, such as BERT \cite{devlin-etal-2019-bert} and RoBERTa \cite{liu2019roberta}, that have set a state-of-the-art baseline \cite{cer-etal-2017-semeval}. These models generate representations at the word or token level. However, for tasks like Semantic Textual Similarity (STS), Sentence-BERT (SBERT) \cite{reimers-gurevych-2019-sentence} recently showed a better performance for sentence-level representations. SBERT is a modification of the pre-trained BERT network that employs Siamese \cite{10.5555/2987189.2987282} and triplet network structures \cite{Schroff_2015}, capable of capturing meaning-related relationships, which allows assessing the degree to which two sentences are semantically similar at reduced computational costs. We used sentence-level information in our research to focus on contextual relevance. It is important to note that SBERT is only used in our research for embedding sentences, not as input for \gls{mt}.

By default, the SBERT base model embeddings output has 768 dimensions. Sentences encoded using 768-dimensional vectors require a substantial amount of memory to be stored. For example, an out-of-domain corpus containing 31 million sentences generates a $31M\times768$ embedding matrix which is computationally expensive. The size of these large vectors does not only require substantial physical memory but also considerably increases the time for computing the semantic similarity (i.e., the STS task), where we need to cautiously mitigate the cost of computing semantic search between in-domain and out-of-domain sentences. To speed up the search process, embedding vectors should be loaded into GPU memory to leverage parallel processing. However, large embeddings like ours can quickly max out GPU memory, leading to issues like memory swapping or batching. These workarounds increase processing time and ultimately negate the speed benefits of using GPUs, creating a major challenge for large-scale semantic search.

In order to mitigate these issues, we decided to work with smaller sized vectors. However, reducing the vector dimensions must be done in a conscious way, in order not to lose important information. To do so, we employ PCA~\cite{Jolliffe2011} as a pooling method into the last embedding neural layers. We select the 32 principal components as our output features. That is, PCA acts as the final layer of our selection network which allows us to reduce the 768-dimensional vectors to 32-dimensional ones. In general, PCA components are equivalent to the output features and are easy to append to any SBERT pre-trained model. 

%To employ the PCA idea in SBERT, we only selected and shuffled 500 K sentences of experimental datasets to train a PCA model. 
We selected and shuffled 500K sentences from the experimental datasets to train the PCA model, assuming they were representative of the entire corpus while keeping the process computationally efficient.
%A pre-trained model called 'stsb-xlm-r-multilingual'\footnote{\url{https://huggingface.co/sentence-transformers/stsb-xlm-r-multilingual}} (trained on over 50 languages), has been selected and subsequently 32 components appended to that. Finally, the reduced model is saved to use for word embedding. 
A pre-trained model called 'stsb-xlm-r-multilingual'\footnote{\url{https://huggingface.co/sentence-transformers/stsb-xlm-r-multilingual}} (trained on over 50 languages) was selected, and its embeddings were reduced to 32 dimensions using PCA. This dimensionality reduction retains the most important features of the embeddings while significantly improving computational efficiency. Finally, the reduced model was saved for use in sentence embedding tasks.

\subsection{Semantic search and ranking in-domain data}
\label{CLIN:sec:semanticssearch}
Once the input data was reduced and embedded, we employed semantic search to identify general-domain sentences that are similar to in-domain data. This idea is mainly inspired by \cite{moore-lewis-2010-intelligent,axelrod-etal-2011-domain,duh-etal-2013-adaptation,wang-etal-2017-sentence}. %In particular, we assume in-domain and out-of-domain sentences as search queries and document entries respectively. In our assumption, queries are responsible to find the most relevant embedding vectors from entries. 
Specifically, we treat in-domain sentences as search queries and out-of-domain sentences as document entries in the semantic search process. In this context, the queries are used to retrieve the most relevant embedding vectors from the entries, effectively matching in-domain sentences to their closest counterparts in the out-of-domain corpus.

The proximity search can be fulfilled by various distance similarity measures methods, such as cosine similarity, Manhattan distance, Euclidean distance, etc. In our research, we used cosine similarity\footnote{We chose cosine similarity because, unlike Euclidean or Manhattan distance, it normalizes vectors and compares only their direction. Since semantic meaning in embeddings is represented by direction rather than magnitude, cosine similarity provides a more reliable measure of contextual similarity.}. However, we could not apply such an exhaustive search (each in-domain sentence compared to each out-of-domain sentence) on our datasets using CPU as was expensive, specifically in terms of computational time. To resolve this issue, we used a GPU implementation of cosine similarity in the PyTorch library. Once the cosine similarity scores were computed over embedding vectors, the torch.topk\footnote{\url{https://pytorch.org/docs/stable/generated/torch.topk.html}} is called. By calling this function the n largest / most similar elements, as well as indices of the given in-domain or out-of-domain embedded tensors, are returned. Below we provide a detailed explanation of this process.

Let $S$, $E$, $q$ and $d$ denote in-domain corpus, out-of-domain corpus, a vectorized search query and an embedded document entry, respectively, where $q\in S$ and $d\in E$. Let $k$ and $l$ be the number of sentences in the given corpora. That is, $S=\left \{q_1,q_2,\text{...},q_k\right \}$, $E=\left \{d_1,d_2,\text{...},d_l \right \}$ and $k\ll l$. According to these definitions, the cosine similarity is defined by Equation \ref{CLIN:eq1} in our data selection method, where 32 is the number of dimensions. 

\begin{equation}
\centering
\label{CLIN:eq1}
cos(\vec{q}, \vec{d}) = \frac{\vec{q}\cdot \vec{d}}{|\vec{q}||\vec{d}|} = \frac{\vec{q}}{|\vec{q}|}\cdot \frac{\vec{d}}{|\vec{d}|} = \frac{\sum_{i=1}^{32}q_{i}d_{i}}{\sqrt{\sum_{i=1}^{32}{q_{i}}^{2}}\sqrt{\sum_{i=1}^{32}{d_{i}}^{2}}}
\end{equation}

Based on the similarity measurement defined in Equation~\ref{CLIN:eq1}, we rank our sentences and pick the top $n$ ($n = 6$ in our experiments)\footnote{Set empirically; tests with $n \in \{1,2,3,4,5,6\}$ showed that translation quality and similarity to in-domain data converged at six.} out-of-domain sentences, which are sorted in descending order, to build pseudo in-domain sub-corpora. Considering that sentences are chosen from an out-of-domain corpus and are distinct bitexts, no further operation is required before feeding those into an \gls{nmt} system.

In summary, our data selection method has four major steps:
\begin{itemize}
    \item Step 1, the input data is converted into vectors using the embedding unit.
    \item Step 2, the vectors' dimensions are reduced to the lower dimensions.
    \item Step 3, we compute the similarity scores between each in-domain and out-of-domain vector.
    \item Step 4, vectors pairs are ranked according to their similarity score achieved from step 3.
\end{itemize}

\section{Experiments}
\label{CLIN:sec:experiments}
To test the quality of our data selection performance to generate parallel in-domain data, we conducted experiments with English-French data and compared our results to different systems' results. These systems are divided mainly into two categories: %(i) for the first category, we trained some models for specific purposes, such as having an in-domain NMT model that only uses the original / given in-domain data for training. i.e., systems that are not trained with the selected data generated by our data selection method. We use them to analyze how well our data selection algorithm worked in terms of helping the model to reach the maximum possible translation performance, providing a comparison point. It is noteworthy that the systems in the first category are not considered baselines. 
%(ii) For the second category, we chose previous researchers' MT systems that were trained using their selected data. These systems after being trained on the selected in-domain data are usually re-trained on the original / given in-domain data to increase the translation performance. Although most systems in this category used DA / re-training in their work, we considered them as actual baselines. This is, first, because there is not much previous work that only uses in-domain parallel data to train MT systems. Second, we aim to evaluate the quality of our generated data to see if it helps the models to improve the translation quality without re-training, i.e., as a stand-alone corpus.
\begin{enumerate}
    \item \textbf{Category 1:} Models trained exclusively on the original in-domain data, without incorporating the data selected by our method. These systems were used to analyze how effectively our data selection algorithm improves translation performance compared to models trained on only the given in-domain data. It is important to note that these systems are not considered baselines but serve as a comparison point to assess the added value of our approach.
    
    \item \textbf{Category 2:} For the second category, we chose previous researchers' \gls{mt} systems that were trained using their selected data. These systems after being trained on the selected in-domain data are usually re-trained on the original / given in-domain data to increase the translation performance. Although most systems in this category used \gls{da} / re-training in their work, we considered them as actual baselines. This is, first, because there is not much previous work that only uses in-domain parallel data to train \gls{mt} systems; and second, because we aim to evaluate the quality of our generated data to see if it helps the models to improve the translation quality without re-training, i.e., as a stand-alone corpus.\end{enumerate}

\subsection{Data}
\label{CLIN:sec:data}
\paragraph{In- and out-of-domain datasets.}
The data we experimented with is a collection of TED talks, referred to as the IWSLT\footnote{International Workshop on Spoken Language Translation} 2014 corpus \cite{Cettolo2015ReportOT}. To validate our models during training time and test the models' performance, we used one development set (dev2010) and two test sets (test2010 and test2011), respectively. In addition to an in-domain corpus, we combined a collection of WMT corpora\footnote{\url{http://statmt.org/wmt15/translation-task.html}} including Common Crawl, Europarlv7, News Commentary v10 and United Nations (UN) \cite{Koehn2005EuroparlAP,tiedemann-2012-parallel} to create a large out-of-domain corpus. IWSLT 2014 and WMT are commonly used in the context of \gls{da} as an in-domain dataset \cite{axelrod-etal-2011-domain,Luong2015StanfordNM,Chen2016BilingualMF,wang-etal-2017-sentence}, which allows for better applicability of our experiments. Data statistics are shown in Table \ref{CLIN:tab:corpora}.

\begin{table}[ht]
\setlength\tabcolsep{4pt}
\renewcommand{\arraystretch}{1.40}
\centering
\begin{adjustbox}{width=\textwidth,center}
\begin{tabular}{l|l|l}
\hline
\hline
\textbf{EN-FR}           & \textbf{Name}                & \textbf{Sentences} \\ \hline\hline
TED training (in-domain) & \multirow{4}{*}{IWSLT 2014}  & 179K                            \\ %\cline{1-1} \cline{3-4} 
TED dev2010              &                              & 887                                 \\ %\cline{1-1} \cline{3-4} 
TED test2010             &                              & 1664                              \\ %\cline{1-1} \cline{3-4} 
TED test2011             &                              & 818                                 \\ \hdashline[1pt/1pt]
\multirow{4}{*}{WMT training (out-of-domain)} & Common Crawl & 3.25M  \\ %\cline{2-3}
                         & Europarl v7                  & 2M                                \\ %\cline{2-3}
                         & News Commentary v 10         & 200K                               \\ %\cline{2-3}
                         & United Nations (UN)          & 25.8M                              \\ \hline \hline
\end{tabular}
\end{adjustbox}
\caption{\textbf{Overview of EN-FR datasets used for training and evaluation.} The in-domain data (IWSLT 2014) includes TED talks for training, dev2010, test2010, and test2011. The out-of-domain data (WMT) consists of Common Crawl, Europarl v7, News Commentary v10, and UN datasets. The table presents the number of sentences for each dataset and split. The total number of out-of-domain samples is approximately 31 million.}

\label{CLIN:tab:corpora}
\end{table}

\paragraph{Selected data.} We used our method (see Section~\ref{CLIN:sec:semanticssearch}) and the monolingual in-domain dataset, i.e., IWSLT, to extract the most similar subsets of in-domain data from out-of-domain data, i.e., WMT. Since the data we extract is not specifically compiled as (authentic) in-domain data, but rather automatically generated based on similarities, it is referred to as \emph{pseudo} in-domain data~\cite{zhang-xiong-2018-sentence}. In this chapter, to distinguish between authentic and pseudo in-domain data, we refer to the latter as selected in-domain data as this reflects the origin of the data. 

\begin{figure}[ht]
	\centering 
	\includegraphics[width=4.2in]{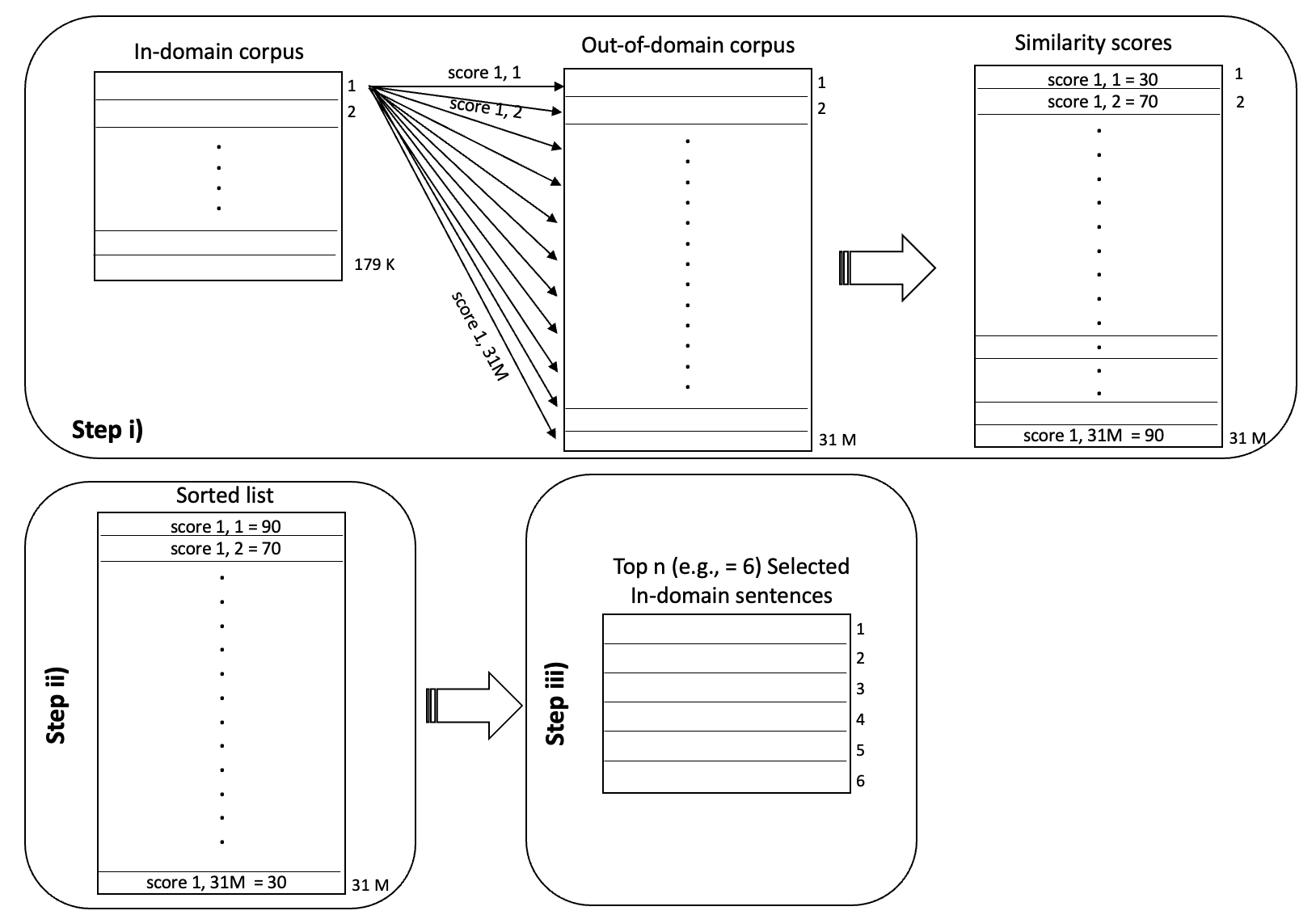} 
	\caption{An iteration of selecting in-domain data} 
	\label{CLIN:fig:dataselection}
\end{figure}

For our experiments, we created six (sub-)corpora based on the sentence ranks determined by the data selection method. To do so, (i) we choose one in-domain sentence and compute its similarity score with every single out-of-domain data point, i.e., one-to-many. So, for each in-domain sentence, we obtain a score list with the size of out-of-domain data, i.e., 31M. (ii) Afterward, the out-of-domain sentences are sorted in descending order according to the similarity scores achieved in step (i); (iii) We only select first n ($n=6$) out-of-domain sentences from the list generated in step (ii). We repeat all aforementioned steps for every single in-domain entry, i.e., 179K. This procedure outputs a $179k\times 6$ matrix according to our input data. Figure \ref{CLIN:fig:dataselection} shows one iteration of data selection.

In practice, the selected sentences (top1–top6) come from whichever parts of the out-of-domain corpora happen to be most similar to the in-domain sentences in SBERT embedding space. This similarity is semantic—primarily topical and contextual.

\paragraph{Note on data mixing.}
All top-$k$ sub-corpora (top1–top6) contain only \emph{selected in-domain} data
(i.e., automatically selected sentences). Mixing is done only among these selected sub-corpora (e.g., top1+top2), and never with the original IWSLT in-domain corpus.

Table \ref{CLIN:tab:DSoutput} shows an example of possible outputs for the proposed data selection algorithm given a monolingual in-domain query, where generated sentences were sorted from the highest score (top1) to the lowest one (top6).

\begin{table*}[ht]
\centering
\Huge
\begin{adjustbox}{width=\textwidth,center}
\renewcommand{\arraystretch}{1.8}
\begin{tabular}{rl|c|l|}
\cline{3-4}
 &
   &
  \multicolumn{2}{l|}{Score (/100)} \\ \hline
\multicolumn{1}{|r}{Monolingual in-domain ($q_i$):} &
  It can be a very complicated thing, the ocean. &
  \multicolumn{2}{c|}{-} \\ \hline
\multicolumn{1}{|c}{Top1–parallel in-domain ($top_{i0}$):} &
  \begin{tabular}[c]{@{}l@{}}EN: Ocean affairs are sensitive and complex.\\ FR: Les affaires maritimes sont délicates et complexes.\end{tabular} &
  \multicolumn{2}{c|}{90.10} \\ \hline
\multicolumn{1}{|r}{Top2–parallel in-domain ($top_{i1}$):} &
  \begin{tabular}[c]{@{}l@{}}EN: This is a dangerous position to be in if the sea is running high.\\ FR: Ainsi, le capitaine peut prendre les effets du capitaine du navire pris, le chirurgien ...\end{tabular} &
  \multicolumn{2}{c|}{86.80} \\ \hline
\multicolumn{1}{|r}{Top3–parallel in-domain ($top_{i2}$):} &
  \begin{tabular}[c]{@{}l@{}}EN: Rip currents and undertow are common, dangerous conditions along ocean beaches.\\ FR: Déchirez les courants et les baïnes sont des conditions communes et dangereuses le long des plages d'océan.\end{tabular} &
  \multicolumn{2}{c|}{86.60} \\ \hline
\multicolumn{1}{|r}{Top4–parallel in-domain ($top_{i3}$):} &
  \begin{tabular}[c]{@{}l@{}}EN: Moving with the waves can be dangerous.\\ FR: Il est dangereux de progresser avec la vague.\end{tabular} &
  \multicolumn{2}{c|}{86.13} \\ \hline
\multicolumn{1}{|r}{Top5–parallel in-domain ($top_{i4}$):} &
  \begin{tabular}[c]{@{}l@{}}EN: Obstacles in the water are particularly dangerous when coupled with currents.\\ FR: Les obstacles dans l’eau sont avant tout dangereux par rapport au courant.\end{tabular} &
  \multicolumn{2}{c|}{85.96} \\ \hline
\multicolumn{1}{|r}{Top6–parallel in-domain ($top_{i5}$):} &
  \begin{tabular}[c]{@{}l@{}}EN: This problem affects not only small islands, but also large islands and countries with extensive coastlines.\\ FR: Ce problème concerne non seulement les petites îles, mais aussi les grandes îles ....\end{tabular} &
  \multicolumn{2}{c|}{85.76} \\ \hline
\end{tabular}
\end{adjustbox}
\caption{\textbf{Example of data selection output}: Scores (/100) for different data subsets selected using monolingual.}
\label{CLIN:tab:DSoutput}
\end{table*}

\begin{figure}[ht]
	\centering 
	\includegraphics[width=\textwidth]{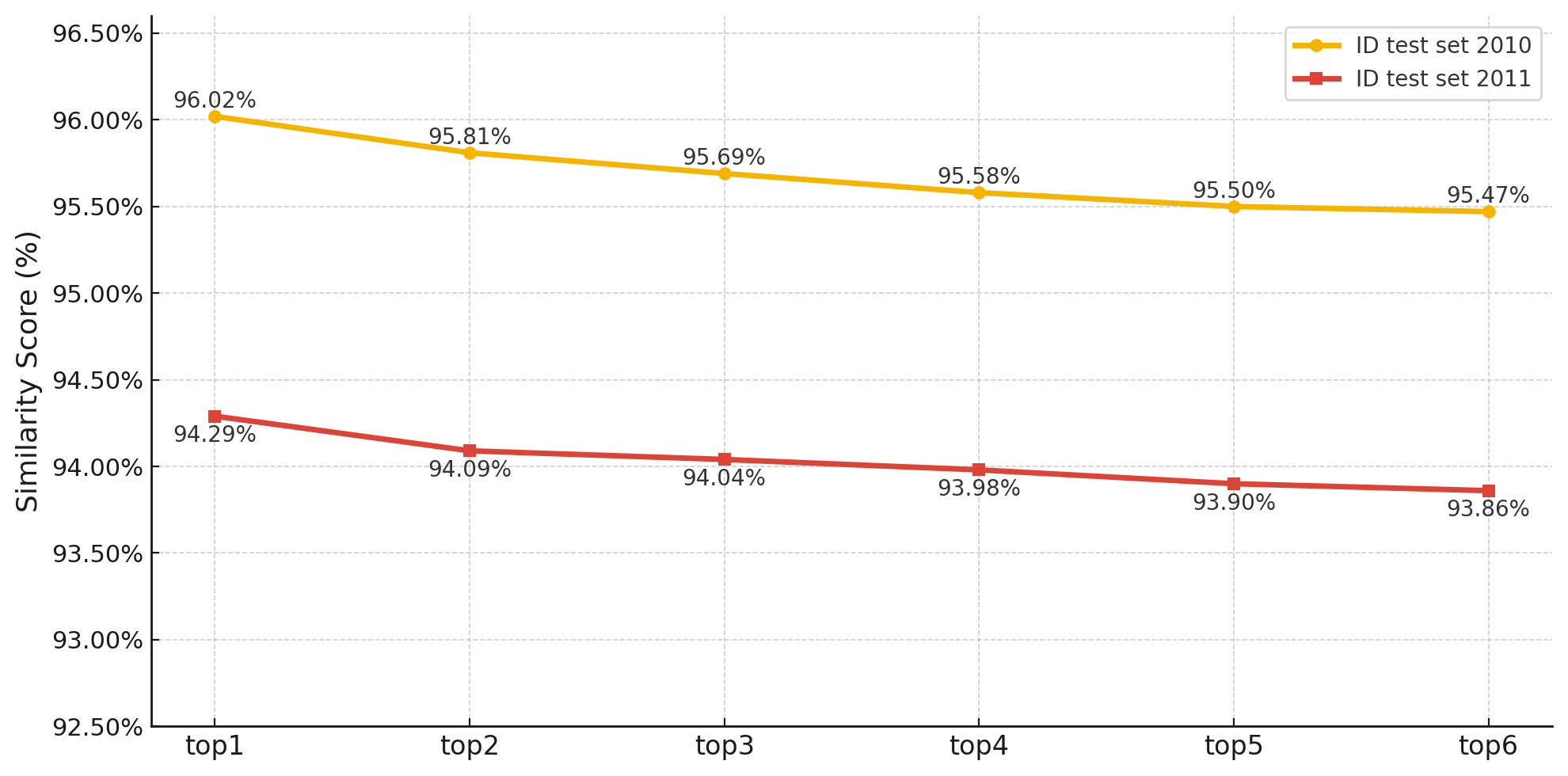} 
	\caption{The difference of selected sub-corpora to in-domain test sets} 
	\label{CLIN:fig:closeness}
\end{figure}

To show the effectiveness of our semantic search and ranking idea, we found the centroids of selected sub-corpora, then compared them to the in-domain test sets' centroids. The centroid is a multidimensional vector, calculated as the average of all other vectors. That is, it is a vector around which other vectors are distributed. Figure~\ref{CLIN:fig:closeness} depicts a gradual drop of similarity score from top1 to top6 for both test sets. To increase the performance of in-domain translation in terms of diversity richness, each sub-corpus is combined with all preceding bitext data (like a stack). For instance, top6 comprises top5, top4, top3, top2, top1; top5 holds top4, top3, top2, top1 and so forth. In that way, that last corpus (i.e., top6) encompasses all in-domain sentences.

\subsection{\gls{nmt} system description}
We used the OpenNMT-py\footnote{\url{https://opennmt.net/OpenNMT-py/}} framework \cite{klein-etal-2017-opennmt} for training our \gls{nmt} models. We trained transformer models \cite{vaswani2017attention} for a maximum of 200K steps; intermediate models were saved and validated every 1000 steps until reached convergence unless being stuck in the early stopping condition (10 validation steps with no improvements). The parameter setup stated in Table~\ref{CLIN:tbl:hyperparameters} was used. 

%We performed a large set of preliminary experiments to determine these values. These experiments span beyond the scope of this work and as such are not further discussed, except for the batch size. The batch size and the learning rate have a direct connection to how the network learns, especially in different conditions of data sparsity / availability. As such, we conducted a thorough investigation of both and we found that using batch size 512 and the default learning rate in the OpenNMT was particularly impactful. Our results are summarised in Section~\ref{CLIN:sec:batch_size}.

Hyperparameters were determined through preliminary experiments based on convergence speed, validation performance, early stopping behavior, and generalization ability. While a full discussion is beyond the scope of this work, batch size and learning rate were particularly beneficial. We found that a batch size of 512 and the default OpenNMT learning rate yielded the best results. See Section~\ref{CLIN:sec:batch_size} for details.

\begin{table}[h]
    \centering
    % \small % Reduce font size
    \setlength\tabcolsep{7pt} % Adjust column spacing
    \begin{adjustbox}{width=\textwidth,center}
    \begin{tabular}{p{0.3\textwidth}r||p{0.3\textwidth}r}
        \hline\hline
        word emb. dim. & 512 & max seq. len. & 150 \\
        num. transf. layers & 6 & learn. opt. & Adam \\
        transf.-ff size & 2048 & opt. lr & 2 \\
        num. heads & 8 & opt. b1 and b2 & 0.9, 0.998 \\
        batch size & 512 & beam size & 6 \\
        % batch type & tokens & & \\
        \hline\hline
    \end{tabular}
    \end{adjustbox}
    \caption{\textbf{Hyperparameters used for training our \gls{nmt} models.} Abbreviations: emb. dim. (embedding dimension), transf. layers (transformer layers), transf.-ff (transformer feed-forward), max seq. len. (maximum sequence length), learn. opt. (learning optimizer), opt. lr (optimizer learning rate), opt. b1 and b2 (optimizer beta1 and beta2).}
    \label{CLIN:tbl:hyperparameters} 
\end{table}

To run all \gls{nmt} systems effectively, we also set other hyperparameters, as suggested by the OpenNMT-py community to simulate Google's default setup \cite{vaswani2017attention}. The \gls{nmt} training was distributed over three NVIDIA Tesla V100 GPUs. We encoded all data, including rare and unknown words as a sequence of subword units using Byte Pair Encoding (BPE) \cite{sennrich-etal-2016-neural}. We built our vocabularies for both source and target languages separately. By doing so, our systems are not only capable of translating out-of-vocabulary tokens, but also rare in-vocabulary ones. The number of merge operations for BPE is 50,000 for all sub-corpora, i.e., For each selected sub-corpora, a separate BPE model was created. Vocabulary sizes of selected in-domain data are shown in Table \ref{CLIN:tbl:vocab}.

\begin{table}[ht]
\centering
\begin{adjustbox}{width=0.80\textwidth,center}
\renewcommand{\arraystretch}{1.3}
\begin{tabular}{l|rr}
\hline
\hline
\multicolumn{1}{c|}{\multirow{2}{*}{\textbf{Selected In-domain Datasets}}} & \multicolumn{2}{c}{\textbf{Vocabulary with BPE}} \\ \cdashline{2-3}[1pt/1pt]%\cline{2-3} 
\multicolumn{1}{c|}{} & \textbf{Source} & \textbf{Target} \\ \hline\hline
Top1 & 48,956 & 49,281 \\ 
Top2 + top1 & 49,896 & 50,055 \\
Top3 + top2 + ... & 50,299 & 50,391 \\ 
Top4 + top3 + ... & 50,596 & 50,568 \\ 
Top5 + top4 + ... & 50,759 & 50,720 \\ 
Top6 + top5 + ... & 50,874 & 50,894 \\ \hline \hline
\end{tabular}
\end{adjustbox}
\caption{\textbf{Vocabulary sizes for the source (EN) and target (FR)} after applying BPE across various combined in-domain datasets.}
\label{CLIN:tbl:vocab}
\end{table}

\subsection{Compared systems}
To show the effectiveness of the proposed data selection method in terms of the quality of generating parallel in-domain data, as mentioned earlier, we compared our results with two disparate categories of \gls{nmt} systems as follows. First, we used the available in-domain and out-of-domain data as well as their mixture to establish the following systems: (i) S1:ID---\gls{nmt} trained on in-domain only; (ii) S2:OOD---\gls{nmt} trained on out-of-domain data; and (iii) S3:ID+OOD---\gls{nmt} trained on the combination of in- and out-of-domain data. It is noteworthy that here, we employed a bitext in-domain data to demonstrate our selection method's productivity, however, the proposed method uses a monolingual in-domain corpus in a real use case.

System S1, S2 and S3 are not intended as baselines, but points of comparisons for different purposes: (i) system S1 resembles the best possible translation quality according to the given in-domain data; (ii) system S2 is trained on a large corpus without including domain-relevant data, thus it is a generic-domain \gls{mt} system; and (iii) system S3 is trained on a mixture of a large generic-domain corpus (the same as for S2) and domain-relevant data and aims to test the impact of the in-domain data on the translation performance, resembling a domain-adapted model. 
Second, we compared our \gls{nmt} systems (based on the proposed data selection method) to four reliable previous \gls{nmt} \gls{da} methods, which we refer to as baselines. These are B4:Luong \cite{Luong2015StanfordNM}, B5:Axelrod \cite{axelrod-etal-2011-domain}, B6:Chen \cite{Chen2016BilingualMF}, B7:Wang \cite{wang-etal-2017-sentence} as mentioned in Section \ref{CLIN:sec:related_work}. While \citeA{wang-etal-2017-sentence} proposed three different models, we only refer to the best one. It is noteworthy that the numbering of baselines starts from B4 to align with system conventions (i.e., S1–S3); That is, B1–B3 do not exist.

\subsection{Results and analysis}
\label{CLIN:sec:results_analayses}

The performance of our \gls{mt} systems is reported on two test sets using case insensitive BLEU \cite{10.3115/1073083.1073135}, TER \cite{snover-etal-2006-study} and chrF2 \cite{popovic-2015-chrf} metrics, as implemented in the sacreBLEU framework\cite{post-2018-call}. We also analyzed the results for any statistically significant differences (see Section \ref{CLIN:subsec:ss}). The vocabulary was built by employing the IWSLT train set and each selected in-domain corpus. Note that according to Table \ref{CLIN:tab:indomain_data_selection} data, S1 achieved the highest BLEU score among the compared systems on both test sets. We considered it as a best-case scenario (i.e., target performance). As such, we aim to generate parallel in-domain sentences that yield results as good as S1. 

\begin{table}[ht]
\renewcommand{\arraystretch}{1.3}
\centering
\begin{adjustbox}{width=\textwidth,center}
\begin{tabular}{l|rrrrrrr}
\hline
\hline
\multirow{2}{*}{\textbf{Systems}} &
  \multirow{2}{*}{\begin{tabular}[c]{@{}l@{}}\textbf{Number of} \\ \textbf{Sentences}\end{tabular}} &
  \multicolumn{3}{c}{\textbf{\gls{nmt}- Test Set 2010}} &
  \multicolumn{3}{c}{\textbf{\gls{nmt}- Test Set 2011}} \\ \cdashline{3-8}[1pt/1pt]%\cline{3-8} 
 &
   &
  \multicolumn{1}{l}{\textbf{BLEU$\uparrow$}} &
  \multicolumn{1}{l}{\textbf{TER$\downarrow$}} &
  \multicolumn{1}{l}{\textbf{CHRF2$\uparrow$}}  &
  \multicolumn{1}{l}{\textbf{BLEU$\uparrow$}} &
  \multicolumn{1}{l}{\textbf{TER$\downarrow$}} &
  \multicolumn{1}{l}{\textbf{CHRF2$\uparrow$}} \\ \hline \hline
S1:ID         & 179K      & 31.9           & 56.6           & 57.0        & 38.3          & 49.7        & 61.0          \\
S2:OOD        & 31.0M     & 25.8           & 66.1           & 53.0        & 30.7          & 59.3        & 47.0          \\
S3:ID+OOD     & 31.1M     & 26.0           & 62.9           & 54.0        & 30.9          & 56.8        & 58.0          \\
\hdashline[1pt/1pt]
B4:Luong      & 17.9M     & 32.2           & N/A            & N/A         & 35.0          & N/A         & N/A           \\
B5:Axelrod    & 9.0M      & 32.2           & N/A            & N/A         & 35.5          & N/A         & N/A           \\
B6:Chen       & 7.3M      & 30.3           & N/A            & N/A         & 33.8          & N/A         & N/A           \\
B7:Wang       & 3.7-7.3M  & 32.8           & N/A            & N/A         & 36.5          & N/A         & N/A           \\
\hdashline[1pt/1pt]
Top1          & 179K      & 21.8           & 69.8           & 50.0          & 25.6          & 64.0          & 53.0            \\
Top2+top1 & 358K      & 26.7           & 63.4           & 54.0          & 31.3          & 57.1        & 57.0            \\
Top3+top2+... & 537K      & 29.1           & 60.4           & 56.0          & 34.3          & 53.9        & 60.0            \\
Top4+top3+... & 716K      & 30.7           & 59.5           & 57.0          & 35.6          & 52.6        & 61.0            \\
Top5+top4+... & 895K      & 30.9           & 59.1           & 57.0          & \textbf{36.7} & \textbf{51.5} & \textbf{62.0}  \\
Top6+top5+... & 1.0M      & \textbf{31.3}  & \textbf{58.3}  & \textbf{58.0} & 36.5          & 50.9        & 62.0            \\
\hline \hline
\end{tabular}
\end{adjustbox}
\caption{\textbf{Evaluation scores for EN-FR \gls{nmt} systems trained on different data domains:} in-domain (S1, B4, B5, B6, B7 and Top1 .. Top6+top5+...), out-of-domain (S2), and a mixture of both (S3). Results are reported on \gls{nmt}-Test Set 2010 and \gls{nmt}-Test Set 2011, including BLEU, TER, and CHRF2 scores.}
\label{CLIN:tab:indomain_data_selection}
%\vspace{-3mm}%Put here to reduce too much white space after your table
\end{table}

Even though system S2 employed an enormous corpus with almost 31M sentences, it could not perform well for in-domain translation. System S3 had a minor improvement (0.2 BLEU points) after mixing OOD with ID, yet less than S1 performance. These results suggest that increasing the amount of training data, even with systematic inclusion of in-domain translation, is not always sufficient for improving in-domain translation and may even lead to under-fitted model. Moreover, system S3 exhibited bias due to an over-reliance on out-of-domain data, causing it to prioritize translation patterns––such as word usage, expressions, and sentence structures––from OOD data, which were less suitable for in-domain translation.

%The proposed sub-corpora (top1 to top6), whose mixture led to some data removal, were employed to train the NMT systems. In this regard, we started from top1, then continued with the mixture of other sub-corpora. In the beginning, there was a sharp improvement from top1 to top2 (+4.9 BLEU points for test set 2010 and +5.7 BLEU points for test set 2011). This growth trend continued until its saturation point in top5 and top6, where the performance started degrading. Given this point, we only selected six sub-corpora to achieve the maximum translation performance. Hence, top5 and top6 obtained the highest BLEU scores among the selected sub-corpora, 36.7 for test set 2011 and 31.3 for test set 2010, respectively.

The proposed sub-corpora (top1 to top6), whose mixture resulted in some data removal, were used to train the \gls{nmt} systems. We began by training the first system using top1 (referred to as system top1). Next, a mixture of top1 and top2 was used to train the second system (system top2), and this process continued by progressively adding more sub-corpora to train subsequent systems (top3, top4, and so on). Initially, performance improved significantly, with BLEU score gains of +4.9 points on the 2010 test set and +5.7 points on the 2011 test set from top1 to top2. This growth trend persisted until it reached a saturation point at top5 and top6, after which performance either plateaued or slightly degraded. Based on these results, we selected six sub-corpora to maximize translation performance. Among the selected systems, top5 and top6 achieved the highest BLEU scores, with 36.7 on the 2011 test set and 31.3 on the 2010 test set, respectively.

Alongside BLEU, we also evaluated our models with TER and chrF2. According to TER, system top1 achieved the highest score among all systems trained on mixed sub-corpora. This would imply that this system would require the most post-editing effort. The TER score for top2 mixed with top1 dropped by 6.4 and 6.9 for test sets 2010 and 2011, respectively. This improvement (TER decrease) per each mixing operation continued until the system top6+top5+... but with a smaller amount of drop. According to the chrF2 metric, there was an increasing trend from top1 to top6+top5+... for test set 2010, where the last sub-corpus obtained 58.0 scores, while for test set 2011, two last systems (top5+top4+... and top6+top5+...) achieved the same chrF2 score (62.0). 

The three metrics have consistently improving trends, which supports the observation that, while more data implies an increase in quality, there is a point after which no (significant) improvements are possible.\footnote{See Section~\ref{CLIN:subsec:ss} for a detailed statistical significance interpretation.}

Although systems B4, B5, B6 and B7 were used inherently for \gls{da} of \gls{nmt} systems (fine-tuned on a large corpus), top5 outperformed the best of them (B7) for test set 2011 without retraining on such enormous generic corpus. That is, our proposed data selection method worked well and consequently generated better quality data. In addition to that, our generated corpora are relatively smaller than their proposed methods. Our largest sub-corpora has 1M sentences, whereas B7 proposed three corpora which the smallest is three times larger than top6. Practically speaking, applying \gls{da} as well as using a large in-domain corpus in the work of~\citeA{wang-etal-2017-sentence} caused translation performance to be only 1.5 BLEU points higher than top6. That would be negligible as we did not benefit from any \gls{da} techniques.

\section{Discussions}
\label{CLIN:sec:discussion}
In this section, first we discuss the statistical significance to evaluate the performance of \gls{mt} models that trained on top1, top1+top2 and so on. Second, we investigate the training time of models trained using our proposed data and compare them with the training time of intended systems (S1, S2 and S3). Third, we show how different batch sizes affect the performance of training models in our research. Finally, we investigate the impact of adding more data on the \gls{mt} models' quality, more specifically, to see how much our mixing idea helped the models to improve translation quality.

\subsection{Statistical significance}
\label{CLIN:subsec:ss}

As mentioned earlier in Section \ref{CLIN:sec:results_analayses}, we also computed pairwise statistical significance of scores shown in Table \ref{CLIN:tab:indomain_data_selection} in terms of BLEU, TER and chrF2 scores by using bootstrap resampling and 95\% confidence interval for both test sets based on 1000 iterations and samples of 100, 200 and 300 sentences. According to our experiment outputs, most results have a statistically significant difference except those system pairs listed in Table \ref{CLIN:tab:ss_exception} for some number of samples. %This evaluation reveals at some point models become similar. As such, at this point, the mixing sub-corpora (top6+top7, top7+top8, etc.) degrades the translation quality of MT systems. 
This evaluation highlights that as the training process progresses and more sub-corpora are added, the models' outputs begin to converge, resulting in minimal or no distinguishable performance improvement. 

At this point of convergence, further mixing of sub-corpora (e.g., top6+top7, top7+top8, etc.) leads to a degradation in translation quality. This suggests that while incremental data initially boosts performance, there is a saturation threshold beyond which additional data not only fails to improve quality but may also introduce inconsistencies or noise that negatively impact the \gls{mt} system.

\begin{table}[ht]
\renewcommand{\arraystretch}{1.3}
\centering
\begin{adjustbox}{width=210pt,center}
\begin{tabular}{l|cc|}
\cline{2-3} 
                                    & \textbf{Test Set 2010} & \textbf{Test Set 2011} \\ \hline
\multicolumn{1}{|r|}{\textbf{BLEU}}  & (Top4, Top5, 100)      & (Top5, Top6, 100)      \\ \hline
\multicolumn{1}{|r|}{\textbf{TER}} & \begin{tabular}[c]{@{}c@{}}(Top4, Top5, 100) \\ (Top4, Top5, 200)\\ (Top4, Top5, 300)\end{tabular} & - \\ \hline
\multicolumn{1}{|r|}{\textbf{CHRF2}} & -                    & -                  \\ \hline \hline
\end{tabular}
\end{adjustbox}
\caption{\textbf{System pairs with no statistically significant difference ($p < 0.05$).} The notation (TopX, TopY, N) indicates that systems TopX and TopY exhibit \textbf{no} statistically significant difference for a specific sample size \(N\) (100, 200, or 300 sentences). When it comes to chrF2 and TER (Test Set 2011), we note that all results are statistically significant.}
\label{CLIN:tab:ss_exception}
%\vspace{-3mm}%Put here to reduce too much white space after your table
\end{table}

\subsection{Training time}
\begin{table}[ht]
\renewcommand{\arraystretch}{1.3}
\centering
\begin{adjustbox}{width=\textwidth,center}
\begin{tabular}{l|crrr}
\hline
\hline
\multicolumn{1}{l|}{\textbf{Systems}} &
\textbf{\begin{tabular}[r]{@{}c@{}}Complete TT\end{tabular}} &
\multicolumn{1}{r}{\textbf{Step}} &
\multicolumn{1}{r}{\textbf{\begin{tabular}[c]{@{}c@{}}Best model TT\end{tabular}}}  &
\multicolumn{1}{r}{\textbf{Step}} \\ 
\hline\hline
S1: ID            & 00:03:53   & 18,000 & 00:00:50   & 5,000   \\
S2: OOD           & 02:05:31   & 93,000 & 01:22:56   & 82,000  \\
S3: ID+OOD        & 01:13:30   & 39,000 & 00:10:00   & 29,000  \\ 
\hdashline[1pt/1pt]
Top1              & 00:04:43   & 16,000 & 00:01:23   & 5,000   \\
Top2 + top1 & 00:04:43   & 20,000 & 00:02:13   & 10,000  \\
Top3 + top2 + ... & 00:06:06   & 26,000 & 00:03:03   & 13,000  \\
Top4 + top3 + ... & 00:06:23   & 27,000 & 00:03:53   & 17,000  \\
Top5 + top4 + ... & 00:10:33   & 35,000 & 00:05:50   & 20,000  \\
Top6 + top5 + ... & 00:08:20   & 35,000 & 00:04:26   & 19,000  \\ 
\hline \hline
\end{tabular}
\end{adjustbox}
\caption{\textbf{The training time (abbreviated as TT) of generated sub-corpora and the first category of compared systems.} The times are reported in days (D), hours (H), and minutes (M) for both the complete training duration and the duration until the best model performance was achieved, along with the corresponding step counts.}
\label{CLIN:tab:training_time}
\end{table}

As can be seen in Table \ref{CLIN:tab:training_time}, the training of the large baseline model (S2) with almost 31M sentences not only took 2 days, 5 hours and 31 minutes but also did not perform well in the context of in-domain translation – it obtained 25.8 and 30.7 BLEU score for test set 2010 and 2011, respectively. That means enormous training data is not always helpful for training in-domain \gls{mt} systems. 

%Moreover, according to Table \ref{CLIN:tab:indomain_data_selection} data, the best performed mixed MT models are top5 and top6 for test sets 2011 and 2010, respectively, while, their training time is also considerably less than S2 and even S3. That resulted while S3 was mixed with parallel in-domain data, whereas proposed MT systems were trained without mixing with IWSLT in-domain corpus. 

According to the results in Table \ref{CLIN:tab:indomain_data_selection}, the best-performing models for in-domain translation are top5 and top6, which achieved the highest BLEU scores for test sets 2011 and 2010, respectively. Furthermore, their training times were significantly shorter than those of both S2 and S3. This is notable because, unlike our proposed \gls{mt} models––trained with data selected through our methodology––S3 was trained using a mixture of parallel in-domain data, including the IWSLT in-domain corpus, whereas the proposed models were trained without mixing such data. 

%The training time of top1 and other sub-corpora with no mixing procedure (Table \ref{CLIN:tab:without_mixing}) is nearly the same, hovering around 1 hour and 20 minutes for finding the best model.
The training times for top1 and other sub-corpora models without any data mixing, as shown in Table \ref{CLIN:tab:without_mixing}, were nearly identical—averaging around 1 hour and 20 minutes to find the optimal model.

\subsection{Batch size effect}\label{CLIN:sec:batch_size}
\begin{figure*}[htp]
	\centering 
	\includegraphics[scale=0.56]{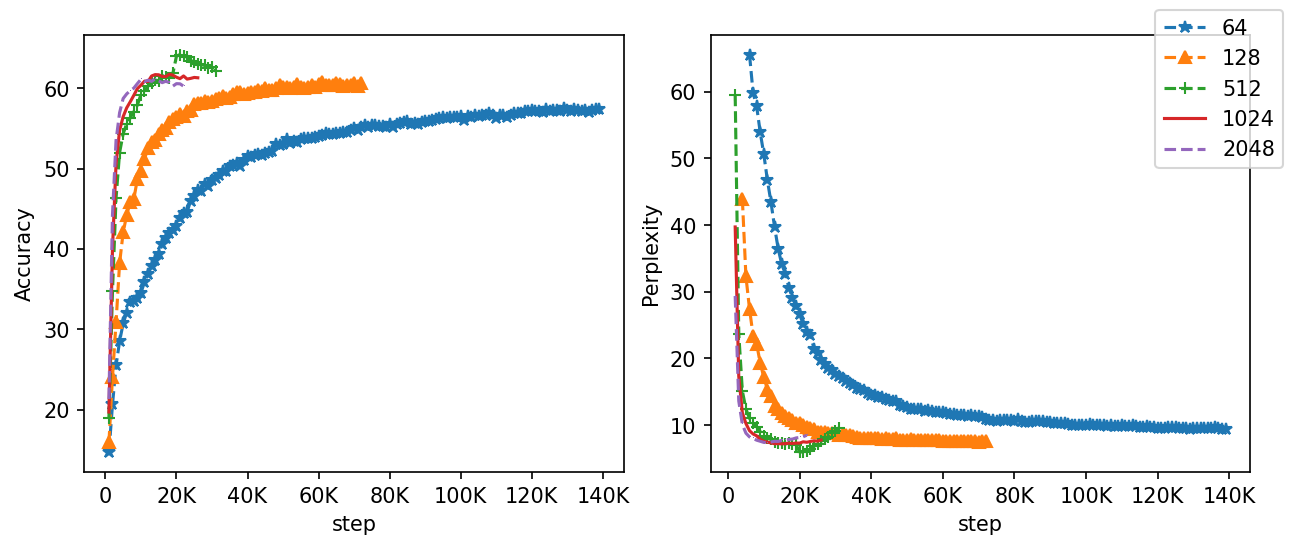} 
	\caption{\textbf{The effect of different batch sizes on training the proposed \gls{nmt} models.} Left and right Figures show validation accuracy and validation perplexity per step, respectively.} 
	\label{CLIN:fig:batch_sizes}
\end{figure*}

In our experiments, we used the default hyperparameters from \citeA{vaswani2017attention} to train the \gls{nmt} models. However, the effects of different batch sizes on the models' performance were investigated as well in order to determine the most suitable one. Our work is pertinent to data selection and the size of selected data has a close correlation with batch sizes. As such, we need to choose a batch size that is optimal for the models according to the training data (i.e., the selected sub-corpora). To this end, we tested five different batch sizes including, 64, 128, 512, 1024 and 2048 for training on top1+..+top6 data. Figure \ref{CLIN:fig:batch_sizes} depicts the accuracy and perplexity for the different batch sizes per step. 

These results show that the selected model with batch size 512 simultaneously reached the highest possible accuracy percentage (61.89\%) and the lowest perplexity score (7.08) within 21K steps. Following these experiments and the results shown in Figure~\ref{CLIN:fig:batch_sizes} we decided to conduct all our experiments (see Section~\ref{CLIN:sec:experiments}) with a batch size of 512. Further investigation is needed in order to define a correlation between (training) data size and batch size for optimal \gls{mt} performance. We leave this for future work.

\subsection{Mixing effect}
In our main experiments, we investigated the translation quality of systems trained on incremental datasets: top1 then top1+top2 and so on. In order to determine whether the quality achieved by the different systems is due to the quality of the data or the quantity (i.e., adding additional data), we trained 6 other models without combining (mixing) the datasets, i.e., a model trained only on top1, a model trained on top2 only and so on. Then we evaluated these models the same way as with the ones presented in Section~\ref{CLIN:sec:experiments}. The results are shown in Table~\ref{CLIN:tab:without_mixing}. 

\begin{table}[ht]
\renewcommand{\arraystretch}{1.3}
\centering
\begin{adjustbox}{width=\textwidth,center}
\begin{tabular}{l|crrrrrr}
\hline
\hline
\multirow{2}{*}{\textbf{Systems}} &
  \multirow{2}{*}{\begin{tabular}[c]{@{}l@{}}\textbf{Sentence} \\ \textbf{Number}\end{tabular}} &
  \multicolumn{3}{c}{\textbf{\gls{nmt}- Test Set 2010}} &
  \multicolumn{3}{c}{\textbf{\gls{nmt}- Test Set 2011}} \\ \cdashline{3-5}[1pt/1pt]\cdashline{6-8}[1pt/1pt]%\cline{3-5}\cline{6-8}
 &
   &
  \multicolumn{1}{r}{\textbf{BLEU$\uparrow$}} &
  \multicolumn{1}{r}{\textbf{TER$\downarrow$}} &
  \multicolumn{1}{r}{\textbf{CHRF2$\uparrow$}} &
  \multicolumn{1}{r}{\textbf{BLEU$\uparrow$}} &
  \multicolumn{1}{r}{\textbf{TER$\downarrow$}} &
  \multicolumn{1}{r}{\textbf{CHRF2$\uparrow$}} \\ \hline
Top1     & 179K     & 21.8      & 69.8       & 50.0           & 25.6       & 64.0       & 53.0  \\
Top2     & 179K     & 21.2      & 72.1       & 49.0           & 24.6       & 67.3       & 52.0  \\
Top3     & 179K     & 21.9      & 71.3       & 49.0           & 25.2       & 66.1       & 52.0  \\
Top4     & 179K     & 21.1      & 71.7       & 49.0           & 24.6       & 66.3       & 52.0  \\
Top5     & 179K     & 20.8      & 72.0       & 49.0           & 24.6       & 67.3       & 51.0  \\
Top6     & 179K     & 21.9      & 69.6       & 49.0           & 24.1       & 66.8       & 51.0  \\ 
\hline \hline
\end{tabular}
\end{adjustbox}
\caption{Comparison of \gls{nmt} system performance on in-domain data selection without mixing sub-corpora.}
\label{CLIN:tab:without_mixing}
%\vspace{-3mm}%Put here to reduce too much white space after your table
\end{table}

%Table \ref{CLIN:tab:without_mixing} shows the performance of in-domain translation of NMT systems trained without mixing the selected sub-corpora. 
According to the three evaluation metrics, all systems performances are on par with system top1 except for minor differences that are unavoidable. That is, according to our experiment (shown in Figure~\ref{CLIN:fig:closeness} in Section \ref{CLIN:sec:data}) the centroids of selected sub-corpora (top1, top2, etc.) are very similar to centroids of in-domain test sets. Recall that a centroid represents the average position of data points in high-dimensional space. This measure helps us compare how closely the sub-corpora align with one another––particularly with the in-domain test sets.

Notably, the centroid of top1 aligns even more closely with the centroids of the 2010 and 2011 test sets than others. To further interpret these results, we computed pairwise statistical significance of BLEU score using bootstrap resampling and 95\% confidence interval. Table \ref{CLIN:tab:SS_without_mixing} indicates whether the differences between two systems' BLEU scores are statistically significant (Y) or not (N). It shows that there was no statistically significant difference between results obtained with systems trained on top2, top3, ..., top6. However, the results (of BLEU) obtained with top1 are statistically significant from all the rest. These are the sentences (from the OOD corpus) that are the most similar to the in-domain data. Table \ref{CLIN:tab:SS_without_mixing} shows the statistical significance computed over both test sets and based on 1000 iterations and samples of 200 sentences. %Note that, Yes and No are abbreviated as 'Y' and 'N' in Table \ref{CLIN:tab:SS_without_mixing}, respectively.

\begin{table}[ht]
\renewcommand{\arraystretch}{1.3}
\centering
\begin{adjustbox}{width=0.6\textwidth,center}
\begin{tabular}{c|cccccc}
\hline\hline
  & Top1 & Top2 & Top3 & Top4 & Top5 & Top6 \\\hline\hline
Top1 &  & Y & Y & Y & Y & Y\\
Top2 &  &  & N & N & N & N\\
Top3 &  &  &  & N & N & N\\
Top4 &  &  &  &  & N & N\\
Top5 &  &  &  &  &  & N\\
Top6 &  &  &  &  &  & \\\hline \hline
\end{tabular}
\end{adjustbox}
\caption{Results of statistically significant test for in-domain data selection without mixing sub-corpora (for $p < 0.05$).}
\label{CLIN:tab:SS_without_mixing}
\end{table}

\begin{figure}[htbp]
  \centering
  \includegraphics[width=\textwidth]{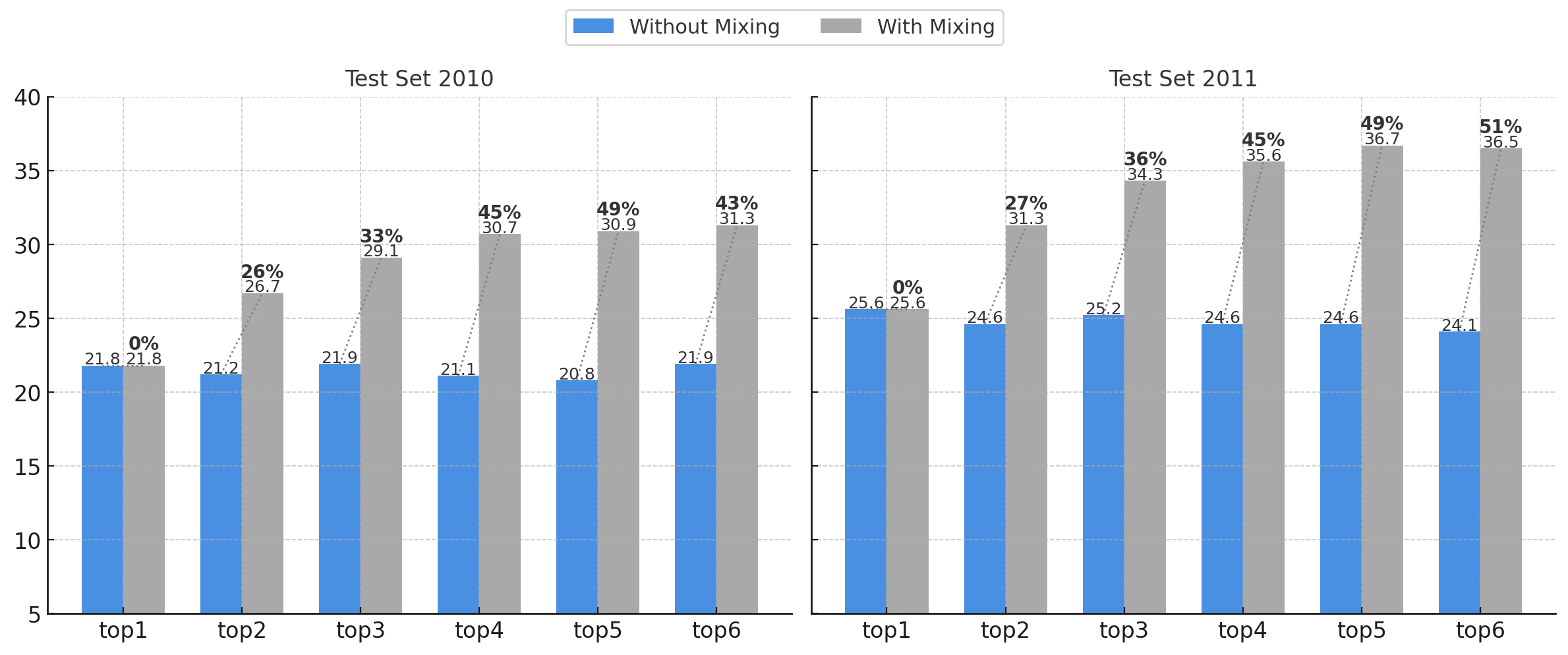}
  \caption{\textbf{Quality improvement (in percentage) after applying the mixing idea on \gls{nmt} systems} for test sets 2010 and 2011.}
  \label{CLIN:fig:mixing_difference}
\end{figure}

Figure \ref{CLIN:fig:mixing_difference} also indicates the quality improvement (in percentage) of \gls{nmt} systems that trained on the mixture data compared to the \gls{nmt} trained on the original sub-corpora without being mixed. According to these figures, the mixing procedure enhanced the translation quality up to 49\% and 51\% for test sets 2010 and 2011, respectively. Overall, there is a gradual increase trend after mixing them with all preceding sub-corpora until the convergence point. 

%Our objectives are twofold: (i) to mix as few sub-corpora as possible and (ii) at the same time increase the translation quality by avoiding the MT models being biased toward in-domain data. For example, top5 and top6 are convergence points for test sets 2011 and 2010, respectively. As such, we stop mixing sub-corpora once we reached our convergence points. For instance, the improvement rate for test set 2010 began with 26\%, then continued to 33\%, 45\% and eventually reached its peak by 49\% before dropping to 43\%. It is noteworthy that top1 had no improvement since we did not mix it with any other sub-corpora.

Our objectives are twofold: (i) to mix as few sub-corpora as possible in order to achieve maximum performance with fewer samples, which reduces training time and computational costs, and (ii) to prevent \gls{mt} models from becoming overly biased toward in-domain data. This helps preserve performance on generic domains and reduces the risk of overfitting. For example, top5 and top6 serve as the convergence points for test sets 2011 and 2010, respectively. We stop mixing sub-corpora once the convergence point is reached.

For test set 2010, the improvement rate started at 26\%, rose to 33\%, then 45\%, peaked at 49\%, and eventually declined to 43\%. For test set 2011, the improvement rate began at 27\%, briefly dropped to 26\%, then increased to 45\%, 49\%, and finally reached a peak of 51\%. It is important to note that top1 showed no improvement since it was not mixed with any other sub-corpora.

\section{Conclusion and future work}
\label{CLIN:sec:conclusions}
In this chapter, we presented a method to help the \gls{mt} community to mitigate a lack of parallel in-domain corpora for many language pairs. Considering multiple standards such as generating high-quality data, designing a scalable architecture, having a reusable pipeline, we present a method for data selection, based on ranking, for the purposes of generating parallel in-domain data as well as \gls{da}. Given a large parallel corpus, the proposed method aims to select data that are semantically similar to a set of in-domain dependencies. Typically, this method would be employed when there is little in-domain or only monolingual data, however, our method is generic and not restricted to the size of the in-domain data. The proposed selection pipeline is made of three main components: (i) a contextual sentence embedding component; (ii) a semantic search component and (iii) a ranking in-domain data component. These have been wrapped and released in a publicly accessible Python tool.

We conducted experiments with different sizes of selected in-domain data. Our experimental results showed that selected parallel corpora generated through our data selection method can be applied directly for domain-specific \gls{nmt} systems. Models trained on the selected data outperformed baseline models and, in some cases, were comparable to or even better than fine-tuned models. We note that our experiments used data up to top-6 as beyond top-6 a larger portion of the selected data becomes more dissimilar and would not contribute to the translation quality of the \gls{nmt} systems.\footnote{This was determined through preliminary empirical evaluation and is not presented or discussed here.} However, for other data and domains this threshold could be different and should be determined on a case-by-case basis. 

As future work, building on the present study, we intend to employ our generated corpora in the context of \gls{da} by further training on a parallel in-domain corpus, which possibly boosts \gls{nmt} systems' performance considerably. Furthermore, our proposed selection method shortens the training time and, at the same time, increases \gls{nmt} translation quality compared to employing an out-of-domain corpus. We would also like to address other important research questions. First, we would like to compare the hard decision on $n$---the number of sentences that can be selected by our algorithm---to a similarity threshold (which is not linked to any specific $n$). Another research direction would be to investigate the effects of our method on other language pairs and domains. 

The selected data and trained models are available at: \url{https://github.com/JoyeBright/DataSelection-NMT}. 

\section{Python tool}\label{CLIN:tool}
%As the volume of data for Machine Translation~(MT) grows, the need for models that can perform well in specific use cases, like patent and medical translations, becomes increasingly important. Unfortunately, generic models do not work well in such cases, as they often fail to handle domain-specific style and terminology. Only using datasets that cover domains similar to the target domain to train MT systems can effectively lead to high translation quality (for a domain-specific use-case)~\cite{wang-etal-2017-sentence,pourmostafa2021novel,6dee920f972949d6befe7c6f8d594a93}. This highlights the limitation of data-driven MT when trained on general-domain data, regardless of dataset size.

%To address this challenge, researchers have implemented various strategies to improve domain-specific translation using DA methods~\cite{saunders2022domain,sharami2023tailoring}. The DA process involves initially training a generic model, which is then fine-tuned using a domain-specific dataset~\cite{chu-wang-2018-survey}. One approach to generating a domain-specific dataset is to select similar data from generic corpora for a specific language pair, and then utilize both general (to train) and domain-specific (to fine-tune) parallel corpora for MT. In line with this approach, we developed \textit{a language-agnostic Python tool implementing the methodology proposed by~Sharami et al.~\shortcite{sharami2022selecting}}. This tool uses monolingual domain-specific corpora to generate a parallel in-domain corpus, facilitating data selection for DA.

Building on the methodology outlined in this chapter, we developed a Python tool to automate and streamline the process of selecting high-quality in-domain data for \gls{da}. This tool enables the \gls{mt} community to easily extract relevant domain-specific data from large general-domain corpora, enhancing translation quality and reducing computational costs. It implements the core components of our approach, including contextual embedding generation, semantic search, ranking, and data selection.

The tool's operation requires three inputs: (i) a parallel generic corpus for the source language, (ii) a parallel generic corpus for the target language (iii) a monolingual domain-specific corpus for the source language. An optional parameter allows users to specify the number of selected sentences. 
Once these inputs are provided, the tool leverages a pre-trained Sentence-BERT (S-BERT) model~\cite{reimers-gurevych-2019-sentence} to create sentence embeddings for the input corpora (i.e., i and iii).
To optimize computational efficiency, the tool applies PCA, reducing the original 768-dimensional embeddings to 32 dimensions. If the size of the corpus (iii) is exceeded by the desired number of selected data, the generic corpora are split into multiple equal parts, and each of these parts is used separately in the subsequent step.

In the next step, semantic search is performed to match general-domain sentences with the most similar domain-specific sentences based on cosine similarity. This is done by comparing the vectors of sentences, generated by S-BERT, and ranking them based on their cosine similarity score. The sentence with the highest similarity score is labeled as Top 1, while the one with the lowest similarity score is labeled as Top $N$. By default, $N$ is set to 5, but users can choose a different value for $N$. For each split, the tool then creates a CSV file that includes information about the domain-specific sentence (labeled as \texttt{Query}), the top selected source and target sentences (labeled as \texttt{$topN_{src}$} and \texttt{$topN_{trg}$}), and their corresponding similarity scores. By concatenating the CSV columns generated, one can obtain as much data as previously requested.

Our tool is particularly useful to the \gls{mt} community as it addresses the scarcity of parallel domain-specific data across different language pairs. By using our tool, users can seamlessly select domain-specific data from generic corpora to train a domain-specific \gls{mt} model. This tool is typically used when there is a lack of domain-specific data or when only monolingual data is available. However, our tool is generic and not limited to the size of the domain-specific data.

Our tool is licensed under the MIT License and is accessible to the public for free at \href{https://github.com/JoyeBright/domain-adapt-mt}{https://github.com/JoyeBright/domain-adapt-mt}.
\chapterseparatorpage
\chapterimage{Assets/separator.pdf}

\prechaptertext{This chapter is based on the following published paper:
\\\\
\hspace*{0.5em}Javad Pourmostafa Roshan Sharami, Dimitar Shterionov, Frédéric Blain, Eva Vanmassenhove, Mirella De Sisto, Chris Emmery, and Pieter Spronck. 2023. \href{https://aclanthology.org/2023.eamt-1.2/}{Tailoring Domain Adaptation for Machine Translation Quality Estimation}. In Proceedings of the 24th Annual Conference of the European Association for Machine Translation, pp. 9–20, Tampere, Finland.
\\\\
}
{Improvements have been made to the title, figures, and certain sections to align the content with the broader context of this dissertation.}

\chapter[Domain Adaptation for Machine Translation Quality Estimation]{Domain Adaptation for Machine Translation Quality Estimation}
\chaptermark{EAMT}
\label{chap:EAMT}

\lettrine[lines=4]{W}{hile} \gls{qe} can play an important role in the translation workflow, its effectiveness relies on the availability and quality of the training data. For many \gls{qe} use cases––including various domains and language pairs––access to high-quality labeled data remains a significant challenge. This is largely due to the significant cost and effort required for manual data annotation, limiting the scalability and effectiveness of \gls{qe} in these contexts. Unlike \gls{mt} data, \gls{qe} data requires the collection of source, \gls{mt}, and a succinct human quality assessment metric, such as HTER scores. Aside from the \textit{data scarcity} challenge, \gls{qe} models should also be generalizable; i.e., they should be able to \textit{handle data from different domains}, both generic and specific. Since few pre-trained \gls{qe} models exist to provide cross-language generalization (unlike \gls{mt} models), training on generic data is crucial for learning foundational patterns in translation quality. Fine-tuning with domain-specific data further enhances performance for specialized contexts. 

To alleviate these two main issues (data scarcity and domain mismatch), this chapter combines \gls{da} and \gls{dag} in a robust \gls{qe} system. Our method first trains a generic \gls{qe} model and then fine-tunes it on a specific domain while retaining generic knowledge. Our results show a significant improvement for all the language pairs investigated, better cross-lingual inference, and a superior performance in \gls{zsl} scenarios as compared to state-of-the-art baselines.
\newpage

\section{Introduction}
\label{EAMT:sec:intro}
Predicting the quality of \gls{mt} output is crucial in translation workflows. Informing translation professionals about the quality of an \gls{mt} system allows them to quickly assess the overall usefulness of the generated translations and gauge the amount of post-editing that will be required~\cite{tamchyna-2021-deploying,murgolo-etal-2022-quality}. \gls{qe} is an approach that aims to reduce the human effort required to judge the quality of an \gls{mt} system by assessing the quality of its output without the need for reference translations.\footnote{In contrast to quality evaluation, which compares \gls{mt} output to reference translations.}\looseness=-1

\gls{qe} can be applied on word-, sentence- or document-levels. The goal of sentence-level \gls{qe}, which is the focus of our work, is to predict a quality label based on a source sentence and its \gls{mt} equivalent. This label, i.e., the quality estimate, can be expressed in various ways such as TER/HTER~\cite{snover-etal-2006-study}, BLEU~\cite{papineni-etal-2002-bleu} or any metric of interest to the user. Training a sentence-level \gls{qe} system typically requires aligned data of the form: \textit{source sentence} (SRC), \textit{target sentence} (TRG), and \textit{quality gold label} (LBL).
%However, obtaining such data is difficult and expensive as it requires a combination of MT, post-editing, and evaluation of the post-edited text against the MT text~\cite{rei-etal-2020-comet,Zouhar2023PoorMQ}.
However, most quality labels are by-products of \gls{mt} and post-editing---a rather difficult and expensive process---limiting the size of the available \gls{qe} data~\cite{rei-etal-2020-comet,Zouhar2023PoorMQ}.\looseness=-1

The WMT \gls{qe} shared task~\cite{specia-etal-2021-findings,zerva-etal-2022-findings} has been a platform since 2012~\cite{ws-2012-statistical} to compare different \gls{qe} systems and to share \gls{qe} data. % (both training and testing). 
%Nonetheless, in general, QE data, i.e., (SRC, TRG, LBL) triplets, is scarce and only has a small coverage over language pairs and domains~\cite{2020arXiv201004480F,fomicheva-etal-2022-mlqe}. 
Despite efforts from initiatives like the \gls{qe} shared task to publicly release \gls{qe} datasets, such resources remain scarce across language pairs and, by extension, also have a limited coverage across domains~\cite{2020arXiv201004480F,fomicheva-etal-2022-mlqe}.
%This scarcity and smallness of data could pose a challenge for all QE models, especially recent ones that utilize large pre-trained language models (LLMs) like~\cite{ranasinghe-etal-2020-transquest,HAN2021225}, since fine-tuning pre-trained models with small datasets has been demonstrated to be quite unstable~\cite{devlin-etal-2019-bert,2020arXiv200605987Z}.
This can pose a challenge for all \gls{qe} models, especially recent ones that utilize large \glspl{plm}~\cite{ranasinghe-etal-2020-transquest,zerva-etal-2022-findings} since fine-tuning pre-trained models with small datasets has been demonstrated to be quite unstable~\cite{2020arXiv200605987Z,rubino-2020-nict}.\footnote{Fine-tuning \glspl{plm} for \gls{qe} has gained popularity and is becoming the new standard in the field.}

%Furthermore, QE models trained on a specific data are hardly generalizable to other domains, and their application on domains that are outside of the training domain leads to poor performance due to \textit{domain mismatch}~\cite{c-de-souza-etal-2014-machine,Zouhar2023PoorMQ}. That is, to develop QE models that can apply in multiple domains / use-cases, it is important to exploit the right balance of generic and domain-specific training data.
%Furthermore, QE models trained on specific data do not generalize well to other domains that are outside of the training domain~\cite{kocyigit-etal-2022-better}. In the previous chapter~\ref{chap:CLIN}, we discussed how \textit{Domain mismatches} can hinder the performance of MT systems. While the concept has been extensively studied in MT, it poses significant challenges for QE. That is, it leads to significant decreases in the performance of QE models~\cite{c-de-souza-etal-2014-machine,Zouhar2023PoorMQ}. To improve the generalizability of QE models, it is important to establish the right balance between domain-specific and generic training data.

Furthermore, \gls{qe} models trained on specific data do not generalize well to other domains that are outside of the training domain~\cite{kocyigit-etal-2022-better}. In Chapter~\ref{chap:CLIN}, we discussed \textit{domain mismatches} in \gls{mt} using \gls{da} techniques for in-domain data selection, which improved translation quality by better handling specialized vocabulary and context. In this chapter, we explore similar challenges in \gls{qe}, where domain mismatches degrade performance due to limited domain-specific labeled data~\cite{c-de-souza-etal-2014-machine,Zouhar2023PoorMQ}. While \gls{da} methods benefit \gls{mt}, applying them to \gls{qe} involves additional complexities like data scarcity and cross-domain generalization issues. To improve the generalizability of \gls{qe} models, it is important to establish the right balance between domain-specific and generic training data.
%Furthermore, developing QE models is challenging because it involves finding a balance between being generalizable and specific to a particular domain. That is, QE models require a large volume of data to be generic, but they must also be specific enough to accurately evaluate the quality of an MT system; otherwise, domain mismatch occurs. %Despite their generality~\cite{10.1145/3458754}, LLMs are not a solution to this issue because they lack information about quality labels, and their use does not guarantee the creation of generic QE model. 
To date, only a few attempts have been made to address this challenge~\cite{de-souza-etal-2014-towards,rubino-2020-nict,lee-2020-two}, and as a result, many \gls{qe} models still struggle to maintain accuracy across different domains, particularly when faced with limited domain-specific training data~\cite{Zouhar2023PoorMQ}.
%Thus, the majority of QE models have difficulty with accurately estimating quality across different domains, whether they are generic or specific~\cite{Zouhar2023PoorMQ}.\looseness=-1

%In this work we aim to tackle the aforementioned two issues that impact the training of LLM-based QE models: (i) the scarcity of QE data, and (ii) the difficulty of estimating data from various domains. To overcome these challenges, \textit{we propose a methodology that uses a small amount of domain-specific data to boost the QE prediction performance}. It is inspired by previous work on domain adaptation (DA) in MT, where a large generic model is initially trained and then fine-tuned with domain-specific data~\cite{chu-wang-2018-survey,Saunders2021DomainAA,pham-etal-2022-multi}.
%We propose to tackle both the data scarcity and the domain mismatch challenge that LLM-based QE models face through \textit{a methodology whereby a small amount of domain-specific data is used to boost the overall QE prediction performance.} This approach is inspired by work on DA in the field of MT, where a large generic model is initially trained and then fine-tuned with domain-specific data~\cite{chu-wang-2018-survey,pham-etal-2022-multi}.

We propose to tackle both the data scarcity and the domain mismatch challenge that \gls{plm}-based \gls{qe} models face through \textit{a methodology whereby a small amount of domain-specific data is used to boost the overall \gls{qe} prediction performance.} This approach builds upon the \gls{da} techniques discussed in Chapter~\ref{chap:CLIN}, where we addressed domain mismatches in \gls{mt} by selecting in-domain data to fine-tune large, generic models, thereby enhancing their ability to handle specialized vocabulary and context effectively. Similarly, in \gls{qe}, we leverage domain-specific data to improve prediction accuracy, ensuring models can generalize better across different domains~\cite{chu-wang-2018-survey,pham-etal-2022-multi}.
%Specifically, we use a DA technique to fine-tune an LLM with a very large QE dataset (generic) to enable generalization of the model. This is followed by fine-tuning with a mix of domain-specific and generic QE data to allow the model to learn both specific and generic domains. Lastly, we fine-tune the model with small volumes of domain-specific QE data to boost performance for the specific domain of interest.

To assess the validity of the proposed approach in \gls{qe} (See Section~\ref{EAMT:sec:methodology} for a detailed description of our methodology), we conducted experiments using small and large, authentic and synthetic data in bilingual, cross-lingual, and \gls{zs} settings. We experimented with publicly available language pairs from English (EN) into German (DE), Chinese (ZH), Italian (IT), Czech (CZ), and Japanese (JA) and from Romanian (RO) and Russian (RU) into English (EN). We used the common test sets from the WMT2021 \gls{qe} shared tasks.\footnote{\url{https://www.statmt.org/wmt21/quality-estimation-task.html}} 

%Our experiments show a statistically significant improvement in the performance of QE models when compared to the baselines\footnote{Baseline models were trained according to common practices in QE (see Section~\ref{EAMT:sec:baseline} for details).}.
Our experiments show a statistically significant improvement in the performance of \gls{qe} models.
Our findings also indicate that not only our implementation leads to better multi-/cross-lingual \gls{qe} models (where multi-/cross-lingual data is provided) but also \gls{zs} \gls{qe} (where no data for the evaluated language pairs was provided at training). 

The main contributions of our research are summarized in this chapter as follows:
\begin{itemize}[leftmargin=*,noitemsep]
    \item A \gls{qe} methodology that employs \gls{da} and \gls{dag}, along with a novel \gls{qe} training pipeline that supports this methodology.
    \item An empirical demonstration of the pipeline's effectiveness, which highlights improvements in \gls{qe} performance, and better cross-lingual inference.
    \item A comparative analysis with state-of-the-art (SOTA) baseline methods that demonstrates the effectiveness of our approach in enhancing \gls{zsl} for the task of \gls{qe}.
    \item Adaptable \gls{qe} pipelines that can be tailored and implemented for other language pairs; i.e., highly generalizable \gls{qe} pipelines.
\end{itemize}

To the best of our knowledge, this is the first \gls{qe} methodology to use \gls{da} and \gls{dag}. Furthermore, it is easily reusable and adaptable: (i) while we used XLM-R in our experiments, one can easily replace it with any preferred \gls{plm} as long as the input-output criteria are met; (ii) we built our tool around Hugging Face~(HF) implementations of \glspl{plm}, meaning one can employ a certain generic model and apply it to any \gls{qe} task by simply fine-tuning it on (newly-collected) \gls{qe} data.\looseness=-3
%\newline\newline

%It should be noted that our objective in this study is not to outperform the results of other QE frameworks but to explore the performance of QE models with and without the proposed DA method.

% This paper presents a description of relevant prior works in Section~\ref{EAMT:sec:related_works}. The methodological approach that we follow is detailed in Section~\ref{EAMT:sec:methodology}. Section~\ref{EAMT:sec:experimental_setup} provides an outline of the datasets, QE and MT frameworks, models training procedures and evaluation metrics. We present our findings in Section~\ref{EAMT:sec:results}, followed by a further analysis of the results and the conclusion of the paper in Section~\ref{EAMT:sec:discussion} and Section~\ref{EAMT:sec:conclusion}, respectively.

\section{Domain adaptation for specialized quality estimation} 
\label{EAMT:sec:methodology}
In this section, we outline our methodology for training \gls{plm}-based \gls{qe} models for a specific domain with limited available in-domain data. This involves: (i) a set of training steps that we found to be particularly effective, %which we recommend for other researchers to consider
and (ii) \gls{dag} techniques to improve the \gls{qe} models' specificity. Also, we provide details on two different training modes we implemented. %(with or without tags).

\subsection{Training steps}
\label{EAMT:sec:training_steps}
\begin{figure}[h!]
    \centering
     \includegraphics[keepaspectratio,width=\columnwidth]{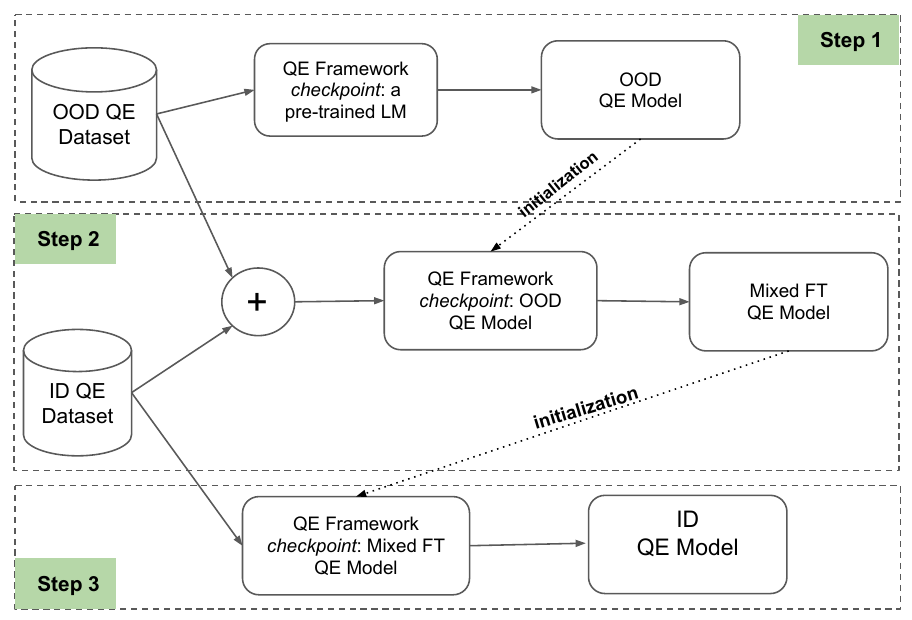}
    \caption{\textbf{Overview of the proposed training steps for specialized \gls{qe}.} The ``+'' sign indicates the oversampling performed in Step 2 to balance the use of ID and OOD data. The dashed arrows indicate the source of the checkpoint used to initialize the models in each stage.}
    \label{EAMT:fig:training_Pipeline}
\end{figure}

We adapt the ``mixed fine-tuning + fine-tuning'' \gls{da} technique that proved promising for \gls{mt}~\cite{chu-etal-2017-empirical} to suit our needs. A visualization of the steps involved can be found in Figure~\ref{EAMT:fig:training_Pipeline}. Our technique involves leveraging both in-domain (ID) and out-of-domain (OOD) \gls{qe} data (see Section~\ref{EAMT:sec:data} for details on the datasets).

\paragraph{Step 1} We train a \gls{qe} model using OOD data until it converges. We employ the experimental framework described in Section~\ref{EAMT:sec:QE} in which a \gls{plm} is fine-tuned to predict \gls{qe} labels. The goal of this step is two-fold: (i) leveraging the \gls{plm}'s cross-lingual reference capabilities and (ii) building a generic \gls{qe} model. This way we ensure that the model can estimate the quality of a broad range of systems, but with limited accuracy on ID data.
 
\paragraph{Step 2} The model's parameters are fine-tuned using a mix of OOD and ID data. We use different ID data, both authentic and synthetic according to the \gls{dag} approaches in Section~\ref{EAMT:sec:DAG}. The objective here is to ensure the model does not forget generic-domain knowledge acquired during the first step while simultaneously improving its ability to perform \gls{qe} on the domain-specific data. This mixing step is often referred to as \emph{oversampling} in \gls{da} literature, where \emph{a smaller subset of OOD data is concatenated with ID data to allow the model to assign equal attention to both datasets}; it aims to further adapt the model to the specific domain of interest. \looseness=-1
%\textit{one of the proposed DAG approaches}.

\paragraph{Step 3} We continue to train the \gls{qe} model on a specific ID dataset until convergence, resulting in a more domain-specific \gls{qe} model than the one obtained in Step 2.

% In Figure~\ref{EAMT:fig:training_Pipeline}, we present an overview of the proposed training steps for specialized QE.

\subsection{Data augmentation for domain adaptation in QE}
\label{EAMT:sec:DAG}
In the previous chapter, we demonstrated that \gls{da} techniques significantly improve translation quality in \gls{mt} by addressing domain mismatch through ID data selection. However, even with these methods, a key challenge remains: \emph{the availability of domain-specific data is often limited, particularly for low-resource languages and specialized domains.} This scarcity of ID data can prevent both \gls{mt} and \gls{qe} models from generalizing effectively across different domains. In \gls{qe}, this issue is compounded by the additional need for labeled data, such as post-edited translations, which are costly and time-consuming to \mbox{produce}.

%To address these challenges, we explore DAG as a solution to both data scarcity and domain mismatch in QE. Our approach aims to maximize the utility of available resources by supplementing ID data through oversampling and synthetic data generation, thereby improving the model's domain-specific accuracy without sacrificing cross-domain generalizability.

%We propose two DAG strategies for DA in QE to optimize ID data utilization:

%In our study, we explore two alternative approaches to oversampling to optimize the utilization of available ID resources and assess the potential benefits of incorporating synthetic ID data into the QE pipeline: %The first DAG uses \textit{all available authentic ID data across all language},
% previous version
%The first DAG employs a fully exploitation approach to augment data using \textit{all available authentic ID data}, 
% END previous version
%whereas the second one relies on generating \textit{synthetic} data using an MT model. The resulting data (ID) is concatenated with the OOD to be used in step (2) of the training pipeline.

To overcome these challenges, we explore two alternative approaches to oversampling in this study. These approaches aim to optimize the utilization of existing ID resources and assess the potential benefits of incorporating synthetic ID data into the \gls{qe} pipeline, thereby improving domain-specific accuracy without compromising cross-domain generalizability.

\paragraph{Approach 1: Concatenating all available authentic ID data across all languages.}
The XLM-R model is multilingual, allowing us to apply it to different language pairs. When there is not enough data to fine-tune it for a specific language, one can use multilingual data. In our work, to increase the amount of authentic data (given the small volume of parallel data for two languages), we construct a multilingual ID dataset: we concatenate all available ID data, which includes different language pairs. %In mathematical terms, Let $D$ be the collection of datasets, and $id_i$ and $id_j$ represent the ID set for the $i$-th and $j$-th language pair in $D$, respectively. 
%For example, $id_1$ could refer to the in-domain data for language pair $EN-DE$ while $id_2$ for $EN-ZH$. 
%We assume that the number of IDs in the collection is $N$, and use the symbol $id_i^\frown id_j$ to denote the concatenation of $id_i$ and $id_j$. Following this approach, we concatenate all the datasets in the collection as $id_1^\frown id_2^\frown \dots^\frown id_n$. 
The rationale behind this approach is to make use of all available authentic resources in order to improve the performance of the \gls{qe} model by providing better cross-lingual references.

\begin{figure}[ht!]
    \centering
    \includegraphics[keepaspectratio,width=0.9\textwidth]{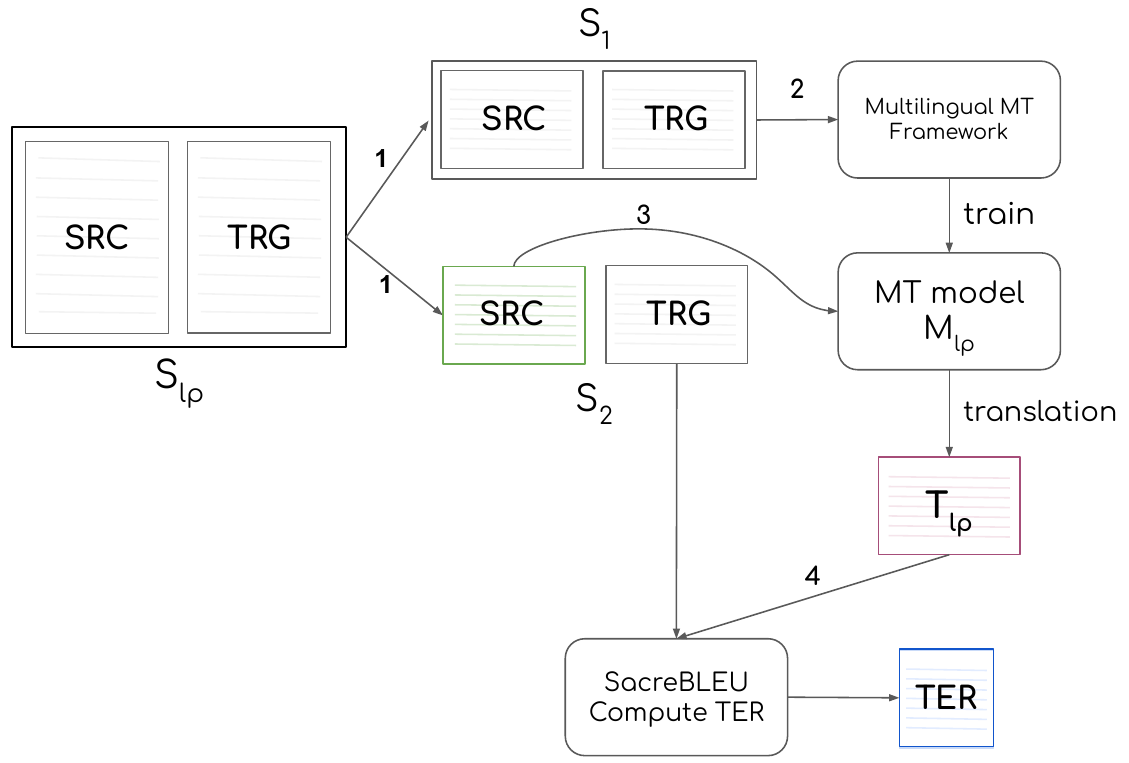}
    \caption{\textbf{Overview of Approach 2 (Generating synthetic ID) of \gls{dag} for \gls{da} in \gls{qe}.} The various steps involved in the approach are indicated close to the corresponding arrows. Arrow 1 represents subsampling. The abbreviations $SRC$, $TRG$, and $T_{lp}$ stand for source, target, and machine-translated text, respectively. The final outputs, which include $SRC$, $T_{lp}$, and quality labels ($TER$), are color-coded for clarity.}
    \label{EAMT:fig:approach2}
\end{figure}

\paragraph{Approach 2: Generating synthetic ID data.}
Given that all available ID resources have been already utilized in Approach 1, we propose to supplement the existing data with artificially generated additional ID data using a trained \gls{mt} model for each language pair, inspired by the research conducted by Negri et al.,~\shortcite{negri-etal-2018-escape} and Lee~\shortcite{lee-2020-two}. This approach aims to tackle the data scarcity problem and further improve the \gls{qe} model's accuracy.
Let $D_{lp}$ denote the publicly available parallel data (SRC, TRG) for a language pair e.g., $lp = \text{EN-DE}$. Then for each ID involved, do:
\begin{enumerate}[leftmargin=*,noitemsep]
    \item Randomly select $N$ samples from $D_{lp}$ to obtain a set $S_{lp}$ of training samples. Divide $S_{lp}$ into two equal sets $S_1$ and $S_2$. %, each containing $N/2$ training samples.
    The division is random to ensure that both sets are representative of the data distribution. 
    \item Train a multilingual \gls{mt} model $M_{lp}$ on $S_1$. %(details of the model can be found in Section~\ref{EAMT:sec:MT}).
    \item Use $M_{lp}$ to translate the sources-side of $S_2$ (or a portion of it), obtaining a set $T_{lp}$ of translated samples.
    \item Compute quality labels (TER/HTER) by comparing $T_{lp}$ with the reference ($TRG$) text from $S_2$.
\end{enumerate}
The result is a triplet ($S_2$, $T_{lp}$, and TER/HTER) to be used to train \gls{qe} models. A visual representation of these steps can be found in Figure~\ref{EAMT:fig:approach2}.

% Figure~\ref{EAMT:fig:approach2} presents an overview of Approach 2 that is employed for data augmentation in the context of domain adaptation for QE.
% \label{EAMT:appendix:approach2}
 
\subsection{Additional indication of domain}
In \gls{nmt}, in order to handle multiple domains and reduce catastrophic forgetting, \gls{da} has been controlled using additional tags added at the beginning or at the end of the sentence~\cite{sennrich-etal-2016-controlling,mattoni-etal-2017-zero,Chu2019MultilingualMA}.
%Previous successful use of tags in NMT, particularly in the context of DA, has been reported in the literature~\cite{sennrich-etal-2016-controlling,Chu2019MultilingualMA,stergiadis-etal-2021-multi}.To handle different domains and reduce catastrophic forgetting~\cite{MCCLOSKEY1989109}  
Following these studies, we explore two training modes: (i) with tag (``TAG'') appending either an \verb|<OOD>| or an \verb|<ID>| tag at the end of sentences based on the dataset domain type (i.e., OOD or ID). The input format in this mode is \verb|<s> SRC </s> TRG <Tag> </s>|, where SRC and TRG represent source and target of the \gls{qe} triplet, and \verb|<s>| and \verb|</s>| are the beginning and separator tokens for the \gls{plm} used in the pipeline\footnote{The \texttt{<s>} and \texttt{</s>} tokens are special symbols typically used in the \gls{plm} to indicate the start and separation of input segments. These tokens help the model structure its input and output during training and inference.}. (ii) without tag (``NO TAG''), where the training steps are the same as detailed in Section~\ref{EAMT:sec:training_steps}, i.e., no additional info is attached. %, but without appending any additional tags to the sentences.

The methodology consists of three interconnected stages: training steps, \gls{dag}, and domain tagging. The training steps gradually adapt the model from OOD to ID data, improving domain-specific performance. \gls{dag} enhances this process by increasing ID data availability through both authentic data concatenation and synthetic data generation. Domain tagging helps the model differentiate between OOD and ID data, reducing domain confusion and improving cross-domain generalization. Jointly, these stages address data scarcity and domain mismatch, ensuring a more robust and adaptable \gls{qe} model.

We use the same domain tags for both original and synthetic ID data because the tags encode 
only the \emph{domain} distinction (ID vs.\ OOD), not the data provenance. The synthetic 
samples generated in \gls{dag}\,2 are intended to represent the same in-domain distribution 
as the original ID data; therefore, assigning different tags would incorrectly signal a domain 
difference where none exists.

% \begin{example}\small 
% A sentence with and without tag:\\
% \noindent\textbf{With tag:} \verb|<s>| Victoriile sârbilor din 1914 au fost plătite cu pierderi grele însă . \verb|</s>| The 1914 Serbian victims were paid with heavy losses . \verb|<ID>| \verb|</s>|

% \noindent\textbf{Without tag:} \verb|<s>| Victoriile sârbilor din 1914 au fost plătite cu pierderi grele însă . \verb|</s>| The 1914 Serbian victims were paid with heavy losses . \verb|</s>|
% \end{example}
\section{Experiments}
\label{EAMT:sec:experimental_setup}
\subsection{Data}
\label{EAMT:sec:data}
We conducted experiments on publicly available data in different languages: from EN into DE, ZH, IT, CZ, and JA and from RO and RU into EN. These languages were chosen to cover diverse linguistic properties, such as morphology (DE, CZ), syntax (JA), and writing systems (ZH, JA). This ensures robust evaluation across varied challenges in \gls{mt}. We categorize the data into three groups according to their use in our pipeline: 

\paragraph{Group 1: for building \textit{ID} and \textit{OOD} \gls{qe} models.} The \textit{ID} data is collected from WMT 2021 shared task on \gls{qe}~\cite{specia-etal-2021-findings}, Task 2, consisting of sentence-level post-editing efforts for four language pairs: EN-DE, EN-ZH, RU-EN and RO-EN. For each pair there are train, development (dev), and test sets of 7$K$, 1$K$, 1$K$ samples, respectively. For our \textit{OOD} data, we used the eSCAPE~\cite{negri-etal-2018-escape} dataset with approximately 3.4$M$ tokenized SRC, machine-translated text~(MTT), post-edited~(PE) sentences. In our experiments, we used only the EN-IT portion of the eSCAPE corpus. We used \texttt{sacrebleu}\footnote{signature:nrefs:1$|$case:lc$|$tok:tercom$|$punct:yes$|$version:2.3.1}~\cite{post-2018-call} to calculate TER~\cite{snover-etal-2006-study} from MTT and PE pairs. We split the data into train, dev, test sets via the \texttt{scikit-learn} package\footnote{random state/seed$=$8, shuffle$=$True, used for all splits.}~\cite{pedregosa2011scikit} with 98\%, 1\%, and 1\% of the total data, respectively.
% \footnote{\cite{datasplit} recommends this for datasets $>$1M examples.}  
To improve the generalization of our models and enable them to better adapt to specific \gls{qe} through the ID dataset, we utilized a larger OOD dataset. This decision is in line with prior studies on \gls{da}, which are described in the related work section (Section~\ref{EAMT:sec:related_works}).

\paragraph{Group 2: for building \gls{mt} systems as a component of \textit{Approach 2} in the proposed \gls{dag} (Section~\ref{EAMT:sec:DAG}).} We collected parallel data---SRC and reference translations (REF)--- from Opus \cite{tiedemann-2012-parallel} for each language pair used in ID (EN-DE, EN-ZH, RO-EN, and RU-EN), all sourced from ParaCrawl. Next, we trained \gls{mt} models for Approach 2 of our methodology by selecting 4$M$ samples and dividing them into two equal parts, each with 2$M$ samples. %Using the \texttt{scikit-learn} package with a ``random seed'' of 777. 
We split either of the two parts into train, dev, and test sets. To save time during evaluation and inference, we set the size of the dev and test splits to be the same as the number of training samples in the ID datasets, which is 7$K$. The reason for reducing the number of samples to 7$K$ is to save time during evaluation and inference. Translating 20K samples takes about three times longer than translating 7K samples.
Moreover, we randomly selected a portion of the SRC (7$K$ out of 2$M$) in the second split, which was not used for training. We passed this portion to the trained \gls{mt} to get MTT. Finally, we computed the TER using the MTT and the corresponding REF via \texttt{sacrebleu}. We set the portion size 7$K$ as the goal was to double the size of the initial ID data. 

\paragraph{Group 3: for testing the ZS capabilities of the trained \gls{qe} models in our proposed methodology.} We used two test sets with language pairs that were not seen during training, namely English to Czech (EN-CZ) and English to Japanese (EN-JA), which were provided by WMT 2021 shared task on \gls{qe} for Task 2. Each test set contained 1K samples. 

A summary of the data groups used in the experiments is presented in Table~\ref{EAMT:table:data_groups}.

\begin{table}[ht!]
    \setlength\tabcolsep{8pt}
    \renewcommand{\arraystretch}{1.2}
    \begin{adjustbox}{width=\textwidth,center}
    \centering
    \begin{tabular}{l|l}
    \hline
    \hline
    \textbf{Data Group} & \textbf{Description} \\
    \hline
    \hline
    Group 1: \textit{ID} and \textit{OOD} \gls{qe} models &
    \begin{tabular}[t]{p{0.85\textwidth}}
    - \textit{ID} data: WMT 2021 (EN-DE, EN-ZH, RU-EN, RO-EN), 7K/1K/1K samples \\ 
    - \textit{OOD} data: eSCAPE (3.4M samples), EN-IT only \\
    - TER via \texttt{sacrebleu}, data split with \texttt{scikit-learn} \\
    \end{tabular} \\ 
    \hline
    Group 2: \gls{mt} systems for DAG &
    \begin{tabular}[t]{p{0.85\textwidth}}
    - Parallel data: Opus (ParaCrawl), 4M samples (EN-DE, EN-ZH, RO-EN, RU-EN) \\
    - Train/dev/test split: 7K samples \\
    - TER from machine-translated text (MTT) and reference (REF) \\
    \end{tabular} \\
    \hline
    Group 3: Zero-shot testing &
    \begin{tabular}[t]{p{0.85\textwidth}}
    - Test sets: EN-CZ, EN-JA (1K samples each) \\
    - Provided by WMT 2021 \gls{qe} shared task \\
    \end{tabular} \\
    \hline
    \hline
    \end{tabular}
    \end{adjustbox}
    \caption{Summary of data groups used in the experiments.}
    \label{EAMT:table:data_groups}
\end{table}

%All necessary scripts for reproducing the data used in our study can be found in~\url{http://anonymous.me}.

\subsection{Frameworks}
\subsubsection{Quality Estimation}
\label{EAMT:sec:QE}
To train all \gls{qe} models of our study, we developed a new \gls{qe} framework with the ability to invoke multilingual models from the HF model repository. The framework is similar in architecture to ``MonoTransQuest''~\cite{ranasinghe-etal-2020-transquest}, but adapted to the needs of our experiments. The differences with ``MonoTransQuest'' are the additional tokens (\verb|<OOD>| and \verb|<ID>|) added  during the tokenization process%\footnote{\url{https://github.com/huggingface/transformers/blob/v4.26.1/src/transformers/tokenization_utils_base.py#L938}, special tokens$=$True}
, as well as the resizing of the model's token embeddings in order to support the added tags. Also, rather than computing the softmax, we directly used logits to estimate the quality labels. This approach saves time, as computing softmax can be expensive~\cite{ruder2016wordembeddingspart2} and we do not need to find the probabilities of each prediction for our experiments. In all our experiments, we chose to use XLM-RoBERTa\footnote{xlm-roberta-large}~(XLM-R)~\cite{conneau-etal-2020-unsupervised}, to derive cross-lingual embeddings, which has shown success in prior studies such as~Ranasinghe et al.,~\shortcite{ranasinghe-etal-2020-transquest}.

\subsubsection{Training and evaluation details of \gls{qe} models}
%In Section~\ref{EAMT:sec:training_steps}, we describe our methodology for training and evaluating QE models. 
During Step 1 (see Section~\ref{EAMT:sec:training_steps}), we trained an OOD \gls{qe} model and evaluated it every 1000 $steps_{HF}$\footnote{$steps_{HF}$ refers to the Hugging Face framework's training or evaluation steps, which are different from the ones we described in Section~\ref{EAMT:sec:training_steps}.} using the train and dev sets from Group 1 (see Section~\ref{EAMT:sec:data}). In Step 2, we trained and evaluated \gls{qe} mix models every 500 $steps_{HF}$ using a mix of OOD and ID data from Group 1. For Step 3, we evaluated the final domain-specific \gls{qe} model after 500 $steps_{HF}$ using only an ID train and dev set. Throughout the training, we used an early stopping mechanism to halt the training process if there was no improvement in the evaluation loss after 5 evaluations. We adjusted the default evaluation $steps_{HF}$ from 500 to 1000 for Step 1 due to the larger number of training samples in that step. This ensures efficient training by reducing evaluation overhead and stabilizing model performance checks.

%In the most realistic scenarios, tags are not available for the QE data (SRC, MTT). Because of that, we excluded tags in Step 3, although they were included in all other steps for input train and dev data.

\subsubsection{Machine Translation} 
\label{EAMT:sec:MT}
%Unlike prior studies, such as \cite{https://doi.org/10.48550/arxiv.2111.00767} on QE DAG that rely on generic/common translation model (e.g., Google machine translator), in our DAG approach we first train a separate NMT model on part of the original dataset. This ensures that the training data and the data used for translation have very close vocabularies, cover similar topics, styles and domain, which will ensure high quality translation~\cite{cite-key}.
Our approach to generating synthetic ID (Approach 2, Section~\ref{EAMT:sec:DAG}) differs from prior studies, such as~Eo et al.,~\shortcite{https://doi.org/10.48550/arxiv.2111.00767}, which rely on a generic/common translation model (e.g., Google machine translate). We first trained a separate \gls{nmt} model on a subset of the original dataset. This approach ensures that the training data and the data used for translation have similar vocabularies, cover comparable topics, styles, and domains, which leads to higher quality translations.

%In our adapted methodology, we used an in-house MT framework to train an MT model for Approach 2 (see Section~\ref{EAMT:sec:DAG}). The framework is based on a pre-trained multilingual model from HuggingFace~(HF), specifically the mBART-50 model~\cite{10.1162/tacl_a_00343}, and used the Seq2SeqTraining\footnote{\url{https://huggingface.co/docs/transformers/main_classes/trainer#transformers.Seq2SeqTrainingArguments}} arguments recommended by HF. If the evaluation loss failed to improve after 5 evaluation steps, we stopped training a model.

We used an in-house \gls{mt} framework to train our models, based on pre-trained mBART-50 \cite{10.1162/tacl_a_00343} from HF. We followed the Seq2SeqTraining
%\footnote{\url{https://huggingface.co/docs/transformers/main_classes/trainer#transformers.Seq2SeqTrainingArguments}} 
arguments recommended by HF and trained the model for Approach 2, stopping the training if the evaluation loss did not improve after 5 evaluations.

%To train our adapted methodology, we used an in-house MT framework based on the pre-trained mBART-50 model from HuggingFace~(HF). We used the Seq2SeqTraining\footnote{\url{https://huggingface.co/docs/transformers/main_classes/trainer#transformers.Seq2SeqTrainingArguments}} arguments recommended by HF and trained the model for Approach 2 (see Section~\ref{EAMT:sec:DAG}). We stopped training the model if the evaluation loss failed to improve after 5 evaluation steps.

We used default hyperparameters recommended by HF for \gls{qe} and \gls{mt}, and our frameworks with modified hyperparameters are available at~\url{https://github.com/JoyeBright/DA-QE-EAMT2023} to reproduce our results. %We used one NVIDIA Tesla V100 GPU for training all models and trained all models with mixed precision (16-bit).
All models were trained on a single NVIDIA Tesla V100 GPU using mixed precision (16-bit).

% \subsection{Evaluation}
% We evaluate QE models performance with Pearson's coefficient (0-100\%) with the quality metric (HTER/TER) on the test sets described in data Groups 1 and 3. Moreover, we use the BLEU~\cite{papineni-etal-2002-bleu} score as a metric to evaluate the translation quality of our MT models.
\section{Results}
\label{EAMT:sec:results}
To assess the performance of our approach, we evaluate output from the trained \gls{qe} models in comparison to the reference quality metric (HTER/TER) on the test sets described in data Groups 1 and 3 (see Section~\ref{EAMT:sec:data}).  We use Pearson's coefficient ($\rho \in [-1, 1]$, which we rescale to $-100$ to $100$ for clarity) to assess the correlation of our predictions with the test set. We evaluate the translation quality of our \gls{mt} models using the BLEU score. %We first present our baseline results and then the results for our models. The interpretation of the results is based on a statistical test, where a p-value of less than $0.05$ indicates significance.

\subsection{Baseline results}
\label{EAMT:sec:baseline}
To establish a baseline for our study, we fine-tuned XLM-R with the ID data for each language pair as provided by WMT 2021 shared task (Group 1––see Section~\ref{EAMT:sec:data}). This is a conventional approach employed in prior research, such as~Ranasinghe et al.~\shortcite{ranasinghe-etal-2020-transquest}, where pre-trained models are utilized to provide cross-lingual reference for training \gls{qe} models. %The results of our baselines are presented in Table~\ref{EAMT:table:main_results}.

We also attempted to compare our work with the models of Rubino~\shortcite{rubino-2020-nict} and Lee~\shortcite{lee-2020-two}. For the latter work, their experiments used the WMT 2020 test sets, while we used WMT 2021, which makes it difficult to compare our results to theirs directly. Furthermore, we could not replicate their models as no code is available (at the time of writing this chapter).  %Our baseline results are presented in Table~\ref{EAMT:table:main_results}.

%It is important to note that we attempted to compare our work with the studies of Rubino~\shortcite{rubino-2020-nict} and Lee~\shortcite{lee-2020-two}. However, we were unable to replicate their models due to the unavailability of code at the time of writing this paper. It should be noted that our objective in this study is not to outperform the results of other QE frameworks but to explore the performance of QE models with and without the proposed DA method.

\subsection{Main results}
In Table~\ref{EAMT:table:main_results} we present our results using the \gls{dag} approaches and the two training modes (TAG and NO TAG). Additional details on the statistical tests for each language pair are available in Section~\ref{EAMT:sup:ss}. The results in Table~\ref{EAMT:table:main_results} show that, in general, all of the proposed \gls{da} methods performed better than the baseline for each language pair, except for Approach 1 (\gls{dag} 1) in the RO-EN language pair. For this language pair, the use of a domain tag led to reduced performance, and the improvement achieved without such a tag was not statistically significant.

\begin{table}[t!]
\setlength\tabcolsep{7pt}
\renewcommand{\arraystretch}{1.3}
\centering
\begin{adjustbox}
{width=\textwidth,center}
\begin{tabular}{l|rrrrrrr}
\hline \hline
\multirow{2}{*}{\pbox{20cm}{\textbf{Language} \\ \textbf{pair}}} &
  \multirow{2}{*}{\textbf{Baseline}} &
  \multicolumn{2}{c}{\textbf{NO TAG}} &
  \multicolumn{2}{c}{\textbf{TAG}} &
  \multicolumn{2}{r}{\multirow{2}{*}{\textbf{Increase \%}}} \\ \cdashline{3-4}[1pt/1pt]\cdashline{5-6}[1pt/1pt]
 &
   &
  \multicolumn{1}{r}{\textbf{DAG 1}} &
  \multicolumn{1}{r}{\textbf{DAG 2}} &
  \multicolumn{1}{r}{\textbf{DAG 1}} &
  \multicolumn{1}{r}{\textbf{DAG 2}} &
  \multicolumn{2}{c}{} \\ \hline \hline
EN-DE        & 47.17 & \multicolumn{1}{r}{49.93} & 49.54 & \multicolumn{1}{r}{\textbf{51.90}}  & 51.25 & \multicolumn{2}{r}{10.03} \\ 
EN-ZH        & 29.16 & \multicolumn{1}{r}{34.75} & 35.27 & \multicolumn{1}{r}{35.62} & \textbf{36.60}  & \multicolumn{2}{r}{25.51} \\ 
RO-EN        & 83.63 & \multicolumn{1}{r}{83.67} & 83.74 & \multicolumn{1}{r}{83.37} & \textbf{84.40}  & \multicolumn{2}{r}{00.92}  \\ 
RU-EN        & 40.65 & \multicolumn{1}{r}{44.91} & 45.40 & \multicolumn{1}{r}{\textbf{47.16}} & 43.98 & \multicolumn{2}{r}{16.01}  \\ \hline \hline
\end{tabular}
\end{adjustbox}
\caption{\textbf{Pearson correlation scores for proposed \gls{qe} models across 4 language pairs}: EN-DE, EN-ZH, RO-EN, and RU-EN. For each language pair, the bold result indicates the highest-performing method compared to the baseline. Results for the first and second \gls{dag} approaches are reported under \gls{dag} 1 and \gls{dag} 2, respectively. The column labeled ``Increase \%" shows the percentage improvement for the highest-performing model (in bold) compared to the baseline.}

\label{EAMT:table:main_results}
\end{table}

We also observe that the increase of performance compared to the baseline for each language pair shown as percentage in the last column of Table~\ref{EAMT:table:main_results} is substantial, except for RO-EN (only 0.92\% increase over the baseline). This is mainly due to the already high baseline performance (83.63), making it challenging to achieve significant improvements. Among the other language pairs, the EN-ZH pair had the largest increase in performance –– just over 25\%. The RU-EN and EN-DE pairs had the second and third highest increases, with improvements of around 16\% and 10\% over their respective baselines. %It is worth pointing out that the results show that the increase percentage correlates to the baseline performance -- the lower the baseline results the higher the increase.

\paragraph{Additional indication of domain results.}
The results indicate that incorporating tags into the \gls{da} training pipeline was generally effective, although in some instances, the improvement was not statistically significant compared to the models that were trained without tags. However, it was observed that at least one model outperformed the same language pair's models that were not trained with tags, when \gls{dag} techniques were used. Specifically, the EN-DE Approach 1 model trained with tags performed better compared to Approach 2 without tags, as did the EN-ZH Approach 1 model trained with tags relative to the same approach without tags. Finally, the RO-EN Approach 2 model trained with tags outperformed Approach 2 without tags, and the RU-EN Approach 1 model trained with tags exhibited better performance than Approach 1 without tags. %These results suggest that the use of tags is beneficial in DA for QE, but the extent of their effectiveness may vary depending on the language pair and the specific DAG approach used. 

\vspace*{-1mm}
\subsection{Data augmentation results}
Upon analyzing the integration of \gls{dag} techniques into the specialized \gls{qe} pipeline, we observe that for most language pairs, both approaches showed better performance than their respective baselines. However, in situations where tags were not employed, Approach 2 only showed statistical significance over Approach 1 in the EN-ZH and RU-EN language pairs. Moreover, when tags were used, Approach 2 led to statistically significant improvements only for EN-DE and EN-ZH. %These findings suggest that the choice of DAG approach and the use of tags should be carefully considered when applying DA in QE. Additionally, DAG was observed to be significant for EN-ZH, for both cases --- with or without tags. 
These findings indicate that the effectiveness of \gls{dag} approaches and tag usage can vary depending on language pair and task context.

Our findings suggest that users should evaluate both \gls{dag} approaches with and without tags on a validation set specific to their language pairs. Tags are recommended for languages with complex syntax or morphology (e.g., EN-DE), whereas omitting tags may be preferable when no performance gains are observed (e.g., RU-EN). For language pairs like EN-ZH, where \gls{dag} consistently shows improvements, \gls{dag} should be applied regardless of tag usage.

\subsection{Zero-shot results}
In order to evaluate the effectiveness of our \gls{qe} models in the context of \gls{zsl}, we compared their performance to the baseline models for the EN-CZ and EN-JA test sets. The results are presented in Table~\ref{EAMT:tbl:zsl}.

The findings show that, for the EN-CZ test set, the \gls{qe} model trained solely on the EN-DE dataset achieved the highest performance among all \gls{qe} baselines, with a Pearson correlation score of 46.97. Additionally, we observe that our proposed \gls{da} pipeline performed even better than the highest-performing baseline for EN-CZ, but only \gls{dag} Approach 1 and 2 with tags were found to be statistically significant. Likewise, for the EN-JA test set, the highest-performing \gls{qe} baseline was the one that was trained solely on the RU-EN dataset, with a Pearson correlation score of 20.32. In contrast to EN-CZ, none of the models that were trained with our pipeline and with the RU-EN dataset outperformed the baselines. Nevertheless, we observed that three models trained with EN-ZH and using our pipeline (Approach 1 with and without tag, and Approach 2 with tag) performed better than the highest-performing baseline.

Overall, these findings suggest that if a \gls{qe} model is conventionally trained with and evaluated on an unseen \gls{qe} dataset, some extent of \gls{zsl} capabilities can be achieved due to the use of XLM-R. However, the proposed \gls{da} pipeline can significantly increase this extent, whether through models trained with the same dataset or other datasets used in the pipeline. %Furthermore, we observed that training a QE model conventionally using certain language pairs may lead to decreased performance. For instance, a model trained exclusively with the EN-DE language pair showed a Pearson correlation of approximately 10. In such cases, the proposed pipeline may enhance performance even when using the same training data. 

\begin{table}[h]
\setlength\tabcolsep{8pt}
\renewcommand{\arraystretch}{1.2}
\centering
\begin{adjustbox}
{width=\textwidth,center}
\begin{tabular}{l|lrrrrr}
\hline\hline
\multirow{2}{*}{\begin{tabular}[c]{@{}l@{}}\textbf{Trained} \\ \textbf{on}\end{tabular}} &
  \multirow{2}{*}{\textbf{Test set}} &
  \multirow{2}{*}{\textbf{Baseline}} &
  \multicolumn{2}{c}{\textbf{NO   TAG}} &
  \multicolumn{2}{c}{\textbf{TAG}} \\ \cdashline{4-7}[1pt/1pt]
                       &       &       & \textbf{DAG 1} & \textbf{DAG 2} & \textbf{DAG 1} & \textbf{DAG 2} \\ \hline\hline
\multirow{2}{*}{EN-DE} & EN-CZ & 46.97 & 48.77 & 48.07 & 47.78 & 47.82 \\
                       & EN-JA & 09.67  & 18.16 & 08.00 & 16.12 & 17.36 \\ \hdashline[1pt/1pt]
\multirow{2}{*}{EN-ZH} & EN-CZ & 35.56 & 49.33 & 48.54 & 47.98 & 46.83 \\
                       & EN-JA & 13.13 & 22.77 & 19.87 & 22.24 & 21.54 \\ \hdashline[1pt/1pt]
\multirow{2}{*}{RO-EN} & EN-CZ & 26.33 & 39.10 & 39.79 & 39.20 & 40.41 \\
                       & EN-JA & 18.88 & 20.34 & 18.55 & 20.11 & 21.22 \\ \hdashline[1pt/1pt]
\multirow{2}{*}{RU-EN} & EN-CZ & 28.42 & 45.58 & 44.85 & 46.43 & 45.22 \\
                       & EN-JA & 20.32 & 17.64 & 17.04 & 17.26 & 19.63 \\ \hline\hline
\end{tabular}
\end{adjustbox}
\caption{\textbf{Performance comparison of the proposed methods and the baseline} 
using Pearson correlation. 
All models were trained on the EN-DE, EN-ZH, RO-EN, and RU-EN datasets and 
evaluated in a \gls{zsl} setting on the EN-CZ and EN-JA test sets. 
Results for the two \gls{dag} strategies are presented under \gls{dag}\,1 
(authentic ID augmentation) and \gls{dag}\,2 (synthetic ID augmentation).}
\label{EAMT:tbl:zsl}
\end{table}

\section{Additional observations}
\label{EAMT:sec:discussion}
\subsection{Cross-lingual inference}
%Furthermore, we observe that the second DAG approach, which was trained with twice the amount of data used in approach 1, resulted in a shorter training time.

Table~\ref{EAMT:table:cross-lingual} presents data that shows that our proposed methodology has an overall advantage over the conventional training method of using a \gls{plm} and fine-tuning it with \gls{qe} data (baselines) in terms of cross-lingual inference. That is, the \gls{qe} models trained with our proposed \gls{da} pipeline not only perform significantly better than baselines on their target domain and language pair but can also estimate the quality of other language pairs to some extent better than their corresponding baseline.

By examining the data closely (bottom to top row of the Table~\ref{EAMT:table:cross-lingual}), we observe that XLM-R provides a limited level of cross-lingual inference, which is insufficient for estimating quality labels due to the absence of prior knowledge about them. However, using Step 1 of our pipeline, which utilizes little inference knowledge, the model achieves an acceptable level of generalization across all language pairs. 

Specifically, Step 1 achieved an average Pearson correlation score of approximately 39, which is higher than all baseline scores by 3 points, except for the RO-EN pair, which achieved around 42. Furthermore, the model trained using Step 1 of the pipeline achieved a Pearson correlation of around 70 when evaluated with the RO-EN test set. This result can be attributed to the training of the model with the Italian (IT) dataset, used as OOD data. %From a linguistic point of view, this result could be explained by the fact that IT and RO belong to the same language family, i.e., the ``romance languages'' (refer to Appendix~\ref{EAMT:appendix:genetic}), which explains the high Pearson correlation score achieved by the model.
The use of Italian data likely provided the model with exposure to linguistic features and patterns that closely align with those found in Romanian (RO), thereby enhancing its ability to make accurate quality estimations for the RO-EN pair. From a linguistic perspective, both Italian and Romanian are part of the ``Romance language family'' (refer to Section~\ref{EAMT:sup:genetic}), which share significant similarities in terms of syntax, morphology, and lexical structure. These shared characteristics, such as similar word formation rules, verb conjugations, and sentence structures, can facilitate cross-lingual generalization. As a result, the model was better equipped to handle Romanian data despite not being explicitly trained on it, leading to a significantly higher Pearson correlation score on the RO-EN test set. This phenomenon highlights the importance of leveraging linguistically related languages for improved cross-lingual model performance in quality estimation tasks.

As we move up the table, we can observe that the model built in Step 2 of our pipeline becomes more specific toward the task and the ID datasets. Consequently, there is an average improvement of around 3.5 Pearson correlation (from 39.36 to 42.83) across the languages.
The improvement can be attributed to the model's enhanced ability to leverage \gls{da}, where exposure to both OOD and ID data helps the model better generalize to in-domain tasks.
%This indicates that our DA pipeline is effective in improving more specific cross-lingual QE performance. Ultimately, fine-tuning Step 2 with any of the ID languages provides a highly domain-specific QE model that is not only better estimates the quality of their language pair, but also performs better cross-lingual inference over its baseline.
By fine-tuning with ID data, the model adapts more closely to the characteristics of the specific language pairs, resulting in better task-specific performance. This targeted adaptation strengthens the model’s capacity to evaluate translation quality for its language pair, leading to improved performance on both in-domain and cross-lingual tasks. Consequently, models fine-tuned in Step 2 not only provide more precise estimates of translation quality within their specific language pair but also exhibit superior cross-lingual inference, outperforming the baseline models. This demonstrates the efficacy of our \gls{da} pipeline in enhancing task-specific and cross-lingual \gls{qe} performance.

\begin{table}[!ht]
\renewcommand{\arraystretch}{1.1}
\centering
\begin{adjustbox}
{width=0.95\textwidth,center}
\begin{tabular}{l|rrrr|r}
\hline\hline
  \multirow{2}{*}{\textbf{Models}} &
  \multicolumn{4}{c|}{\textbf{Test Sets}} &
  \multirow{2}{*}{\textbf{AVG}}
  \\ \cdashline{2-5}[1pt/1pt]%\cline{2-5}
                                          & \textbf{EN-DE} & \textbf{EN-ZH} & \textbf{RO-EN} & \textbf{RU-EN} & \\ \hline\hline
\cellcolor[HTML]{FFFF00}Baseline & \underline{47.17}          & 19.67          & 44.96          & 32.91          & 36.17             \\
\cellcolor[HTML]{FFFF00}EN-DE    & \underline{49.93}          & 22.66          & 78.97          & 39.55          & 47.77             \\
\textbf{$\Delta$}                         & \underline{02.76}           & 02.99           & 34.01          & 06.64           & \textbf{11.60}       \\ \hdashline[1pt/1pt]
\cellcolor[HTML]{FFE699}Baseline & 30.34          & \underline{29.16}          & 47.55          & 36.87          & 35.98            \\
\cellcolor[HTML]{FFE699}EN-ZH    & 43.46          & \underline{34.75}          & 80.51          & 42.67          & 50.34             \\
\textbf{$\Delta$}                         & 13.12          & \underline{05.59}           & 32.96          & 05.80            & \textbf{14.36}      \\ \hdashline[1pt/1pt]
\cellcolor[HTML]{A9D08E}Baseline & 24.64          & 23.56          & \underline{83.63}          & 39.97          & 42.95              \\
\cellcolor[HTML]{A9D08E}RO-EN    & 43.02          & 24.31          & \underline{83.67}          & 38.74          & 47.43             \\
\textbf{$\Delta$}                         & 18.38          & 00.75           & \underline{00.04}           & -01.23          & \textbf{04.48}        \\ \hdashline[1pt/1pt]
\cellcolor[HTML]{F4B084}Baseline & 22.40           & 24.67          & 57.17          & \underline{40.69}          & 36.23              \\
\cellcolor[HTML]{F4B084}RU-EN    & 25.36          & 26.06          & 75.34          & \underline{44.91}          & 42.91             \\
\textbf{$\Delta$}                         & 02.96           & 01.39           & 18.17          & \underline{04.22}           & \textbf{06.68}        \\ \hdashline[1pt/1pt]
Step2                            & 38.29          & 24.72          & 76.96          & 31.35          & 42.83 
\\  \hdashline[1pt/1pt]
Step1                            & 30.80           & 16.57          & 70.14          & 39.93          & 39.36     
 \\  \hdashline[1pt/1pt]
XLM-R  &-02.74	&07.30	&02.97	&03.12	&02.66  
\\
\hline \hline
\end{tabular}
\end{adjustbox}
\caption{\textbf{Performance comparison of proposed models and baselines across all test sets} using Pearson correlation as the metric. $\Delta$ represents the difference between them. The ``AVG'' column reports the mean Pearson correlation across all four language pairs (EN-DE, EN-ZH, RO-EN, RU-EN), and the differences ($\Delta$) are computed with respect to these averaged values.
``XLM-R (not trained)'' refers to the raw XLM-R encoder used without any QE fine-tuning, included as a lower-bound reference. 
Step 1: model trained with OOD. 
Step 2: model trained with \gls{dag} approach 1 and OOD (Approach 2 in Step 2 had similar results, not included). 
Models and baselines are color-coded for clarity, with bold numbers indicating the average $\Delta$ across all language pairs, and underlined numbers representing each model's performance on their respective test sets.}
\label{EAMT:table:cross-lingual}
\end{table}

\subsection{OOD Performance}
The main goals of \gls{da} are to quickly create an adapted system and to develop a system that performs well on ID test data while minimizing performance degradation on a general domain. In our study, we showed that models from Step 1 or Step 2 can be fine-tuned quickly using the user's data (achieving the first of these goals). Our main focus was on the assessment of ID \gls{qe}. Here, we test the performance of our ID models on an OOD test set in order to assess the degree of degradation. Our results, summarized in Table~\ref{EAMT:table:OOD_performance}, indicate that not only did all ID models outperform the corresponding baselines on the OOD test set, but that incorporating ID data in Approaches 1 and 2 did not compromise the performance with respect to OOD. However, comparing the models' performance with models trained solely on OOD, we see a small yet statistically significant performance drop, which is inevitable and, in most cases, acceptable.
\begin{table}[ht]
\setlength\tabcolsep{6pt}
\renewcommand{\arraystretch}{1.55}
\centering
\begin{adjustbox}{max width=\textwidth,center}
\begin{tabular}{l|rrrr|r|rr}
\hline\hline
\multirow{2}{*}{\textbf{Training setup}} &
  \multicolumn{4}{c|}{\textbf{ID Models (Step 3)}} &
  \textbf{OOD-only} &
  \textbf{DAG\,1} &
  \textbf{DAG\,2} \\
& \textbf{EN-DE} & \textbf{EN-ZH} & \textbf{RO-EN} & \textbf{RU-EN} & \textbf{EN-IT} & \textbf{(OOD+ID)} & \textbf{(OOD+SynthID)} \\
\hline\hline

\textbf{Baseline} 
& 11.95 & 3.59 & 11.60 & 3.43 
& \multirow{4}{*}{64.33} 
& \multirow{4}{*}{65.24}
& \multirow{4}{*}{64.76} \\

\textbf{Our pipeline} 
& 54.62 & 59.30 & 52.51 & 47.36 
&  &  &  \\

\cdashline{1-5}[1pt/1pt]

$\boldsymbol{\Delta_{\text{Baseline}}}$ 
& 42.67 & 55.71 & 40.91 & 43.93 
&  &  &  \\

$\boldsymbol{\Delta_{\text{OOD}}}$ 
& \textbf{-09.71} & \textbf{-05.03} & \textbf{-11.82} & \textbf{-16.97} 
&  &  &  \\

\hline\hline
\end{tabular}
\end{adjustbox}

\caption{\textbf{Performance on the OOD test set} (Pearson correlation).  
``OOD-only'' is the model trained exclusively on the out-of-domain EN-IT data (Step~1).  
``DAG\,1'' denotes Step~2 fine-tuning using all authentic ID data concatenated across language pairs;  
``DAG\,2'' uses synthetic ID data generated via Approach~2 (Section~\ref{EAMT:sec:DAG}).  
$\Delta_{\text{Baseline}}$ compares each ID model with its corresponding baseline, and  
$\Delta_{\text{OOD}}$ compares each ID model with the OOD-only model.
Note that OOD-only, DAG~1, and DAG~2 do not vary across language pairs and therefore appear once for clarity.}
\label{EAMT:table:OOD_performance}
\end{table}

\subsection{Training time}
\begin{figure}[t]
    \centering
    \includegraphics[width=0.7\textwidth]{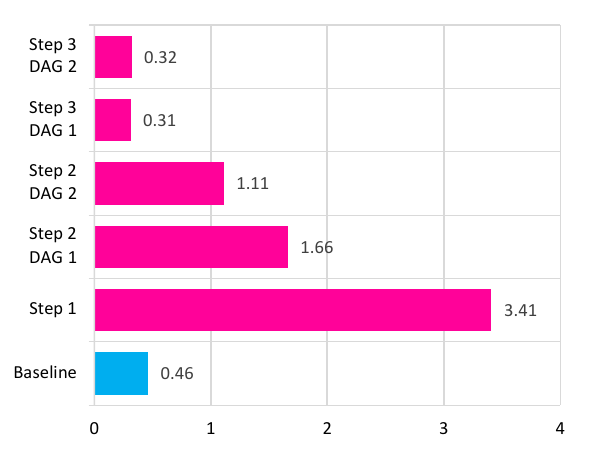}
    \caption{\textbf{Training time (in hours) for models in the EN-ZH language pair}, where Step X refers to the training step outlined in Section~\ref{EAMT:sec:training_steps}, and \gls{dag} X denotes the \gls{dag} approach used in the second step of the pipeline. The term ``Baseline'' denotes a model fine-tuned from XLM-R. The X and Y axes represent the training time in hours and the approaches used to train the model, respectively.}
    \label{EAMT:fig:training_time}
\end{figure}

Compared to the conventional approach of using a \gls{plm} and fine-tuning it with \gls{qe} data (baselines), our proposed \gls{da} methodology results in a significant improvement in performance, regardless of whether we include tags in the sentences or not. However, it requires two additional training steps: Step 1, training an OOD \gls{qe} model, and Step 2, fine-tuning the model using a mix of OOD and ID \gls{qe} data. These additional steps require more time. Step 1 and Step 2 (with both \gls{dag} approaches) are reused (i.e., not trained) for each language pair, and Step 3 of the pipeline took almost the same amount of time across all languages. That is why we present the consumed time for EN-ZH in Figure~\ref{EAMT:fig:training_time}, and use it to discuss training times for other language pairs as well. Models trained with tagged data have a similar training time.

The data presented in Figure~\ref{EAMT:fig:training_time} indicates that Step 1 has the highest training time of approximately 3.4 hours. It is noteworthy that this long training time is partly due to the fact that the model was evaluated after every 1000 $steps_{HF}$, which consequently resulted in a longer running time in comparison to other models that were evaluated after every 500 $steps_{HF}$. The model that was trained is publicly accessible, and other individuals can utilize it to fine-tune with new ID datasets, avoiding the need for retraining for each specific ID data. This applies to both \gls{dag} approaches, given that the target language pair was used in Step 2 of the pipeline. If not, Step 1 must be fine-tuned with a new set of \gls{qe} data.

\section{Related work}
\label{EAMT:sec:related_works}
% \subsection{Machine Translation Quality Estimation} Initially, QE systems were developed as statistical, regression models (only sentence- and document-level QE) trained on linguistic features of the source and MTT~\cite{specia-etal-2013-quest,specia-etal-2015-multi}. With neural approaches taking over the field of Natural Language Processing, including MT and QE, more sophisticated models appeared. These models were mostly based on LSTMs~\cite{10.1162/neco.1997.9.8.1735} at first (e.g., DeepQuest by Ive et al.,~\shortcite{ive-etal-2018-deepquest}) and later on Transformer~\cite{NIPS2017_3f5ee243} (e.g., QEBrain by Wang et al.,~\shortcite{wang-etal-2018-alibaba}). The advent of the Transformer also led to the recent availability of LLMs such as BERT~\cite{devlin-etal-2019-bert} and XLM-R, which have become fundamental to QE. 

% A number of QE systems utilize such models~\cite{ranasinghe-etal-2020-transquest,HAN2021225} to build latent representations of source/target sentences or words, or for feature extraction, used to calculate a quality estimate. Despite their impressive performance, all QE systems, including those employing LLMs like MTransQuest~\cite{ranasinghe-etal-2020-transquest}, neural architectures like Predictor-Estimator~\cite{kim-etal-2017-predictor}, and models like SiameseQE~\cite{shterionov-etal-2019-less}, require QE data for training~\cite{zheng-etal-2021-self}. However, such data is scarce and expensive to obtain.

\subsection{Data scarcity in \gls{qe}} 
The issue of data scarcity in \gls{mt} \gls{qe} has been explored in numerous previous studies. The work of \shortciteA{rubino-sumita-2020-intermediate} involves the use of pre-training sentence encoders and an intermediate self-supervised learning step to enhance \gls{qe} performances at both the sentence and word levels. This approach aims to facilitate a smooth transition between pre-training and fine-tuning for the \gls{qe} task. %Furthermore, the authors claim that their proposed method (intermediate step) does not require the use of annotated data. %The authors of this study evaluated the effects of domain adaptation (one-step fine-tuning) on the pre-trained model and developed a self-supervised learning approach for intermediate training.
Similarly, \shortciteA{fomicheva-etal-2020-unsupervised} proposed an unsupervised method for \gls{qe} that does not depend on additional resources and obtains valuable data from \gls{mt} systems. %They also developed a new multilingual dataset for QE that includes low- and high-resource language pairs.

\shortciteA{2022arXiv221210257Q} conducted a recent study on the impact of various types of parallel data in \gls{qe} \gls{dag}, and put forward a classifier to differentiate the parallel corpora. Their research revealed a significant discrepancy between the parallel data and real \gls{qe} data, as the most common \gls{qe} \gls{dag} technique involves using the target side of parallel data as the reference translation~\cite{baek-etal-2020-patquest,2022arXiv221210257Q}, followed by translation of the source side using an \gls{mt} model, and ultimately generating pseudo \gls{qe} labels~\cite{freitag-etal-2021-results}. %However, our study diverges from this conventional approach and concentrates on a straightforward yet effective DAG methods to mitigate this gap. 
\shortciteA{kocyigit-etal-2022-better} proposed a negative \gls{dag} technique to improve the robustness of their \gls{qe} models. Their method involves training a sentence embedding model to reduce the search space for \gls{qe} predictions by using contrastive loss. This approach improves the model's capacity to differentiate between high- and low-quality translations by maximizing the distance between dissimilar sentence pairs in the embedding space, thereby refining the overall \gls{qe} performance.

\subsection{Domain adaptation in \gls{qe}} 
To tackle the challenges with translating data when training data comes from diverse domains, researchers have extensively used \gls{da} in \gls{mt}. \gls{da} involves training a large generic model and then fine-tuning it with domain-specific data~\cite{chu-wang-2018-survey,saunders2022domain,2021arXiv211206096P,pham-etal-2022-multi} and Chapter~\ref{chap:CLIN}. In \gls{mt}, one way to achieve \gls{da} is by appending tags to sentences to handle different domains~\cite{sennrich-etal-2016-controlling,mattoni-etal-2017-zero,vanmassenhove-etal-2018-getting,Chu2019MultilingualMA} and reduce catastrophic forgetting. 

Despite being useful in \gls{mt}, to the best of our knowledge, \gls{da} has not been widely used in \gls{qe}. Dongjun Lee~\shortcite{lee-2020-two} proposed a two-step \gls{qe} training process similar to our own, and Raphael Rubino~\shortcite{rubino-2020-nict} pre-trained XLM and further adapted it to the target domain through intermediate training. Both studies demonstrated that adding a step before fine-tuning improves performance compared to fine-tuning alone. However, unlike our methodology, neither of them included sentence tags or conducted additional fine-tuning (such as Step 3 in our methodology). As a result, their \gls{qe} models are not as specialized for the target domain as ours.
A few researchers have made attempts to integrate aspects of \gls{da} into \gls{qe}. For instance, in an effort to improve \gls{qe} performance in domain-specific scenarios, Arda Tezcan~\shortcite{Tezcan_2022} included fuzzy matches into MonoTransQuest with the aid of XLM-RoBERTa model and \gls{dag} techniques. %Another research that tried to apply DA in QE is~\cite{de-souza-etal-2014-towards}, which analyzed multitask and online learning methods for MT QE across various domains. However, their approach does not fully utilize the benefits of DA as it was developed before successful DA approaches.

\section{Conclusion and future work}
\label{EAMT:sec:conclusion}
This chapter addresses two key challenges related to \gls{qe} of \gls{mt}: (i) the scarcity of \gls{qe} data, which is mitigated by utilizing pseudo-\gls{qe} data followed by fine-tuning on authentic data, and (ii) the difficulty of estimating translations across diverse domains, addressed through domain-specific fine-tuning using in-domain~(ID) data. The primary aim of this study is to enhance the performance of \gls{qe} models by addressing these challenges. To do so, we propose a solution that utilizes \gls{da} techniques adopted from \gls{mt}. We adapt the ``mixed fine-tuning + fine-tuning'' approach~\cite{chu-etal-2017-empirical} and extend it with \gls{dag} as an alternative to the traditional oversampling technique. We adopt a three-step training methodology: (i) we fine-tune XLM-R, a language model, with a large generic \gls{qe} dataset, which enables the model to generalize; (ii) we fine-tune the model with a mix of out-of-domain~(OOD) and ID data derived from two \gls{dag} approaches; and (iii) we fine-tune the model with a small amount of domain-specific data, which leads to a more specific model. We evaluated models' performance with and without domain tags appended to the sentences. As seen in previous studies, sentence tags help the model differentiate between various domains by providing contextual cues. Our findings partially confirm this effect, indicating that domain tags contribute to improved domain-specific performance.

Our experiments show significant improvements across all language pairs under consideration, indicating that our proposed solution has a beneficial impact in addressing the aforementioned challenges. Our study also demonstrates the effectiveness of both proposed \gls{dag} approaches and shows that using domain tags improves the performance of the models. Additionally, we find that our model outperforms the baseline in the context of \gls{zsl} and in cross-lingual inference. % We also discussed the superiority of our proposed approach in terms of cross-lingual inference. %The models that we trained using our proposed methodology are publicly accessible, enabling anyone to implement each step and fine-tune them using their own data.

Moving forward, there are several directions for future work based on our findings. First, it would be interesting to investigate the performance of our pipeline on low-resource language pairs, where there is limited ID data available. This is particularly relevant given the smaller coverage of \gls{qe} datasets compared to parallel data in \gls{mt}. Second, we only used one type of OOD data in our experiments (EN-IT); it would be useful to explore other OOD data over different language pairs for \gls{qe}. Third, it would be valuable to study the performance of other \glspl{plm} and not only XLM-R. Fourth, since the choice of languages employed in the pipeline was based on availability, we would suggest exploring a more regulated approach for selecting the languages to be used in the proposed pipeline. Specifically, the optimal transfer languages can be selected based on their data-specific features, such as dataset size, word overlap, and \gls{sw} overlap, or dataset-independent factors, such as genetic (see Section~\ref{EAMT:sup:genetic}) and syntactic distance~\cite{lin-etal-2019-choosing}. 

\vspace{1cm}

\section{Supplemental material}
\label{sec:amta:supplementary}
This section provides additional information that complements the findings discussed in the main chapter.

\subsection{Statistically significance test results}
\label{EAMT:sup:ss}

The statistical significance test results for the predictions in Table~\ref{EAMT:table:main_results} for the language pairs EN-DE, EN-ZH, RO-EN, and RU-EN are shown in Table~\ref{EAMT:tab:ss_all}.

\begin{table}[!ht]
\setlength\tabcolsep{5pt}
\renewcommand{\arraystretch}{1.2}
\centering
\begin{adjustbox}{width=0.85\textwidth,center}
\begin{tabular}{l|l|cccc}
\hline\hline
\begin{tabular}[c]{@{}l@{}}\textbf{Language}   \\ \textbf{pair}\end{tabular} & \multicolumn{1}{l|}{\textbf{Models}} & \textbf{NO TAG 1} & \textbf{NO TAG 2} & \textbf{TAG 1} & \textbf{TAG 2} \\ \hline\hline
\multirow{4}{*}{EN-DE} & Baseline & Y & Y & Y & Y \\
                                & NO TAG 1 & - & N & N & Y \\
                                & NO TAG 2 & - & - & Y & Y \\
                                & TAG 1    & - & - & - & Y \\ \hdashline[1pt/1pt]
\multirow{4}{*}{EN-ZH} & Baseline & Y & Y & Y & Y \\
                                & NO TAG 1 & - & Y & Y & N \\
                                & NO TAG 2 & - & - & N & N \\
                                & TAG 1    & - & - & - & Y \\ \hdashline[1pt/1pt]
\multirow{4}{*}{RO-EN} & Baseline & N & Y & Y & Y \\
                                & NO TAG 1 & - & N & Y & Y \\
                                & NO TAG 2 & - & - & N & N \\
                                & TAG 1    & - & - & - & N \\ \hdashline[1pt/1pt]
\multirow{4}{*}{RU-EN} & Baseline & Y & Y & Y & Y \\
                                & NO TAG 1 & - & Y & Y & Y \\
                                & NO TAG 2 & - & - & N & Y \\
                                & TAG 1    & - & - & - & N \\ \hline\hline
\end{tabular}
\end{adjustbox}
\caption{\textbf{Statistically significant test results} with a p-value less than 0.05. The letter ``Y" in the table indicates that the corresponding prediction in Table~\ref{EAMT:table:main_results} is statistically significant, while ``N" indicates that it is not.}
\label{EAMT:tab:ss_all}
\end{table}

% \subsection{Training Steps}
% In Figure~\ref{fig:training_Pipeline}, we present an overview of the proposed training steps for specialized QE.
% \label{appendix:training_steps}
% \begin{figure}[h]
%     \centering
%      \includegraphics[keepaspectratio,width=\columnwidth]{Training_pipeline.3.pdf}
%     \caption{\textbf{Overview of the proposed training steps for specialized QE.} The ``+'' sign indicates the oversampling performed in Step 2 to balance the use of ID and OOD data. The dashed arrows indicate the source of the checkpoint used to initialize the models in each stage.}
%     \label{fig:training_Pipeline}
% \end{figure}

% \subsection{Data Augmentation: Approach 2}
% Figure~\ref{EAMT:fig:approach2} presents an overview of Approach 2 that is employed for data augmentation in the context of domain adaptation for QE.
% \label{EAMT:appendix:approach2}
% \begin{figure}[htbp]
%     \centering
%     \includegraphics[keepaspectratio,width=\columnwidth]{Assets/EAMT/Approach2.6.pdf}
%     \caption{\textbf{Overview of Approach2 (Generating synthetic ID) of data augmentation for domain adaptation in QE.} The various steps involved in the approach are indicated close to the corresponding arrows. Arrow 1 represents subsampling. The abbreviations $SRC$, $TRG$, and $T_{lp}$ stand for source, target, and machine-translated text, respectively. The final outputs which include $SRC$, $T_{lp}$ and quality labels ($TER$) are color-coded for clarity.}
%     \label{EAMT:fig:approach2}
% \end{figure}

\subsection{Machine translation performance}
We utilized multilingual \gls{mt} systems to generate synthetic ID data. Table~\ref{EAMT:tab:mt_DAG2} displays the results of the top-performing models used in generating this data.

\begin{table}[t!]
\setlength\tabcolsep{7pt}
\renewcommand{\arraystretch}{1.2}
\centering
\begin{adjustbox}{width=0.65\textwidth,center}
    \begin{tabular}{l|rr}
        \hline\hline
        \textbf{Language pair} & \textbf{BLEU~$\uparrow$} & \textbf{Eval Loss~$\downarrow$} \\ \hline\hline
        EN-DE & 41.25 & 01.09 \\
        EN-ZH & 32.28 & 01.52 \\
        RO-EN & 49.60 & 00.96 \\
        RU-EN & 41.29 & 01.61 \\  \hline\hline
    \end{tabular}
\end{adjustbox}
    \caption{\textbf{\gls{mt} performance used as a component of Approach 2} in the proposed DAG~(Section~\ref{EAMT:sec:DAG}).}
    \label{EAMT:tab:mt_DAG2}
\end{table}

\subsection{Genetic distance}
\label{EAMT:sup:genetic}

\begin{figure}[!ht]
    \centering
    \includegraphics[width=0.9\textwidth]{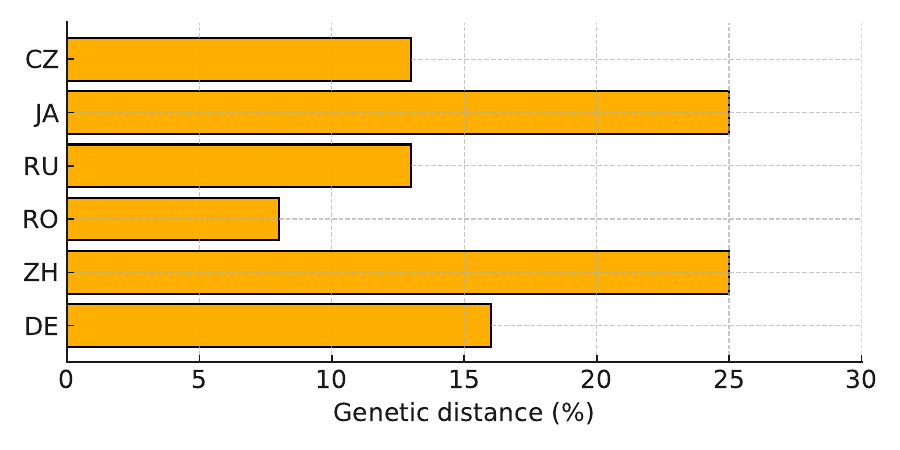}
    \caption{\textbf{Genetic distance between IT and other languages:} DE, ZH, RO, RU, JA, and CZ.}
    \label{EAMT:fig:genetic}
\end{figure}
In \gls{mt}, measuring the similarity between languages is important for effective cross-lingual learning. One such measure is the \emph{genetic distance} between languages, which has been shown to be a good indicator of language similarity for independent data~\cite{lin-etal-2019-choosing}. To illustrate this, we calculate\footnote{\url{http://www.elinguistics.net/Compare_Languages.aspx}} and present the genetic distance scores between Italian (used as OOD data) and the other languages included in our study in Figure~\ref{EAMT:fig:genetic}. The genetic distance is represented as a numerical value ranging from 0 (indicating the same language) to 100 (the greatest possible distance).

\chapterseparatorpage
\chapterimage{Assets/separator.pdf}

\prechaptertext{This chapter is based on the following published paper:
\\\\
\hspace*{0.5em}Javad Pourmostafa Roshan Sharami, Dimitar Shterionov, and Pieter Spronck. 
\href{https://acl-bg.org/proceedings/2025/RANLP\%202025/pdf/2025.ranlp-1.111.pdf}
{Analysis of Vocabulary and Subword Tokenization Settings for Optimal Fine-tuning of MT: A Case Study of In-domain Translation}. In \textit{Proceedings of Recent Advances in Natural Language Processing (RANLP 2025)}, pp. 970–979, Varna, September 8–10, 2025.
\\\\
}
{Improvements have been made to the title, figures, and certain sections to align the content with the broader context of this dissertation.
}

\chapter[Analysis of Vocabulary and Subword Tokenization Settings for Optimal Fine-tuning of Machine Translation]{Analysis of Vocabulary and Subword Tokenization Settings for Optimal Fine-tuning of \gls{mt}}

\chaptermark{BPE}
\label{chap:BPE}

\lettrine[lines=4]{T}{he} choice of vocabulary and \gls{sw} tokenization has a significant impact on both training and fine-tuning of language and translation models. Fine-tuning is a common practice in optimizing a model with respect to new data. However, new data potentially introduces new words (or tokens), which, if not considered, may lead to suboptimal performance. In addition, the distribution of tokens in the new data can differ from the distribution of the original data. As such, the original \gls{sw} tokenization model could be less suitable for the new data. With this work, we aim to gain better insights on the impact of \gls{sw} tokenization and vocabulary generation on the performance of \gls{nmt} models fine-tuned to a specific domain. To do so, we compare several strategies for \gls{sw} tokenization and vocabulary generation and investigate the performance of the resulting models.

Our findings show that the best way to fine-tune for \gls{da} is to consistently use both BPE and vocabulary from the in-domain data, which helps the model pick up on important domain-specific terms. At the same time, it is crucial not to lose sight of the vocabulary of the base (pre-trained) model—maintaining coverage of this vocabulary ensures the model keeps its general language abilities. The most successful configurations are those that introduce plenty of frequent domain terms while still retaining a substantial portion of the base model vocabulary, leading to noticeably better translation quality and adaptation, as seen in higher BLEU scores. These benefits, however, often come with greater computational costs, such as longer training times, since the model must learn more new tokens. Conversely, approaches that skip important domain terms or combine mismatched tokenization and vocabulary do not perform as well, making it clear that both domain-specific adaptation and broad vocabulary coverage matter—and that these gains are realized when the vocabulary preserves a good portion of the base (pre-trained) model.

While using in-domain BPE and vocabulary yields the best \gls{da}, it substantially reduces out-of-domain translation quality. Hybrid configurations that combine base and domain vocabularies help balance this trade-off, maintaining broader translation capabilities alongside improved domain performance.
\newpage

\section{Introduction and background}\label{BPE:introduction}
Fine-tuning is a common practice in optimizing MT and \gls{plm} with respect to new data. It is often in the context of DA where an existing model is tuned to perform better on a specific domain (different from what the model was originally trained for)~\cite{luong-etal-2015-effective,dakw:17fine,70c446486deb46eba84b4e2d06c5f963,mahdieh2020rapid,Shivang_Chopra_2023}. The positive effect of fine-tuning has been demonstrated in various previous works. For example, \citeA{luong-etal-2015-effective} trained an NMT model on English-German general-domain data and then fine-tuned it on a conversational data in the same languages, leading to an increase of 3.8 BLEU~\cite{Papineni2002} points compared to the original model. \citeA{sharami2022selecting} show that fine-tuning is preferred (as it leads to better results) than training from scratch, even if the data allows the latter. To improve the translation performance on a new domain (without degrading the performance on the generic domain) is to ensemble the fine-tuned model with the already trained baseline, as done by~\citeA{freitag2016fast}. However, while they achieve a substantial increase of quality (+7.2 BLEU points ), they note that because the in-domain data comprises of new vocabulary and linguistic features that are different from the generic data, the performance of the fine-tuned models drops for the generic domain task, especially when it comes to domain-specific contexts (e.g., medical and legal domains).

While newly introduced data brings in new information, i.e., new, unseen words, it could be that, statistically, tokenization into \glspl{sw} is significantly different from the tokenization of the original model~\cite{lim2018exploring,essay80128,sato-etal-2020-vocabulary,Dongjun_Hwang_2024}. If not properly addressed, this new information may have an adverse effect on the system.

To address this problem, \citeA{sato-etal-2020-vocabulary} proposed a method to adapt the embedding layers of the initial model to the target domain by projecting the general word embedding obtained from target-domain monolingual data onto source-domain embeddings. They reported a 3.86 and 3.28 BLEU points gain in English$\rightarrow$Japanese and German$\rightarrow$English translation, respectively.

In this chapter, we investigate the impact of using different \glspl{sw} and vocabularies on the performance of fine-tuned NMT systems. We identify a best-case setup and preferable setups under constrained fine-tuning conditions, such as limited domain-specific data. That is, we aim to investigate which fine-tuning conditions (or settings) of a domain-specific model lead to the best performance. Specifically, our objectives are:
\begin{enumerate}
\item Identify optimal \gls{sw} combination choices and vocabulary configurations for a given MT model and fine-tuning dataset.  

\item Determine the best fine-tuning conditions under data limitations.

\end{enumerate}

To achieve the aforementioned objectives, we use one large dataset (\(\sim\)12.7 million parallel sentences) for training and a smaller in-domain dataset (\(\sim\)248,000 parallel sentences) for fine-tuning multiple MT systems. This setup allows us to examine the extent to which a model trained on a substantial amount of general-domain data can be improved by fine-tuning with additional domain-specific data, which alone would be insufficient to train a robust model from scratch.

Each fine-tuned alternative, is trained on a different set of options of how the \glspl{sw} and the vocabulary are created. We analyze these fine-tuning strategies to find the best setup based on available data. In our case study, for example, we have access to the data of both models (initial and fine-tuned). However, as already discussed in~\cite{freitag2016fast,dakw:17fine,Eduardo_Zimelewicz_2024}, initial models are mostly deployed in an application; thus data might not be available at the production time. As such, it is paramount to have a guideline based on the available data that determines how to best generate \glspl{sw} and vocabularies. In this work, we use Byte-Pair Encoding (BPE)~\cite{sennrich-etal-2016-controlling} for \gls{sw} units.\footnote{Throughout this chapter, we use ``subword'' (SW) and ``BPE'' synonymously, as all experiments use BPE for \gls{sw} tokenization.}

It is noteworthy that the point of this research is to investigate the best fine-tuning setup, rather than identifying the best model. Typically, fine-tuning involves tokenizing the new data using the vocabulary originally employed in training the model. This ensures consistency in \gls{sw} tokenization and prevents discrepancies in word representations. However, it is not always evident whether this practice yields the best translation performance, especially when the new domain introduces a significantly different linguistic distribution or unseen vocabulary. Thus, we explore alternative approaches to \gls{sw} tokenization and vocabulary creation to determine if different configurations could lead to better fine-tuning outcomes. Specifically, given a pre-training dataset A—whether in-domain, out-of-domain, synthetic, e.g., generated using methods like those in \cite{sharami-etal-2023-python}, or authentic—and a fine-tuning dataset B, we investigate which tokenization and vocabulary configurations best enable the model to retain and adapt pre-training-derived parameters in a way that improves translation quality on B.

This chapter is organized as follows. Section~\ref{BPE:DPs} presents the key decision points that frame our study. Section~\ref{BPE:data} describes the datasets used in our experiments. Section~\ref{BPE:experiments} details our experimental design and the overall training framework. Section~\ref{BPE:results} reports and analyzes the results, incorporating relevant recommendations. In Section~\ref{BPE:limitations}, we discuss the limitations of our approach. Finally, Section~\ref{BPE:conclusion} summarizes our findings and outlines directions for future work.

\section{Decision points}
\label{BPE:DPs}

% As noted in Section~\ref{BPE:introduction}, with this work, we aim to identify the most effective way of fine-tuning an MT system with respect to \emph{SW segmentation} and \emph{vocabulary creation}. 

Given a model \( M \) trained on a dataset \( D \), which represents a specific domain \( d \), and a fine-tuning dataset \( E \) representative for domain \( e \), the following decision points need to be made:

\subsection{SW tokenization}
A key decision in fine-tuning is determining how to tokenize words into SW units. This choice affects how the model processes domain-specific terminology and generalizes across datasets. We consider three approaches:
\begin{itemize}
    \item \textbf{Reusing the original \gls{sw} model} trained on \( D \) (\( D_{\text{SW}} \)). This maintains consistency with the pre-trained model.  
    \item \textbf{Training a \gls{sw} model on the combined dataset} (\( (D+E)_{\text{SW}} \)) to integrate both the original and fine-tuning data. 
    \item \textbf{Training a new \gls{sw} model on the fine-tuning dataset} (\( E_{\text{SW}} \)) to better capture domain-specific terminology.  
\end{itemize}

\subsection{Vocabulary creation}
\gls{sw} tokenization techniques arose in response to two major challenges in NMT:  
(i) the lack of generalizability—models often fail to process words not seen during training, leading to out-of-vocabulary (OOV) problems and degraded performance; and  
(ii) the need to limit vocabulary size, as large vocabularies increase memory consumption and computational cost, which remains a practical constraint in current neural translation systems, particularly when working with large models or limited GPU resources.

Fine-tuning introduces a third, less often addressed challenge: whether the vocabulary used during adaptation adequately captures the token distribution of the fine-tuning dataset. If not, domain-specific content may be poorly represented, limiting the effectiveness of adaptation.

We consider three strategies for vocabulary construction:

\begin{itemize}
    \item \textbf{Reusing the original vocabulary} — the vocabulary that the pre-trained model \( M \) was originally trained with (denoted \( |D| \)). This strategy ensures full compatibility with the pre-trained token embeddings and does not require any modifications to the embedding space.
    
    \item \textbf{Expanding the vocabulary} — augmenting the original vocabulary with additional tokens found in the fine-tuning dataset \( E \), resulting in a combined vocabulary \( |D+E| \). This approach aims to better cover domain-specific terms in \( E \) while retaining compatibility with \( M \)'s original vocabulary.
    
    \item \textbf{Constructing a new vocabulary solely from the fine-tuning data} — generating the vocabulary exclusively from \( E \) (denoted \( |E| \)). This strategy maximizes domain-specific representational capacity but introduces a mismatch with the pre-trained vocabulary of \( M \).
\end{itemize} 

\vspace{1em}
\noindent \textbf{Handling vocabulary-embedding alignment.}  
In the first strategy (\( |D| \)), the embedding space remains unchanged, as all tokens are already present in the pre-trained model.

In the second and third strategies (\( |D+E| \) and \( |E| \)), we introduce new tokens absent from the original vocabulary. To accommodate these, we extend the embedding matrix by appending randomly initialized vectors for the new tokens while preserving the original embeddings.

The key distinction lies in the degree of divergence from the original model. Strategy 2 retains the original vocabulary and extends it with tokens from \( E \), maintaining alignment with the pre-trained structure. In contrast, Strategy 3 derives both the vocabulary and BPE model entirely from \( E \), resulting in a larger mismatch with the pre-trained model and necessitating greater adaptation during fine-tuning.

\bigskip

Since \gls{sw} tokenization and vocabulary creation are interdependent, we explore all feasible combinations, resulting in nine configurations. These include applying each \gls{sw} model (\( D_{\text{SW}}, E_{\text{SW}}, (D+E)_{\text{SW}} \)) with different vocabulary choices (\( |D|, |D+E|, |E| \)).

Following these decision points, given a fine-tuning dataset, we can consider three \gls{sw} models. With these models, we (i) \textit{tokenize the vocabulary sources}, and (ii) \textit{tokenize the training sets for fine-tuning}. Typically, these two processes are tied to each other, i.e., once the \gls{sw} model is learned and applied to the training data, the vocabulary is the set of \gls{sw} units that appear in the (processed) data. However, this is not a hard constraint. 

For instance, dataset \( E \) can be processed with \( E_{\text{SW}} \), but the vocabulary used for training can still be based on \( D \) and derived from applying \( D_{\text{SW}} \). Such mismatched configurations, though theoretically possible, can lead to tokenization inconsistencies and degrade model performance. Since they are suboptimal, we exclude them from this study.

\section{Data}
\label{BPE:data}

We used two datasets: (i) a large \textit{out-of-domain} corpus consisting of approximately 12.7 million English-German sentence pairs drawn from the WMT18 dataset\footnote{\url{http://statmt.org/wmt18/translation-task.html}. The original WMT18 dataset is considerably larger; we selected a representative subset to reduce computational costs. The selected subset and all preprocessing scripts are released for reproducibility.}, and (ii) a smaller (\(\sim\)248K sentence pairs) \textit{in-domain medical} corpus extracted from the multi-domain English-German data introduced by~\cite{koehn-knowles-2017-six}.

\paragraph{Out-of-domain dataset}
\label{generic_data}
The out-of-domain corpus used to train our base model is a randomly selected subset of the WMT18 English-German dataset, which contains parallel data from various domains. We selected approximately 12.7 million sentence pairs to balance domain coverage with training efficiency.

\paragraph{In-domain dataset}
\label{in_domain_data}
For fine-tuning, we used 248{,}099 English-German sentence pairs from the medical domain of the multi-domain dataset introduced by~\cite{koehn-knowles-2017-six}. We used the cleaned and re-split version provided by~\cite{aharoni-goldberg-2020-unsupervised}, which removes duplicates and prevents data leakage between train, dev, and test sets.

\paragraph{Combined dataset (\( D+E \))}  
For configurations requiring both \( D \) and \( E \), we oversampled the in-domain medical data to match the size of the WMT18 subset and concatenated them. The combined data was shuffled and used to train BPE models or extract vocabularies. This ensures that both domains are equally represented, avoiding bias toward the larger out-of-domain corpus.

\section{Experiments}
\label{BPE:experiments}
To investigate the impact of \gls{sw} and vocabulary generation choices on fine-tuning, we followed the decision points outlined in Section~\ref{BPE:DPs} and ran experiments using the English-German data described in Section~\ref{BPE:data}. We compared the resulting fine-tuned models using BLEU~\cite{Papineni2002}, TER~\cite{snover-etal-2006-study}, chrF2~\cite{popovic-2015-chrf}. Additionally, we measured training time and estimated CO\textsubscript{2} emissions using \texttt{CodeCarbon}~\cite{benoit_courty_2024_11171501}.

\subsection{Experimental design}

Given our task—fine-tuning a model trained on the WMT18 out-of-domain dataset (\( D \)) using in-domain medical data (\( E \))—a total of 9 theoretical configurations exist, arising from three possible vocabulary sources (\( D \), \( E \), or \( D+E \)) and three BPE models trained on the same sources. Each configuration couples one vocabulary source with one BPE model, which is used for both vocabulary construction and fine-tuning data tokenization.

However, as discussed in Section~\ref{BPE:DPs}, we imposed constraints to ensure tokenization consistency. Specifically, we excluded configurations where the vocabulary is derived from one source (\( D \) or \( E \)), but the fine-tuning tokenization is performed using a BPE model trained on the combined dataset (\( D+E \)). These mismatches introduce inconsistencies, as the vocabulary may not align with how the data is tokenized. Since both \( D \) and \( E \) are available, we ensure that the same BPE model is used for both vocabulary construction and fine-tuning tokenization.

In total, there are 9 possible configurations (from all combinations of BPE models and vocabulary sources). However, we exclude 2 inconsistent configurations, leaving 7 valid configurations for our experiments (see Table~\ref{tbl:configurations}).

\begin{table}[ht]
\centering
\renewcommand{\arraystretch}{1.2}
\resizebox{0.85\textwidth}{!}{%
\begin{tabular}{l|l|l}
\hline\hline
\textbf{Config.} & \textbf{BPE for vocab + FT data} & \textbf{Vocabulary source} \\
\hline\hline
C1  & \( D_{\text{BPE}} \)       & \( D \)       \\
C2  & \( D_{\text{BPE}} \)       & \( D+E \)     \\
C3  & \( D_{\text{BPE}} \)       & \( E \)       \\
C4  & \( E_{\text{BPE}} \)       & \( D \)       \\
C5  & \( E_{\text{BPE}} \)       & \( D+E \)     \\
C6  & \( E_{\text{BPE}} \)       & \( E \)       \\
C7  & \( (D+E)_{\text{BPE}} \)   & \( D+E \)     \\
\hline\hline
\end{tabular}%
}
\caption{\textbf{Valid fine-tuning configurations.} Each row represents a consistent setup where the same BPE model is used for both tokenizing the fine-tuning data and constructing the vocabulary. \( D \) refers to the WMT18 out-of-domain dataset; \( E \) refers to the in-domain medical dataset.}
\label{tbl:configurations}
\end{table}

\subsection{Model architecture and training Setup}
\label{BPE:nmt_system}
\paragraph{Framework and model architecture}
We used the OpenNMT-py\footnote{\url{https://opennmt.net/OpenNMT-py/}} framework~\cite{klein-etal-2017-opennmt} to train and fine-tune Transformer-based NMT models~\cite{vaswani2017attention}. Each model had 6 encoder and 6 decoder layers, 512-dimensional embeddings, 8 attention heads, and a feed-forward size of 2048. We used the Noam optimizer schedule with a learning rate of 2.0, 8{,}000 warmup steps, and label smoothing of 0.1. Batching was done over 10{,}240 tokens with gradient accumulation over 4 steps.

\paragraph{Training setup}
All models were trained for up to 200{,}000 steps, with validation and checkpointing every 1{,}000 steps. We applied early stopping after 10 validations without improvement. All experiments—including the base and fine-tuned models—were run on a single NVIDIA A40 GPU.

\paragraph{Base model}
The base model was trained on the WMT18 out-of-domain dataset. We applied BPE with 50K merge operations to both source and target sides. The resulting vocabularies and tokenized data were used to train the initial Transformer model, which served as the starting point for all fine-tuning experiments.

\paragraph{Fine-tuning}
Fine-tuning was done on the in-domain medical dataset using the same model architecture and training settings. Each configuration (C1–C7) used a specific combination of vocabulary source and BPE model (see Table~\ref{tbl:configurations}). The base model checkpoint was reused across all configurations, and only the vocabulary and tokenized data differed. BLEU was used to track validation performance.

\paragraph{BPE settings}
We trained separate BPE models for the source and target sides. The number of merge operations depended on dataset size: 8K merges for corpora with fewer than 100K lines, 30K for those between 100K and 1M, and 50K for larger ones. This choice is supported by prior work, which shows that smaller vocabularies benefit Transformer models~\cite{kudo-2018-subword}, and that 2K–8K merges perform best for low-resource datasets~\cite{adlaon-marcos-2024-finding}. Our BPE models were used consistently for both vocabulary construction and fine-tuning data tokenization.

\section{Results and analysis}
\label{BPE:results}
In this section, we present the evaluation results and statistical comparisons of our fine-tuning setups. We also explore vocabulary overlaps to understand how token and vocabulary choices impact performance and adaptation.

\begin{table*}[t]
\setlength\tabcolsep{4pt} 
\renewcommand{\arraystretch}{1.3}
\centering
\begin{adjustbox}{width=\textwidth,center}
\begin{tabular}{c|l|l|r|r|r|r|r}
\hline\hline
\textbf{Config} & 
\textbf{BPE model}& 
\textbf{Vocab SRC} & 
\textbf{BLEU$\uparrow$} & 
\textbf{chrF2$\uparrow$} & 
\textbf{TER$\downarrow$} & 
\textbf{CO\textsubscript{2} (g)$\downarrow$} & 
\textbf{Time (h)$\downarrow$} \\
\hline\hline
C1 & \( D_{\text{BPE}} \)       & \( D \)       & 53.6 & 69.4 & 49.3 & 1658.69 & 07:45 \\
C2 & \( D_{\text{BPE}} \)       & \( D+E \)     & 53.4 & 69.5 & 49.9 & 1198.66 & 05:15 \\
C3 & \( D_{\text{BPE}} \)       & \( E \)       & 51.7 & 68.4 & 50.9 & 907.24  & 04:00 \\
C4 & \( E_{\text{BPE}} \)       & \( D \)       & 46.6 & 64.5 & 53.0 & 723.94  & 03:11 \\
C5 & \( E_{\text{BPE}} \)       & \( D+E \)     & 53.1 & 68.9 & 49.7 & 729.04  & 03:15 \\
C6 & \( E_{\text{BPE}} \)       & \( E \)       & \textbf{54.8} & \textbf{69.8} & \textbf{48.9} & 1587.41 & 09:30 \\
C7 & \( (D+E)_{\text{BPE}} \)   & \( D+E \)     & 53.2 & 69.1 & 50.1 & 543.84  & 03:08 \\
\hline\hline
\end{tabular}
\end{adjustbox}
\caption{\textbf{Evaluation scores of fine-tuned models.} \textbf{Note:} For each configuration, the BPE model shown is used to tokenize both the vocabulary source and the fine-tuning data. All models were fine-tuned on the in-domain dataset \( E \) and evaluated on the same test set. CO\textsubscript{2} emissions and training times were recorded during fine-tuning.}
\label{tbl:result}
\end{table*}

\subsection{Analysis of fine-tuning results}
\label{BPE:analysis}

Table~\ref{tbl:result} summarizes the performance of all fine-tuning configurations. To better interpret these results, we performed pairwise bootstrap tests on BLEU scores (Section~\ref{BPE:significance}), interpreting \(p\)-values as a continuous measure of confidence without enforcing a strict threshold. TER and chrF2 metrics supplemented the analysis to refine the ranking.

We ranked configurations using the following criteria:
\begin{enumerate}
    \item BLEU scores, weighted by the strength of statistical evidence from \(p\)-values.
    \item TER to resolve ties or unclear BLEU differences.
    \item chrF2 as a final tiebreaker if both BLEU and TER were inconclusive.
\end{enumerate}

Accordingly, the ranking from best to worst is:
\[
C6 \succ C1 \succ C5 \succ C2 \succ C7 \succ C3 \succ C4,
\]
where \( \succ \) denotes a configuration that performs better or more reliably than the next.

\paragraph{Top configuration (\(C6\)).}  
Configuration \(C6\) uses both BPE and vocabulary exclusively from the in-domain data \(E\), resulting in the highest BLEU and best TER and chrF2 scores. Statistical tests show that \(C6\) significantly outperforms all other configurations, confirming the advantage of aligning tokenization and vocabulary strictly with the fine-tuning domain.

\paragraph{Strong middle tier (\(C1\), \(C5\), \(C2\)).}  
Configurations \(C1\), \(C5\), and \(C2\) achieve similar BLEU scores, with statistical evidence showing no clear superiority among them. \(C1\) (BPE and vocabulary from out-of-domain \(D\)) slightly leads numerically, while \(C5\) (in-domain BPE, combined vocabulary) offers better TER than \(C2\) (out-of-domain BPE, combined vocabulary), which justifies the order. These results suggest incorporating some in-domain vocabulary or combining datasets can yield competitive results if full in-domain access (for both BPE and vocabulary) is not possible.

\paragraph{Lower performing configurations (\(C7\), \(C3\), and \(C4\)).}  
Configurations \(C7\) and \(C3\) perform moderately but are consistently behind the mid-tier cluster. Configuration \(C4\) ranks last, likely because of a mismatch between its in-domain BPE and out-of-domain vocabulary, which impairs tokenization and reduces fine-tuning effectiveness.

\paragraph{Practical recommendations.}  
For optimal fine-tuning, use both BPE and vocabulary consistently derived from the in-domain data, as exemplified by configuration \(C6\). When full access to in-domain data or vocabulary is limited—due to privacy, proprietary constraints, or resource availability—fine-tuning remains possible but may yield reduced adaptation effectiveness. In such cases, configurations like \(C1\) and \(C2\) offer robust alternatives by leveraging available data while balancing performance and practicality. It is important to avoid mixing BPE and vocabulary from mismatched domains, as this often leads to suboptimal tokenization and degraded translation quality. Overall, aligning tokenization and vocabulary with domain data maximizes fine-tuning benefits, but adapting with limited data can still provide meaningful improvements compared to no adaptation.

\subsubsection{BLEU score statistical comparison}
\label{BPE:significance}

We conducted pairwise bootstrap tests on BLEU scores using 1,000 iterations. Table~\ref{tbl:ss_bleu} shows the \( p \)-values for all configuration pairs. Diagonal entries represent self-comparisons.

\begin{table}[ht]
\centering
\begin{adjustbox}{width=0.85\columnwidth,center}
\begin{tabular}{l|*{7}{c}}
\hline\hline
 & \textbf{C1} & \textbf{C2} & \textbf{C3} & \textbf{C4} & \textbf{C5} & \textbf{C6} & \textbf{C7} \\
\hline
\textbf{C1} & -- & 0.545 & 0.000 & 0.000 & 0.200 & 0.852 & 0.102 \\
\textbf{C2} & 0.447 & -- & 0.000 & 0.000 & 0.172 & 0.792 & 0.063 \\
\textbf{C3} & 1.000 & 1.000 & -- & 0.000 & 0.995 & 1.000 & 0.989 \\
\textbf{C4} & 1.000 & 1.000 & 1.000 & -- & 1.000 & 1.000 & 1.000 \\
\textbf{C5} & 0.774 & 0.813 & 0.006 & 0.000 & -- & 0.966 & 0.286 \\
\textbf{C6} & 0.149 & 0.174 & 0.000 & 0.000 & 0.026 & -- & 0.012 \\
\textbf{C7} & 0.896 & 0.929 & 0.010 & 0.000 & 0.701 & 0.987 & -- \\
\hline\hline
\end{tabular}
\end{adjustbox}
\caption{\textbf{Pairwise bootstrap \( p \)-values for BLEU scores} (1,000 iterations). Diagonal entries represent self-comparisons. \textit{All values are provided for reference only; no statistical significance threshold is applied.}}
\label{tbl:ss_bleu}
\end{table}

\subsection{Training time and CO\textsubscript{2} emissions}
The training times and estimated CO\textsubscript{2} emissions in Table~\ref{tbl:result} show the resource demands of each fine-tuning setup. The best-performing configuration, \(C6\), took the longest—about 9.5 hours—and had a higher carbon footprint, likely because it had to learn many new domain-specific tokens. 

Other configurations such as \(C7\), \(C4\), and \(C5\) completed training notably faster, around three hours, and had lower CO\textsubscript{2} emissions compared to \(C6\). This can be attributed to several factors. Configurations \(C7\) and \(C5\) utilize vocabularies with higher overlap to the baseline tokens, meaning fewer new domain-specific tokens need to be learned, which reduces training complexity and time. On the other hand, \(C4\) exhibits both a vocabulary and BPE mismatch, which limits effective fine-tuning and results in quicker but less effective training. In summary, configurations with less vocabulary adaptation or mismatched tokenization require less training time and energy but tend to yield lower translation quality.

Overall, while domain-aligned fine-tuning boosts performance, it can require more time and energy—something to consider in real-world applications.

\subsection{Vocabulary overlap analysis}
\label{BPE:vocab_overlap}

To better understand how vocabulary choice influences fine-tuning, we measured overlap between each configuration’s vocabulary and the baseline WMT vocabulary in terms of token frequency coverage. This approach more accurately reflects the practical impact of commonly used tokens on model performance, as it weights tokens by how often they occur in the data. Table~\ref{tbl:vocab_overlap} summarizes the key statistics:

\begin{itemize}
    \item \textbf{SRC/TGT Overlap (\%)}: The percentage of total token frequency (i.e., the sum of token counts) in the baseline WMT vocabulary that is also present in the configuration’s vocabulary, calculated for source (English) and target (German) separately. This reflects not just the number of shared tokens, but their practical frequency in baseline data.
    \item \textbf{New Tokens}: The number of tokens in the configuration’s vocabulary that do not appear in the baseline vocabulary (after filtering), representing domain-specific or new tokens introduced by the configuration.
\end{itemize}

\begin{table*}[t]
\setlength\tabcolsep{4pt}
\renewcommand{\arraystretch}{1.5}
\centering
\small
\begin{adjustbox}{width=\textwidth,center}
\begin{tabular}{l|l|l|r|r|r|r|r}
\hline\hline
\textbf{Config} & 
\textbf{BPE Model} & 
\textbf{\makecell[t]{Vocab\\Src}} & 
\textbf{BLEU$\uparrow$} & 
\textbf{\makecell[t]{SRC\\Overlap \%}} & 
\textbf{\makecell[t]{TGT\\Overlap \%}} & 
\textbf{\makecell[t]{New\\SRC}} & 
\textbf{\makecell[t]{New\\TGT}} \\
\hline\hline
C6 & \( E_{\text{BPE}} \)       & \( E \)           & 54.8 & 82.84 & 77.81 & 13,022 & 13,559 \\
C1 & \( D_{\text{BPE}} \)       & \( D \)           & 53.6 & 100   & 100   & 0      & 0      \\
C5 & \( E_{\text{BPE}} \)       & \( D+E \)         & 53.1 & 83.04 & 77.95 & 14,289 & 14,157 \\
C2 & \( D_{\text{BPE}} \)       & \( D+E \)         & 53.4 & 83.04 & 77.95 & 11,736 & 11,804 \\
C7 & \( (D+E)_{\text{BPE}} \)   & \( D+E \)         & 53.2 & 97.61 & 95.72 & 14,300 & 15,077 \\
C3 & \( D_{\text{BPE}} \)       & \( E \)           & 51.7 & 83.04 & 77.95 & 11,736 & 11,804 \\
C4 & \( E_{\text{BPE}} \)       & \( D \)           & 46.6 & 90.70 & 90.46 & 0      & 0      \\
\hline\hline
\end{tabular}
\end{adjustbox}
\caption{\textbf{Vocabulary overlap and BLEU scores per configuration.} Configurations are listed in order of their overall performance ranking. SRC and TGT overlap percentages indicate the proportion of baseline tokens retained. New Tokens columns count tokens unique to the configuration's vocabulary. \textbf{Note:} The BPE model in each configuration is applied to both the vocabulary source and the fine-tuning data.}
\label{tbl:vocab_overlap}
\end{table*}

The results demonstrate that configurations incorporating in-domain vocabulary (e.g., \(C6\), \(C5\), and \(C7\)) introduce a substantial number of new domain-specific tokens, which is associated with their superior BLEU scores and more effective \gls{da}. In contrast, \(C1\), relying solely on the baseline vocabulary, achieves complete overlap but lacks critical domain-specific terms, limiting its adaptability. The notably poor performance of \(C4\) corresponds with its lower vocabulary overlap and the evident mismatch between its BPE model and vocabulary source, underscoring the detrimental impact of inconsistent tokenization strategies.

These findings robustly support our practical recommendation: for optimal fine-tuning, vocabulary and BPE should be consistently derived from the same in-domain data. Such alignment ensures richer domain-specific token representation, ultimately leading to enhanced translation accuracy and better overall model performance.

\subsection{Out-of-domain performance analysis}
\label{BPE:ood_analysis}

To quantify the impact of different fine-tuning strategies on generalization, we evaluated all configurations on the original out-of-domain (WMT18, $D$) test set. Table~\ref{tbl:ood_performance} reports BLEU scores for each configuration, alongside the absolute and relative drop with respect to the pre-trained base model (before fine-tuning).

\begin{table}[ht]
\centering
\renewcommand{\arraystretch}{1.2}
\resizebox{0.9\columnwidth}{!}{%
\begin{tabular}{l|l|r|r|r}
\hline\hline
\textbf{Config} & \textbf{BPE/Vocab Source} & \textbf{BLEU} & \textbf{Drop} & \textbf{Drop (\%)} \\
\hline
Base & $D_{\text{BPE}}, D$ (pre-trained) & 33.9 & --   & --   \\
C2 & $D_{\text{BPE}}, D+E$  & 15.1 & $-18.8$ & $-55.5$ \\
C7 & $(D+E)_{\text{BPE}}, D+E$ & 15.0 & $-18.9$ & $-55.8$ \\
C3 & $D_{\text{BPE}}, E$    & 13.3 & $-20.6$ & $-60.8$ \\
C1 & $D_{\text{BPE}}, D$    & 13.1 & $-20.8$ & $-61.4$ \\
C4 & $E_{\text{BPE}}, D$    & 10.2 & $-23.7$ & $-69.9$ \\
C6 & $E_{\text{BPE}}, E$    & 7.7  & $-26.2$ & $-77.3$ \\
C5 & $E_{\text{BPE}}, D+E$  & 7.0  & $-26.9$ & $-79.4$ \\
\hline\hline
\end{tabular}
}
\caption{\textbf{Out-of-domain BLEU scores.} Performance on the WMT18 ($D$) test set for all configurations, ranked by smallest drop relative to the pre-trained base model.}
\label{tbl:ood_performance}
\end{table}

The results illustrate the trade-off introduced by \gls{da}: as the model is adapted to the in-domain data, out-of-domain performance drops substantially across all configurations. This degradation is most pronounced when both BPE and vocabulary are derived solely from in-domain data (C5, C6), indicating strong domain specialization at the expense of generalization.

For practitioners seeking to balance \gls{da} and general translation quality, we recommend hybrid configurations such as $C_2$ and $C_7$, which use either the original BPE with a combined vocabulary or a combined BPE and vocabulary. These setups moderate the drop in out-of-domain BLEU, preserving more general-domain competence while still offering improved \gls{da}.

\section{Limitations}
\label{BPE:limitations}

Despite the systematic and thorough analysis, we acknowledge several drawbacks and limitations of our work. Addressing these in the future would complement this research and expand the understanding of the impact of data processing on model performance.

\begin{itemize}
    \item \textbf{Focus on MT:} Our evaluation focused on NMT systems. However, neural language models are also impacted by how the training and fine-tuning data is processed and used, as well as the limitations placed on the vocabulary. This is even more pertinent with the progress in \glspl{llm}. We did not analyze the performance of \glspl{llm}, which is a more complex task, especially in the case of multi-lingual \glspl{llm} capable of translation.

    \item \textbf{Fine-tuning data:} Our study focused exclusively on the medical domain for fine-tuning. Future research could consider additional specialized domains to evaluate the generalizability of the findings.

    \item \textbf{Use of BPE only:} We employed BPE only and did not consider other methods such as SentencePiece \cite{kudo-richardson-2018-sentencepiece} or LMVR \cite{Ataman2017}. This was a deliberate choice, as it was up to us which model and method to use during training and fine-tuning.

    \item \textbf{Hyperparameters:} We used the model's default hyperparameters and did not perform hyperparameter optimization or tuning. This was not necessary as we aimed to compare the impact of the \gls{sw} algorithms under the same conditions. However, we acknowledge that fine-tuning hyperparameters would impact the performance of original and fine-tuned models, and we hypothesize a correlation with how the vocabulary is constructed.
\end{itemize}

\section{Conclusion and future work}
\label{BPE:conclusion}

In this work, we presented a systematic analysis of vocabulary and \gls{sw} tokenization settings for fine-tuning \gls{nmt} models, using a large out-of-domain corpus (WMT18) and a specialized in-domain medical dataset as a case study. By comparing seven realistic fine-tuning setups that varied in BPE tokenization and vocabulary generation, we identified clear practical guidelines for \gls{da}.

Our results show that the most effective fine-tuning is achieved when both BPE and vocabulary are derived from the in-domain data, allowing the model to better capture frequent and relevant domain-specific terms. At the same time, we find that maintaining a substantial overlap with the vocabulary of the base model (originally trained on out-of-domain data) is essential for preserving general language coverage and ensuring stable adaptation. The best-performing configurations in our experiments balanced these two needs: they introduced many new, high-frequency in-domain tokens while still retaining a good portion of the base model vocabulary. However, this approach tends to require more computational resources, such as increased training time and higher energy consumption, due to the need for the model to learn and integrate more new tokens.

It is important to note that while maximizing \gls{da} can significantly boost in-domain performance, it may lead to a substantial drop in out-of-domain translation quality. Hybrid configurations that combine base and domain vocabularies help balance this trade-off, preserving broader translation capabilities while still delivering improved domain performance.

If, in addition to the in-domain data, the original out-of-domain data or its BPE/vocabulary are also accessible, combining these resources can help preserve general language coverage and stabilize adaptation. In all cases, our findings highlight the importance of aligning both BPE and vocabulary with the domain of the adaptation data, while retaining overlap with the base model’s vocabulary to ensure generalization.

For future work, we plan to extend our evaluation to other domains and language pairs, and to investigate how these findings generalize to \glspl{llm} and multilingual systems. We are also interested in exploring adaptive methods for selecting which tokens to retain or introduce during fine-tuning, with the aim of optimizing both performance and computational efficiency.

All datasets, models, and scripts for this chapter are publicly available at \url{https://github.com/JoyeBright/subword-ft-guide}.

\chapterseparatorpage
\chapterimage{Assets/separator.pdf}

\prechaptertext{This chapter is based on the following published paper:
\\\\
\hspace*{0.5em}Javad Pourmostafa Roshan Sharami, Dimitar Shterionov, and Pieter Spronck. 2024. \href{https://aclanthology.org/2024.amta-research.9/}{Guiding In-Context Learning of LLMs through Quality Estimation for Machine Translation}. In Proceedings of the 16th Conference of the Association for Machine Translation in the Americas (Volume 1: Research Track), pages 9–20, Chicago, the USA.
\\\\
}
{Improvements have been made to the title, figures, and certain sections to align the content with the broader context of this dissertation.}

\chapter[In-Context Learning of LLMs through Quality Estimation for Machine Translation]{In-Context Learning of LLMs through Quality Estimation for Machine Translation}
\chaptermark{AMTA}
\label{chap:AMTA}

\lettrine[lines=4]{T}{he} emergence of LLMs marks a significant shift in how we approach language tasks, including \gls{mt}. Unlike traditional \gls{mt} systems, LLMs such as GPT or XGLM possess strong \gls{zs} and few-shot capabilities. Instead of retraining, they can adapt to new tasks or domains through prompt-based learning, leveraging a handful of carefully selected \glspl{ice} via \gls{icl}. This paradigm introduces a more flexible, scalable, and data-efficient alternative to conventional supervised training. However, the quality of output from LLMs––particularly in \gls{mt}––is closely tied to the quality of \glspl{ice} provided along with the query, i.e., the text to translate. The effectiveness of these \glspl{ice} is influenced by various factors, such as the domain of the source text, the order in which the \glspl{ice} are presented, the number of these examples, and the prompt templates used. Naturally, selecting the most impactful \glspl{ice} depends on understanding how these affect the resulting translation quality.%which ultimately relies on translation references or human judgment. 
Assessing translation quality remains a complex task. While various automatic and semi-automatic metrics exist to support evaluation, ultimate judgment often relies on human perception. In the context of \gls{mt}, translation quality is typically measured against reference translations––such as those found in test sets or test suites––but such resources are not always readily available. In these cases, QE becomes the only viable alternative.

This chapter presents a methodology for \gls{icl} that relies on a search algorithm guided by domain-specific \gls{qe}. Leveraging the XGLM model, our methodology estimates the resulting translation quality without the need for translation references, selecting effective \glspl{ice} for \gls{mt} to maximize translation quality. % Our study also includes a comparison between fine-tuning a pre-trained language model (PLM) tailored to MT (mBART-50) and the proposed ICL approach. The study results demonstrate a significant performance enhancement compared to existing ICL methods in the literature. Additionally, our experiments show that while fine-tuning mBART-50 incurs significant computational costs and results in a notably lower BLEU score compared to our proposed ICL approach, it leads to improved contextual translation performance.
% Our results demonstrate significant improvements over existing ICL methods and higher translation performance compared to fine-tuning mBART-50, a pre-trained language model (PLM) with demonstrated performance for the MT task. This was achieved while remaining computationally efficient. %The script for running our experiments is publicly available at \url{anynomous.com}.
Our findings demonstrate that this approach not only outperforms existing \gls{icl} strategies and conventional fine-tuning but also yields translations that are more closely aligned with human references, ultimately reducing the need for post-editing.
\newpage

\section{Introduction}\label{AMTA:sec:introduction}
Pre-trained \glspl{llm} quickly gained popularity (and continue to do so) due to their performance on a large set of \gls{nlp} tasks, including \gls{mt}~\cite{zhu2023multilingual,xu2024paradigm}. However, the accuracy of their outputs is significantly influenced by the quality of the \emph{in-context} examples provided to them~\cite{10.1162/tacl_a_00324,alves-etal-2023-steering}.\footnote{For simplicity, we sometimes refer to it as ``example(s)'' throughout this paper.}
%The accuracy of outputs generated by pre-trained large language models~(LLMs) in Machine Translation~(MT) and other natural language processing (NLP) tasks~\cite{zhu2023multilingual,xu2024paradigm}, is significantly influenced by the quality of the examples provided to them, also known as in-context examples (ICEs)\footnote{For simplicity, we sometimes refer to it as ``example(s)'' throughout this paper.}~\cite{10.1162/tacl_a_00324}.
If these examples %are not appropriately selected –– that is, if they 
do not align well with the specific task and source domain, the LLMs' outputs can be inaccurate.\footnote{This has been observed in models such as GPT-J and LLaMA2, which rely heavily on the quality and relevance of in-context examples to generate accurate results~\cite{peng-etal-2024-revisiting}.} Therefore, there is a critical need to develop (better) methods for selecting appropriate examples that match the task and source domain being translated. These methods collectively fall under the umbrella of \gls{icl}~\cite{liu-etal-2022-makes}. %This is particularly critical, given the limited number of ICEs that can be inputted into the LLMs~\cite{agrawal-etal-2023-context}.

% LLMs have achieved unprecedented results on many (NLP) tasks, including machine translation (MT), a.o.~\cite{zhu2023multilingual,xu2024paradigm}. %And while prompting is an important factor on this quality~\cite{zhang2023prompting}, the used ICEs have also a substantial, if not even larger, impact.
Traditionally, creating ICEs for \gls{mt} involves either random selection~\cite{sia-duh-2023-context} or using a strategy such as maximizing an evaluation metric like BLEU, to choose examples that improve the metric~\cite{agrawal-etal-2023-context}. %from the development set 
 %(i.e., task-level prompts)
 %This strategy is based on the rationale that development sets closely resemble the domain of the test dataset. However, such approaches may not be practical in real-world MT scenarios, as labeled development sets are not always accessible. 
The former was initially used for its simplicity and ease of implementation. However, relying on randomness %in these methods 
can lead to inconsistent results and pose significant computational costs~\cite{lu-etal-2022-fantastically}. %For example, a randomized technique, such as the task-level prompt, typically demands nearly 100 random trials and approximately 78 hours of inferring, which can be costly in terms of both computational resources and time. %to achieve translation results comparable to those in the existing literature. 
Recent SOTA \gls{icl} approaches focus on retrieving training examples that are closely relevant to the context of source sentences of test sets using unsupervised retrievers, such as BM25~\cite{10.1561/1500000019}.
% This goes to related work
%accomplish this either through the embedding space of a pre-trained language model (PLM) –– a method previously shown to be effective in capturing context and domain~\cite{aharoni-goldberg-2020-unsupervised} –– or via unsupervised retrievers such as BM25 ~\cite{10.1561/1500000019} to provide additional context to the LLMs~\cite{shin-etal-2021-constrained,das-etal-2021-case,rubin-etal-2022-learning,agrawal-etal-2023-context}.
%SOTA ICL 
Studies have also shown that a range of factors, such as order~\cite{lu-etal-2022-fantastically}, template~\cite{10.1162/tacl_a_00324}, domain, and number of ICEs, significantly impact the output quality~\cite{agrawal-etal-2023-context, raunak-etal-2023-dissecting}. %However, these factors may vary depending on the specific LLM being used.
%Naturally, these factors have different impacts on different LLMs, making it essential to systematically analyze each phenomenon for individual LLMs and source texts. %Importantly, their impact can vary even within individual source texts within a test set. %For example, %using XGLM~\cite{lin-etal-2022-shot}, 
%a source sentence with 16 ICEs may yield a BLEU score of 50, while the same sentence with only one ICE could achieve a BLEU score of 100. %
%Consequently, determining the optimal quantity and quality of ICEs, their order, and templates necessitates systematically analyzing each phenomenon for individual LLMs and translated sources. 

Naturally, the most effective \glspl{ice} for a given source text are the ones that would maximize the resulting translation quality, which we typically judge based on translation references or human assessment. However, when such are not available, one needs to resort to \gls{qe} techniques. %Notably, these ICE-related factors differ among LLMs, necessitating a systematic analysis tailored to individual LLMs and source texts. Nonetheless, carrying out such an analysis poses significant challenges, requiring substantial time and computational resources due to the rapid growth and diversity of LLMs in the field.
%Nevertheless, such an undertaking would be not only time-consuming but also computationally expensive, given the rapid growth and diversity of LLMs.
%However, selecting the right options, i.e., quality, quantity, etc., boils down to knowing how they would impact the resulting translation quality, which would ultimately depend on translation references or human judgment.
In \gls{mt}, \gls{qe} has become a standard approach for evaluating an \gls{mt} system's output without relying on reference translations~\cite{blain-etal-2023-findings}.
Recently, \citeA{lee-2020-two}, \citeA{10.1007/978-981-99-7894-6_7}, and our previous chapter~\cite{sharami-etal-2023-tailoring} (see Chapter~\ref{chap:EAMT}) showed the effectiveness of domain-specific \gls{qe} when it comes to domain-specific \gls{mt} (in contrast to the ineffectiveness of generic \gls{qe}). \gls{qe} models are particularly valuable in scenarios where reference translations are unavailable, costly to have, or domain-specific, making them ideal for real-world deployment and feedback loops. Building on this and to address the aforementioned challenges, our work proposes to leverage domain-specific \gls{qe} to assist in the selection of \glspl{ice}, with the goal of determining the suboptimal number and combination of \glspl{ice} to maximize \gls{mt} quality, all without reference translations.
%We propose that domain-specific QE models, recently shown to be effective~\cite{lee-2020-two,sharami-etal-2023-tailoring}, could also be integrated into ICL methods to evaluate the quality of selected ICEs, including determining the optimal number of ICEs, without requiring translation references. 
As \gls{qe} would assess the impact of different \gls{ice} combinations and sequences, we hypothesize that this integration has the potential to not only improve translation performance but also reduce processing time, as \gls{qe} could result in smaller sets of \glspl{ice}, which would reduce the inference times~\cite{petrov2023language}. This is particularly crucial considering the limited number of \glspl{ice} that can be fed into LLMs due to their constrained input capacity~\cite{agrawal-etal-2023-context}. In this chapter, we present our work on selecting \glspl{ice} on a per-source basis. Specifically, we aim to answer the following \gls{rq}: \textit{How effective are domain-specific \gls{qe} models in determining \glspl{ice} for translation tasks in an LLM?}

%In addition, we follow two objectives: (i) Given the success of ordering ICEs based on their n-gram overlap match with the source, as per~\cite{agrawal-etal-2023-context}, we aim to investigate the efficacy of this approach within our proposed methodology. (ii) We aim to evaluate the impact of using our proposed methodology for ICL in comparison to fine-tuning a pre-trained multilingual MT model with regard to translation quality and computational costs. By considering all computational factors, this comparison allows us to determine whether ICL presents a more advantageous approach compared to fine-tuning a pre-trained MT model.  
    
Our proposed \gls{icl} methodology for \gls{mt} combines an unsupervised retriever to select \glspl{ice} with \gls{qe} to assess their impact on the translation quality, determining which \gls{ice} combination to include. Instead of feeding all selected examples, we only select examples whose \gls{qe} points to maximizing the LLM translation quality. %In our proposed approach, we pass the examples with their respective order determined by the retriever module to the LLM. After obtaining the translated output from the LLM, along with the source, we feed them to the QE model and keep a record of the scores estimated by the QE model. Finally, we select the combination of ICEs for each source, yielding the highest estimated BLEU score. 

Our findings on German-English translations demonstrate that our proposed approach outperforms the current SOTA \gls{icl} strategies, including R-BM25~\cite{agrawal-etal-2023-context}––a retrieval-based method that ranks \glspl{ice} using BM25 with a reranking step––as well as a fine-tuned mBART-50 model~\cite{tang2020multilingual}. Specifically, our method yields up to 8.3 BLEU and 1.35 COMET points higher than mBART-50, and achieves statistically significant improvements over baseline retrieval methods while maintaining a lower computational and environmental footprint. Furthermore, we show that the patience-based search guided by domain-specific \gls{qe} not only identifies more effective \gls{ice} combinations but also produces translations that are more length-aligned with human references, thereby reducing the need for post-editing. Unlike fine-tuning, which requires high computational cost and CO\textsubscript{2} emissions, our method is highly reusable and efficient for inference-time adaptation, making it both practical and sustainable.
%Further information about the compared systems is provided in section~\ref{AMTA:baselines}.
%We also assessed the impact of using our proposed methodology in comparison to fine-tuning a pre-trained multilingual MT model, namely mBART-50~\cite{tang2020multilingual}, with regard to translation quality and computational costs. Our findings show that while fine-tuning mBART-50 incurs significant computational costs and results in a notably lower BLEU score compared to ICL methods, it leads to improved contextual translation performance.
%By considering all computational factors, this comparison allows us to determine whether ICL presents a more advantageous approach compared to fine-tuning a pre-trained MT model.  
%Furthermore, our analysis reveals that although the translation quality can potentially be enhanced by ordering ICEs based on their n-gram overlap with the translated source, this improvement lacks statistical significance. Additionally, our experiments show that while fine-tuning a pre-trained model customized for MT incurs significant computational costs and results in a notably lower BLEU score compared to ICL methods, it leads to improved contextual translation performance.

We outline our methodology in Section~\ref{AMTA:sec:methodology} and present empirical evaluation in Section~\ref{AMTA:sec:experiment_setup} (Experiments Setup) and Section~\ref{AMTA:sec:results} (Results). Further analysis is presented in Section~\ref{AMTA:sec:analysis}. We review related work in Section~\ref{AMTA:sec:related_work} and conclude in Section~\ref{AMTA:sec:conclusion} with a summary and suggestions for future research.

\section{In-context learning using quality estimation for machine translation}
\label{AMTA:sec:methodology}
To utilize LLMs for effective \gls{mt}, as noted in Section~\ref{AMTA:sec:introduction}, what is needed is a set of examples to provide the context (and thus guide or steer the LLM toward a correct, context-specific translation) –– that is, a set of \glspl{ice} –– and what is further important is the number of \glspl{ice} and their combination.\footnote{The question of the order of examples is not specifically discussed in this chapter but is left for future work.} Ultimately, what is required is that the \glspl{ice} provide context that is neither too specific nor too broad and can effectively boost the translation. Our goal with this work is to develop a methodology that optimizes both these aspects in order to deliver high-quality \gls{mt}. Our methodology for identifying effective \glspl{ice} involves two key components: (1) an unsupervised retriever that locates examples closely related to the sentence to be translated and (2) a search algorithm that uses \gls{qe} to select a combination of examples that leads to the improvement of translation quality, i.e., aiming to maximize the BLEU score. %The way these two are connected is as follows: the retrieved examples (initial ICEs) are fed into the search algorithm. Subsequently, the search algorithm determines the optimal combination of ICEs for each source text based on QE analysis, aiming to maximize the BLEU score. %we present details in Section~\ref{AMTA:retrieval_module}

\subsection{Unsupervised retriever ranking}\label{AMTA:retrieval_module}
Building on the idea of selecting relevant examples introduced in Chapter~\ref{chap:CLIN}, we continue to focus on identifying training pairs that are most useful for a given translation task. Instead of relying on dense semantic representations, here we turn to a simpler yet effective method for ranking: the \emph{BM25} algorithm~\cite{10.1145/2682862.2682863}. %we use \emph{BM25} ranking algorithm~\cite{10.1145/2682862.2682863} because it has been shown to work well in similar settings, especially for retrieving relevant ICEs in unsupervised settings~\cite{agrawal-etal-2023-context}. BM25 sorts training pairs (source text and their translations) based on their relevance to a given query, i.e., the sentence to be translated. BM25 determines relevance by analyzing how often words from the query appear in each source text and adjusting for document length to prevent longer texts from being unfairly favored. This ensures that training pairs with source texts closely matching the sentence to be translated are prioritized, enhancing the selection of pertinent in-context examples. 
%source text from the test set.
BM25 has been widely used in information retrieval and has shown strong performance in similar settings, particularly for finding relevant \glspl{ice} in unsupervised scenarios~\cite{agrawal-etal-2023-context}.
It works by scoring how well each candidate sentence matches the input query—the sentence to be translated—based on overlapping words, while also adjusting for sentence length. This helps us surface examples that are more likely to be useful, without requiring additional model training or complex representations.

Subsequently, we select the top $K$ sentence pairs ranked by the algorithm, where $K$ is a hyperparameter that controls the number of pairs to be fed into the search algorithm. 
%A detailed explanation of this module's implementation is provided in Section~\ref{AMTA:retrieval_module}.
%we pass the examples with their respective order determined by the retriever module to the LLM. After obtaining the translated output from the LLM, along with the source, we feed them to the QE model and keep a record of scores estimated by the QE model. Finally, we select the combination of ICEs for each source, yielding the highest estimated BLEU score. 
\subsection{Search algorithm coupled with \gls{qe}}\label{AMTA:sec:search_alg}
%We propose an algorithmic approach that combines a search algorithm with QE to identify the most effective combination of ICEs. Our method focuses on selecting ICEs that contribute to enhancing translation quality from a predefined set of ICEs provided by the BM25 algorithm. %We incorporate early stopping patience into the search process to manage computational costs.
Our search algorithm comprises three main phases: \emph{Selection, Translation}, and \emph{Estimation}. During the Selection phase, the algorithm selects the highest-ranked training\footnote{We use the term ``training'' because these examples are drawn from the training set in our experiments. However, they are not used to train any models.} example from the initial \glspl{ice} provided by the unsupervised retriever ranking method (out of $K$ \glspl{ice}). This selected example is then concatenated with the previously selected \glspl{ice}. In the first iteration, no \glspl{ice} have been selected before. %A predetermined template is used for ICE creation (refer to~\ref{AMTA:prompt_creation}). 
In the Translation phase, the model receives both the selected \glspl{ice} and the test sentence and it generates a translation only for the test sentence.
%encoded using the LLM, and the resulting encoded text is then 
In the Estimation phase, the LLM output (translated text) and the original source text are inputted into the domain-specific \gls{qe} model to estimate the quality of the translation. Our proposed methodology relies on sentence-level \gls{qe}. 

Next, the selected \gls{ice}, together with its estimated quality and the LLM translation output, are appended to an intermediate list. To track the highest quality obtained thus far, the algorithm sorts the list in descending order based on the estimated quality. To avoid duplication, the selected \gls{ice} is removed before the next iteration. This iterative process continues until the best-estimated translation quality no longer improves within the specified patience threshold. Alternatively, the process terminates once all $K$ \glspl{ice} have been selected.

\begin{figure*}[h]
    \centering
    \includegraphics[keepaspectratio,width=\textwidth]{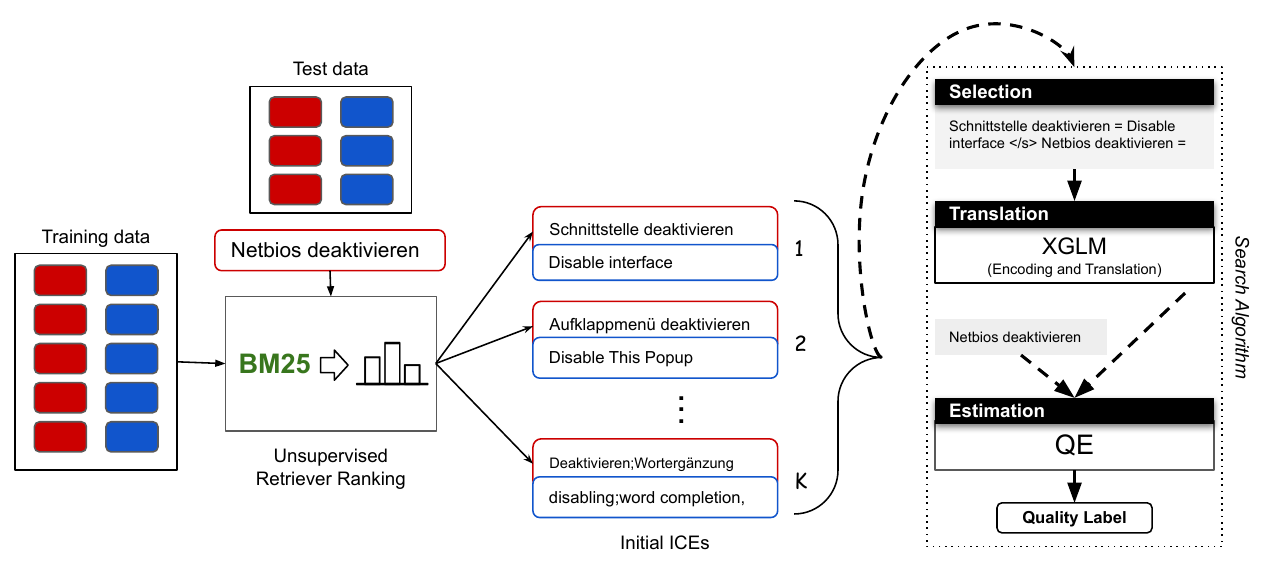}
\caption{\textbf{Overview illustration showing an iteration of our proposed methodology.} All selected \glspl{ice} are combined into a single prompt together with the test sentence, and the model generates a translation only for the test input. The \glspl{ice} are not translated in this process; instead, they serve as contextual examples that prime the model for better translation. %``XGLM'' is the LLM %used in our experiments
%. ``QE'' represents the quality estimation model. %``Quality Label'' denotes the quality gold label, which in our experiment is BLEU.
}
    \label{AMTA:fig:overview}
\end{figure*}

This methodology allows for the systematic selection of \glspl{ice} that improve translation quality compared to previous \gls{icl} methodologies while efficiently managing the computational resources required for the search process. This efficiency is achieved by integrating early stopping conditions with predetermined patience. Notably, we do not explore permutations of initial \glspl{ice}, as doing so would require a large number of attempts, leading to high computational costs during the search process. A pseudocode outlining the search methodology can be found in Algorithm~\ref{AMTA:search_pseudo}. The phases of translating a source text of a test set using our methodology are depicted in Figure~\ref{AMTA:fig:overview}.

\begin{figure*}[!ht]
\renewcommand{\baselinestretch}{1.2}
\begin{algorithmic}[1]
\scriptsize
\Function{Search}{...}
    \State $\text{temp} \gets [(\text{``'', 0.0, ``''})]$
    \State $\text{prompt} \gets \text{``''}$
    \State $\text{itr} \gets 0$
    \State $\text{best\_qe\_score} \gets 0.0$
    \State $\text{patience\_counter} \gets 0$
    \While{$\text{itr} < \text{iteration} \text{ and } \text{patience\_counter} < \text{early\_stop\_patience}$}
        \State $\text{available\_Prompts} \gets \Call{GenerateAvailablePrompts}{...}$ \Comment{Initial ICEs}
        \If{$\text{available\_prompts}$ is not empty}
            \State $\text{selected\_prompt\_index} \gets \text{itr} \bmod k$ \Comment{Phase 1: Selection}
            \State $\text{selected\_prompt} \gets \text{available\_prompts}[\text{selected\_prompt\_index}]$
            \State $\text{prompt} \gets \Call{ConstructFullPrompt}{...}(see~\ref{AMTA:LLM})$
            \State $\text{input\_ids[0]} \gets \Call{EncodePrompt}{...}$ \Comment{Phase 2: Translation}
            \If{$\text{length}(input\_ids) > \text{LLM\_max\_length}$}
                \State \Return $\text{temp}$
            \EndIf
            \State $\text{output} \gets \Call{GenerateOutput}{...}$ 
            \State $\text{final\_output} \gets \Call{DecodeOutput}{...}$
            \State $\text{qe\_input} \gets \Call{PrepareQEInp.}{\text{source}, \text{final\_output}}$ \Comment{Phase 3:~Estimation}
            \State $\text{qe\_score} \gets \Call{EstimateQuality}{\text{qe\_input}, \text{model\_QE}}$
            
            \State $\text{temp}.\text{append}((\text{prompt}, \text{current\_qe\_score}, \text{final\_output}))$
            \If{$\text{current\_bleu\_score} \geq 100$}
                \State \Return $\text{temp}$
            \EndIf
            \State $\text{temp} \gets \Call{SortTemp}{...}$
            \If{$\text{qe\_score} \leq \text{best\_qe\_score}$}
                \State $\text{patience\_counter} \gets \text{patience\_counter} + 1$
            \Else
                \State $\text{patience\_counter} \gets 0$
            \EndIf
            \State $\text{best\_qe\_score} \gets \text{temp}[0][1]$
        \EndIf
        \State $\text{itr} \gets \text{itr} + 1$
    \EndWhile
    \State \Return $\text{temp}$
\EndFunction
\end{algorithmic}
    \captionof{algorithm}{\textbf{Pseudocode outlining the proposed Search Algorithm.} Each phase of the methodology is annotated alongside the relevant code. Function arguments are omitted for simplicity. The first element of the returning list (\textit{temp}) includes the selected prompt, its associated \gls{qe} score, and the translated text.}
    \label{AMTA:search_pseudo}
\end{figure*}

% Overview

\section{Experiments setup}
\label{AMTA:sec:experiment_setup}
We conducted four main experiments to test the effectiveness of our methodology. Three of these experiments compare our methodology to existing \gls{icl} ones in different settings, or \emph{Modes}. %(detailed in Section~\ref{AMTA:sec:modes}). %The fourth one compares our methodology to fine-tuning, aiming to assess which method is preferred (with respect to obtaining better translations).
The fourth experiment compares our methodology %as an ICL method 
to a fine-tuned mBART-50, aiming to assess which method is preferred (with respect to obtaining better translations).

It is important to note that we do not fine-tune the LLM. %The process of building QE is aligned with what we discussed in Chapter\ref{chap:EAMT}.
Details regarding the \gls{qe} model used in our experiments are provided in Section~\ref{AMTA:qe}. The \gls{qe} model we use follows the same general training setup as described in Chapter~\ref{chap:EAMT}, but with a few adjustments to better align with the goals of this experiment.

%Experiment 1 evaluates the effectiveness of our proposed methodology compared to existing methods in the literature. Experiment 2 explores the impact of re-ordering ICEs based on unigram overlaps versus using the standard BM25 ranking. Experiment 3 examines the reduction in BLEU score relative to Experiment 1. 

%\subsection{Unsupervised Retriever}\label{AMTA:retrieval_module}
%We employed an unsupervised retriever method, BM25~\cite{10.1145/2682862.2682863}, to create the initial ICEs utilized by the proposed search algorithm. The initial number of ICEs was set at 16 to maintain consistency with earlier studies for a valid comparison.

%A detailed explanation of the implementation can be found in Section~\ref{AMTA:sec:BM25}.
%implemented through \textit{BM25Okapi} in the \textit{rank\_bm25} package~\cite{10.1145/2682862.2682863}\footnote{\url{https://github.com/dorianbrown/rank_bm25}}. 

%\paragraph{Ordering ICEs Based on N-grams}  Given the use of ordering ICEs based on their n-gram overlap match with the source, as per~\cite{agrawal-etal-2023-context}, one of our objectives is to investigate how ordering the initial ICEs according to their n-grams (specifically unigrams) similarity with the source text affects the quality of translation. Therefore, in addition to ICEs ranked by BM25, we reorder them based on n-grams similarity. This involves tokenizing the sentences using NLTK word tokenizer and calculating the level of overlap between each. Sentences with greater levels of overlap matches will be given precedence in the list and fed into the LLMs earlier.

\subsection{Search algorithm}
We conducted experiments using the search algorithm (see Section~\ref{AMTA:sec:search_alg}) employed in three different ways (modes):
% \subsubsection{Modes}\label{AMTA:sec:modes}
% \begin{enumerate}[label=Mode \arabic*:, leftmargin=*]
\paragraph{Mode 1:} This mode uses \gls{qe} with \glspl{ice} ordered by BM25 to assess the effectiveness of combining BM25 and \gls{qe} in the proposed \gls{icl} methodology.
    
    %\item This mode entails the use of QE along with ICEs ordered by n-gram overlap, with the purpose of investigating the impact of n-gram ordering on the proposed methodology. 
\paragraph{Mode 2:} This mode investigates the impact of ordering \glspl{ice} according to uni-gram overlap, alongside \gls{qe}, on the proposed methodology. Given the success of ordering \glspl{ice} based on their n-gram overlap match with the source, as demonstrated in~\cite{agrawal-etal-2023-context}, we assess how this ordering, based on \glspl{ice}' uni-gram overlap with the source text, influences the translation quality. This involves reordering \glspl{ice} according to their n-gram overlap, which is calculated using the NLTK word tokenizer. Higher overlap matches prioritize \glspl{ice} in the list and feed them into LLMs earlier.

\paragraph{Mode 3:} %This mode does not involve QE and orders initial ICEs based on BM25 ranking. 
Instead of relying on \gls{qe}, in this mode we compute the BLEU score directly on the test set. This mode is not realistic because the method has access to the gold references—which is never available at inference time in MT evaluation—and can therefore select \glspl{ice} that explicitly maximize translation quality; in other words, the labels are exposed. Nonetheless, this mode serves as an upper-bound scenario and is used only as a best-case comparison point.

% \end{enumerate}

\subsection{Early stopping conditions} 
The search algorithm generates up to 16 candidates. In each mode, we conducted experiments using three early stopping patience values (3, 8, and 16), determining the maximum number of \glspl{ice} ($K$) generated. We included Patience 16, which implies no early stopping, to evaluate the model's performance with the maximum \glspl{ice}. Additionally, the search process halts if the estimated \gls{qe} label reaches or exceeds 100, preventing further evaluations.

%The search algorithm generates a maximum of 16 candidates. In each of the aforementioned modes, we conducted experiments using three distinct early stopping patience values, which set the maximum number of ICEs ($K$) that can be generated by the search algorithm. We experimented with values 3, 8, and 16, where Patience 16 implies no early stopping. Patience 16 was specifically examined in order to evaluate the model's performance with the maximum ICEs generated.

%In addition, we included a termination condition that activates when the (estimated) label reaches or exceeds 100. Upon activation, this condition halts the search process, preventing further attempts to evaluate other ICEs. %This feature has been implemented to enhance time effectiveness.

\subsection{Quality estimation}
\label{AMTA:qe}
%Our proposed methodology requires a sentence-level QE model to effectively guide the selection process of the search algorithm. 
Following~\cite{ranasinghe-etal-2020-transquest,lee-2020-two} and our earlier approach presented in Chapter~\ref{chap:EAMT}, we develop a domain-specific \gls{qe} model. %This development involves two key phases: 
First, we trained a \gls{qe} model using out-of-domain (OOD) data (detailed in the paragraph after next) to ensure generalizability; and second, we fine-tuned the model using the training set described in Section~\ref{AMTA:sec:data} to provide domain-specific \gls{qe} model and address domain mismatch, which is critical~\cite{koehn-knowles-2017-six}. %issue in both MT and QE

In our experiments, we used BLEU as the quality label because our study focused on translation performance rather than post-editing effort, which is typically evaluated using (H)TER~\cite{specia-farzindar-2010-estimating}. % in the context of QE experiments. 
We employed the ``MonoTransQuest'' architecture from the TransQuest framework~\cite{ranasinghe-etal-2020-transquest}, known for its success in prior \gls{qe} studies. However, instead of employing softmax computation, we directly utilized logits to estimate the quality labels. This strategy saves computation time, as softmax computation can be resource-intensive~\cite{ruder2016wordembeddingspart2}. %but also eliminates the need to determine probabilities for each prediction in our experiments.

% \subsubsection{QE data}
% \label{AMTA:sec:syntheticQEdata}
We utilized the German-English ``EuroPat v3'' dataset, accessed through Opus~\cite{tiedemann-2012-parallel}%\footnote{\url{https://opus.nlpl.eu/}}
, to develop our generic \gls{qe} model. We chose this dataset because it provides ample data samples (around 20M), ensuring broad coverage of vocabulary––a critical aspect in developing generic models.

However, as \gls{mt} datasets like EuroPat typically consist of pairs of source and translated text, it was necessary to synthetically create post-editing text (since the \gls{qe} data creation process requires a triplet input: source text, machine-translated text, and post-edited text). To accomplish this, we used a pre-trained multilingual \gls{mt} model, namely mBART-50 that supported the language pair used in our experiment. We translated 1M randomly chosen source texts from EuroPat. Afterward, the resulting translations were considered as machine-translated text, with the corresponding reference translations acting as post-edited text.

Using SacreBLEU, we calculated the BLEU score, comparing the translated text with its corresponding post-edited text. This approach, which has been demonstrated to be effective in \gls{qe}~\cite{negri-etal-2018-escape,lee-2020-two,sharami-etal-2023-tailoring}, enabled us to use the source and (machine-) translated text as input and the BLEU score as the target value for the \gls{qe} model.
For building domain-specific \gls{qe}, we utilized the training set detailed in Section~\ref{AMTA:sec:data} and applied the aforementioned approach to synthetically generate BLEU scores for the entire dataset.

\subsection{Dataset and evaluation metrics}\label{AMTA:sec:data}
We used a dataset comprising German-to-English translation pairs within the IT domain, sourced from~\cite{aharoni-goldberg-2020-unsupervised}. This dataset was chosen due to the challenges that \gls{mt} systems and LLMs face when translating out-of-domain contexts, particularly in specialized fields, as noted in previous studies~\cite{koehn-knowles-2017-six,agrawal-etal-2023-context}. The specialized and constrained nature of the IT domain provided an ideal setting for evaluating our methodology's performance under these conditions.

The dataset utilized in this study consisted of approximately 222k training sentences, 2k development sentences, and 2k test sentences. 
To assess the translation effectiveness of the models, we employed metrics such as BLEU from SacreBLEU~\cite{post-2018-call} and COMET~\cite{rei-etal-2020-comet}. 

It is noteworthy that we used the development set during \gls{qe} model fine-tuning to avoid any exposure to the test set. This ensured that the test data remained fully unseen and could serve as a reliable benchmark for evaluating our method's generalization performance.

\subsection{Number of \glspl{ice}}\label{AMTA:sec:p_q}
We use between 1 and 16 \glspl{ice}. These may originate either from a random approach or from an advanced (guided) selection, i.e., the \glspl{ice} selection process is not random but follows specific rules instead of sampling examples arbitrarily from a pool. In line with prior work~\cite{agrawal-etal-2023-context}, we adopt the symbols \( p \) and \( q \) to denote the number of task-level and example-specific prompts, respectively. However, our method does not distinguish between these two levels: all selected \glspl{ice} are combined into a single prompt together with the test sentence. Therefore, in our experiments \( p \) and \( q \) refer to the same quantity—the number of \glspl{ice} included in the final prompt. We retain this notation only for comparability with previous studies. While we cap the maximum at \( q = 16 \), any value between 1 and 16 is feasible in our setup.

\subsection{Multilingual large language model}\label{AMTA:LLM}
For our experiments and hypothesis validation, we used XGLM~\cite{lin-etal-2022-shot}. %a language model developed by Facebook. 
This choice stems from the outstanding performance of the model in the \gls{mt} field. This also ensures a fair comparison of our proposed methodology with previous research, such as~\cite{agrawal-etal-2023-context}, which introduced SOTA approaches in \gls{icl} for \gls{mt}.

We used the 7.5 billion-parameter XGLM implementation and tokenizer by Hugging Face\footnote{\url{https://huggingface.co/docs/transformers/model_doc/xglm}}, consistent with previous research. % \subsubsection{Prompt Template}\label{AMTA:prompt_creation}
%In our study, using the LLM detailed in Section~\ref{AMTA:LLM}, 
We employed a template from~\cite{lin-etal-2022-shot} to maximize translation performance. $</s>$ serves as the \gls{ice} separator in this template.  ``BLANK'' denotes an empty string within the template.
\begin{equation}
{\small \begin{split}
\text{\{source text}_1\} & = \text{\{target text}_1\} </s> \\
\text{\{source text}_2\} & = \text{\{{target text}}_2\} </s> \\
\ldots  & = \ldots </s> \\
\text{\{source text}_n\} & = \text{BLANK} \notag
\end{split}}
\end{equation}

\subsection{Compared systems}\label{AMTA:baselines}
%To evaluate the efficacy of our proposed ICL methodology in contrast to methodologies used in previous studies, we conducted a comparative analysis by comparing our findings to those obtained through different methods proposed in the existing literature. These methods consist of random sampling, task-level sampling, BM25, R-BM25, and mBART-50, which will be elaborated in the following paragraphs.
%To evaluate the efficacy of our proposed ICL methodology, 
We conducted a comparative analysis with methods from previous studies; \textit{random} and \textit{task-level sampling}, \textit{BM25}, \textit{R-BM25}, and \textit{fine-tuned mBART-50}. %Our baseline is defined as the method with the highest performance, detailed in Section~\ref{AMTA:sec:SelectedBL}.
\paragraph{Random:}
We conducted three random trials, each based on a parameter $p$, which defines the number of \glspl{ice} to include in the prompt. For each trial, we randomly sampled $p$ integers between 1 and the size of the training set. These integers were used to select the corresponding translation pairs from the training data. For example, if $p = 3$ and the randomly generated numbers are 1, 10, and 100, we select those three examples from the training set.
%We conducted three random trials, generating random numbers based on parameter $p$. These numbers, ranging from 1 to the size of the training set, selected corresponding translation pairs. In cases where multiple numbers are generated, we identify and choose the translation pairs associated with each of these numbers. 
%For example, if the randomly generated numbers are 1, 10, and 100, we select the respective examples from the training set. 
To create the prompt\footnote{In the literature, the term ``prompt'' is often used interchangeably with ``\gls{ice}''}, in addition to the \glspl{ice}, we need the source side intended for translation. We use the source from the development set, unlike more advanced methods in \gls{icl}, where the source text from the test set is typically employed. %That is, we concatenate the selected examples with each source text from the development set using the prompt template. %The reason for selecting the development set over the test set in this approach is that it has a higher potential to match the test set content.
The reason for selecting the development set over the test set in this approach is that development sets are generally from the same distribution, domain, and context as the test set. This similarity increases the likelihood that the examples in the development set will better match the content and context of the test set, thereby enhancing the relevance and effectiveness of the prompts.

The generated prompt is inputted into the LLM for translation. Then, the BLEU score of the development set is computed. The random number that produces the highest score among the trials is selected, and the \glspl{ice} linked to this number are concatenated with the test set's source text.

\paragraph{Task-level:}
Based on the work of \citeA{agrawal-etal-2023-context}, the task-level approach~\footnote{While the term ``task-level'' is used, it is important to note that this approach uses random sampling to generate a diverse set of \glspl{ice}, aiming to enhance the performance of LLMs in translation tasks.} is similar to the random approach but differs in the number of trials used. We employ 100 trials for the task-level approach, a significantly higher number than the random approach. The reason for using more trials is to generate a greater variety of \glspl{ice}, aiming to enhance the performance of LLMs in the translation task. However, this results in longer execution times compared to the random approach.

\paragraph{BM25:}\label{AMTA:sec:BM25}
%Previous studies suggest that unsupervised retrievers, such as BM25, offer substantial improvements in ICL performance compared to random and task-level approaches~\cite{liu-etal-2022-makes,luo2023dricl,wang2024learning}. 
Using the Moses Tokenizer~\cite{koehn-etal-2007-moses}, we first tokenize the training set's source samples. % based on their respective language. %\footnote{\url{https://github.com/luismsgomes/mosestokenizer}}
% Then, a BM25 model is created for the tokenized corpus by employing the \textit{BM25Okapi} implementation within the \textit{rank\_bm25} package.%~\cite{10.1145/2682862.2682863}.
The BM25 model is then created over the tokenized corpus, where it computes term frequency and inverse document frequency statistics necessary for scoring the lexical relevance of test queries.\footnote{\url{https://github.com/dorianbrown/rank_bm25}}
%~\footnote{We used \textit{BM25Okapi} implementation within the \textit{rank\_bm25} package~\cite{10.1145/2682862.2682863}.}

Next, the test set is %iterated through, and each source text is 
tokenized using the tokenized source. 
The algorithm then searches for similar training samples based on BM25 criteria, selecting the top $q$ matches for the model. %We used the template mentioned earlier (refer to \ref{AMTA:prompt_creation}) to create the prompts. 
This methodology utilizes the test set as opposed to random and task-level approaches using the development set. The reason is that BM25 retrieves \glspl{ice} by ranking examples based on lexical similarity to each specific source sentence, eliminating the need for repeated trials or development set proxies.

\paragraph{Re-rank BM25 (R-BM25):}
%BM25 aims to find translation examples with the highest n-gram overlap with the source sentence of the test set~\cite{luo2023dricl}. However, since there is no link between retrieved examples (i.e., they score independently), the top matches may not include all the n-grams present in the source text. This poses an issue in ICL since the input size of the LLMs is typically limited. That is, they are unable to receive a large number of ICEs to make up for the lack of coverage of certain n-grams. To address this issue, \cite{agrawal-etal-2023-context} proposed a re-rank version of BM25 called R-BM25.
BM25 aims to find translation examples with the highest n-gram overlap with the source sentence~\cite{luo2023dricl}. However, since retrieved examples score independently, top matches may lack coverage of all source n-grams. This poses an issue in \gls{icl} due to LLM input size limitations. To address this, \citeA{agrawal-etal-2023-context} proposed R-BM25, a recall-based variant of BM25 that promotes broader n-gram coverage. It works by extracting all word n-grams from the test sentence and from each BM25-retrieved candidate. At each step, the candidate that contributes the most uncovered n-grams is selected and added to the prompt. Once an example is selected, the n-grams it covers are given less weight in future iterations––essentially telling the algorithm: ``this part is already covered, focus on what is missing." This iterative process continues until a desired level of n-gram coverage is achieved.
%, employing a recall-based n-gram overlap%~\cite{agrawal-etal-2023-context} 
%to extract word n-grams and their numbers from the test source and BM25 retrieved examples. %This is accomplished by computing a recall-based n-gram overlap score, which is explained in detail in the original paper (refer to~\cite{agrawal-etal-2023-context}). 
%The example that receives the highest score is included in the selected prompts set. For the next iteration of selection, the test source n-grams are down-weighted by a certain factor. The process is performed repeatedly until a specific threshold is reached.

\paragraph{Fine-tuning mBART-50:}
Different \gls{icl} methodologies, including our own, are assessed in comparison to the process of fine-tuning a pre-trained multilingual \gls{mt} model, specifically mBART-50. The selection of mBART-50 is based on its alignment with the language specifications of the experiment and its proven track record of achieving success in \gls{mt} tasks through the utilization of \glspl{plm}~\cite{Yuan2022AnIM,pham-etal-2022-effective}. The fine-tuning of mBART-50 is carried out using the training data outlined in Section~\ref{AMTA:sec:data}.

% \vspace*{-7.35mm}
\subsection{Computational costs}
We monitored and reported the computational costs of the models utilized in our experiments using the \emph{carbontracker} package.\footnote{\url{https://github.com/lfwa/carbontracker}} This involved calculating the carbon footprint (CO\textsubscript{2}eq) emissions, time to prediction (TTP), and electricity consumption (kWh) associated with our experiments. Our experiments were conducted using NVIDIA A40 GPUs.

The script for running our experiments is publicly available at \url{https://github.com/JoyeBright/ICLviaQE}.

\section{Experiments results}\label{AMTA:sec:results}
This section presents the results of our experiments. %To ensure a fair comparison, 
We conducted a statistical analysis test (t-test) to determine if our models significantly outperformed the compared systems. 

\subsection{Evaluation metrics}
% \subsection{Compared Systems}\label{AMTA:sec:SelectedBL}
Among prior work, the results in Table~\ref{AMTA:table:main-scores} show that R-BM25 with 16 \glspl{ice} outperforms all other methods. Because of this strong performance, we used it as our baseline for further comparisons.

It is notable that there is a positive correlation between the number of examples and evaluation scores (consistent through all methods---Random, Task-level, BM25, and R-BM25), although at the expense of prediction time (i.e., TTP). Employing 16 examples significantly improved performance compared to using only one example in the random approach. 
% The results from the methods in the literature, shown in Table~\ref{AMTA:table:main-scores}, revealed that R-BM25 with 16 ICEs leads to superior performance compared to the other methods. As a result, we considered the R-BM25 model as our baseline for further comparative analysis. Moreover, while the random approach with only one ICE yielded a BLEU score of 10.38, indicating the lowest performance, this score significantly improved to 31.65 when employing 16 examples. This correlation between the number of examples and evaluation scores was observed across all methods (Random, Task-level, BM25, and R-BM25), although at the expense of prediction time (i.e., TTP).

\begin{table}[t!]
\centering
\renewcommand{\arraystretch}{1.1}
\setlength\tabcolsep{4pt} 
\resizebox{0.9\linewidth}{!}{%
\begin{tabular}{l|rrrrrr}
\hline\hline
\textbf{Method}  & \textbf{\(\boldsymbol{p + q}\)} & \textbf{BLEU} & \textbf{COMET} & \textbf{TTP} & \textbf{CO2} & \textbf{GPU} \\
\hline\hline
Random     & 1 + 0  & 10.38 & 0.6895 & 01:51 & 00.13 & 00.39 \\ 
Random     & 16 + 0 & 31.65 & 0.7844 & 02:20 & 00.19 & 00.58 \\ 
Task-level & 1 + 0  & 29.17 & 0.7586 & 62:50 & 09.83 & 29.10 \\ 
Task-level & 16 + 0 & 32.88 & 0.8083 & 78:30 & 12.80 & 35.91 \\ 
BM25       & 0 + 1  & 39.24 & 0.7833 & 00:56 & 00.06 & 00.19 \\ 
BM25       & 0 + 16 & 44.50 & 0.8120 & 00:58 & 00.07 & 00.19 \\ 
R-BM25     & 0 + 1  & 40.88 & 0.7990 & 01:01 & 00.06 & 00.21 \\ 
R-BM25     & 0 + 16 & \textbf{45.20} & \textbf{0.8218} & 01:04 & 00.07 & 00.21 \\\hdashline[1pt/1pt] 
M 1, P = 3 & 0 + 16 & 45.72 & 0.8395 & 01:49 & 00.22 & 00.67 \\ 
M 1, P = 8 & 0 + 16 & \textbf{46.43} & \textbf{0.8501} & 03:48 & 00.50 & 01.51 \\ 
M 1, P = 16 & 0 + 16 & \textbf{46.78} & \textbf{0.8554} & 05:11 & 00.68 & 02.05 \\ \hdashline[1pt/1pt]
M 2, P = 3 & 0 + 16 & 46.05 & 0.8400 & 01:30 & 00.21 & 00.64 \\ 
M 2, P = 8 & 0 + 16 & \textbf{46.59} & \textbf{0.8518} & 03:52 & 00.51 & 01.52 \\ 
M 2, P = 16 & 0 + 16 & \textbf{46.52} & \textbf{0.8564} & 05:00 & 00.66 & 02.01 \\ \hdashline[1pt/1pt]
M 3, P = 3 & 0 + 16 & 49.89 & 0.8532 & 01:36 & 00.22 & 00.66 \\ 
M 3, P = 8 & 0 + 16 & \textbf{52.63} & \textbf{0.8725} & 03:14 & 00.45 & 01.40 \\ 
M 3, P = 16 & 0 + 16 & \textbf{53.50} & \textbf{0.8791} & 04:08 & 00.55 & 01.65 \\ \hdashline[1pt/1pt]
mBART-50   & N/A    & 42.76 & 0.8659 & 11:20 & 01.88 & 04.82 \\ 
\hline\hline
\end{tabular}
}
\caption{\textbf{Method Performance in BLEU and COMET Scores.} \( M \) 1 to 3 denotes Mode 1 to 3; \( P \) is the patience value. For consistency with prior work, \(p\) and \(q\) both refer to the number of \glspl{ice} included in the final prompt (see Section~\ref{AMTA:sec:p_q}). ``N/A'' (not applicable) indicates that fine-tuning does not use \glspl{ice}. Bold font represents the highest translation performance. Two numbers are in bold if they are statistically similar (t-test, \( p\_value = 0.05 \)). TTP is in (hh:mm), CO2 in (kg), and GPU in (kWh).}
\label{AMTA:table:main-scores}
\end{table}

% \subsection{Our Methodology}
%Table~\ref{AMTA:table:main-scores} also shows the evaluation of our proposed methodology across the various modes. 
Analyzing the performance of our methods in Mode 1 (referred to as ``M 1'', with P = 3, 8, or 16 in Table~\ref{AMTA:table:main-scores}), we observe that our proposed methodology with different patience thresholds consistently outperforms all previous methods, including the baseline. This trend holds for both the COMET and BLEU metrics across all the methods. Specifically, our method exhibits a minimum improvement of 0.52 points in the BLEU score (from 45.20 to 45.72) with patience threshold of 3 and a maximum improvement of 1.58 points in the BLEU score (from 45.20 to 46.78) with a patience threshold of 16 compared to R-BM25 with 16 examples.

Consequently, our methods in Mode 1 are ranked based on their performance, with patience 3 being the least effective model, followed by patience 8, and finally patience 16, representing the most effective method. This ranking indicates that increasing the patience threshold can significantly enhance the translation performance. However, the improvement with patience 16 is not statistically significant compared to patience 8, suggesting that more \glspl{ice} do not necessarily enhance translation performance. Similarly, while more substantial contextual improvement (as indicated by the COMET) is observed at the maximum patience threshold (16), it is not statistically significant compared to patience 8.

The Mode 2 results demonstrate that using any of the three patience thresholds leads to better results compared to the methods in the literature. However, this improvement is not statistically significant when compared to the respective experiments in Mode 1. This suggests that ordering the examples according to n-gram (unigram) similarity does not enhance the translation performance in our methodology. 

While ordering \glspl{ice} by unigram overlap (Mode~2) does not improve results, the initial BM25 retrieval naturally returns examples with relatively high lexical similarity. However, our QE-guided search does not optimize for n-gram overlap: it ranks combinations only by estimated translation quality. Thus, the gains of our method cannot be attributed solely to high n-gram similarity.

When it comes to Mode 3, we should stress that this is an unrealistic scenario, but used as the highest bound. The results indicate that with a patience of 3, the BLEU score is 4.17 points lower (49.89-45.72). With a patience of 8, this gap increases to 6.2 points (52.63-46.43), and with a patience of 16, it widens further to 6.72 points (53.50-46.78). These differences arise from the \gls{qe} model estimations in our experiment compared to the scenario where reference labels are available to the search algorithm.

\subsection{Time to prediction (TTP)}
Among the methods examined, task-level execution required the most time, with approximately 62 hours for one example and 78 hours for 16 examples, primarily due to the large number of trials involved. Our method (Mode 1) with a patience value of 16 is relatively time-intensive, taking approximately 5 hours, while a patience value of 3 is comparable to the baseline method, differing by only around 50 minutes. Mode 2 is nearly equivalent to Mode 1 in terms of TTP, whereas Mode 3, where the reference labels are accessed, requires less time than Modes 1 and 2. %This is primarily because the search algorithm accesses the reference labels. 
In addition, the search algorithm incorporates a termination condition, and given that \gls{qe} estimation rarely triggers this condition, numerous \glspl{ice} are left unattempted, resulting in significant time savings. 

It is also important to note the time required to train the \gls{qe} models used in the prediction process. As shown in Table~\ref{AMTA:app:QEstats}, the training time for the generic \gls{qe} model is $+/-$ 5 hours and 55 minutes, while the specific \gls{qe} model takes about $+/-$~6 hours and 54 minutes. Although these training times are significant, it is crucial to recognize that \gls{qe} models, similar to \gls{mt} models, can be reused for the same language pair and domain, thereby amortizing the initial training cost over multiple predictions.

% QE stats
\begin{table}[t!]
\centering
\renewcommand{\arraystretch}{1.2}
\begin{adjustbox}{width=\textwidth,center}
\begin{tabular}{l|rr}
\hline\hline
Metric & Generic Model & Specific Model \\ 
\hline\hline
Training Time (hh:mm) & 05:55 & 06:54 \\ 
CO2 Emissions (kg) & 1.41 & 1.46 \\ 
Electricity Consumption (kWh) & 3.63 & 3.76 \\ 
\hline\hline
\end{tabular}
\end{adjustbox}
\caption{\textbf{Training Time, CO2 Emissions, and Electricity Consumption for \gls{qe} Models}}
\label{AMTA:app:QEstats}
\end{table}

% \subsection{Fine-tuning vs. ICL}
The last row of Table~\ref{AMTA:table:main-scores} shows the scores of the translations obtained with the mBART-50 model fine-tuned on the same training set as in \gls{icl}. %results concerning the fine-tuning of a pre-trained multilingual model, specifically mBART-50, using the same training dataset as in ICL (referred to as ``FT: mBART-50''). 
Despite mBART-50 being tailored for \gls{mt} across 50 languages, it did not outperform the R-BM25 method with 16 examples (best from the existing methods); it was better only than Random, Task-level, BM25, and R-BM25, each with only 1 example. However, when considering translation performance from a contextual perspective, the COMET results indicate that fine-tuning mBART-50 leads to superior performance compared with lexical overlap. Nevertheless, fine-tuning took significantly longer than identifying \glspl{ice} and obtaining inferences from the XGLM.

Compared to our methodology, especially when considering the least performing method (M 1, P = 3), fine-tuning is significantly worse---6.47\% (42.76 to 45.72). This highlights the substantial benefits of \gls{icl} compared to fine-tuning. Nonetheless, it is noteworthy that various factors might contribute to this observation: e.g., the model's size might be a critical factor, especially during deployment, where larger models like XGLM could pose challenges.\footnote{However, this does not undermine the generalizability of our conclusions; rather, it points to the need for balancing performance with feasibility when selecting models for real-world applications.}

To contextualize this comparison, it is important to note that fine-tuning and \gls{icl} are typically performed using different model classes. Encoder--decoder architectures such as mBART-50 are the standard choice for supervised MT fine-tuning, whereas decoder-only models such as XGLM are commonly used for \gls{icl}. Our evaluation therefore reflects the typical use of each paradigm rather than a controlled architecture ablation. While architectural and size differences may influence absolute scores, this does not undermine the validity of our comparison, whose purpose is to assess the practical trade-offs between fine-tuning and \gls{icl} in MT workflows.

%Nonetheless, it is important to note that deployment constraints, particularly the large size of models like XGLM, may pose practical challenges. While our results demonstrate the effectiveness of ICL with XGLM, these deployment challenges could hinder replicating such performance in resource-constrained settings. However, this does not undermine the generalizability of our conclusions; rather, it points to the need for balancing performance with feasibility when selecting models for real-world applications.

\section{Analysis}\label{AMTA:sec:analysis}
\begin{figure}[t!]
    \centering
    \includegraphics[width=0.8\linewidth]{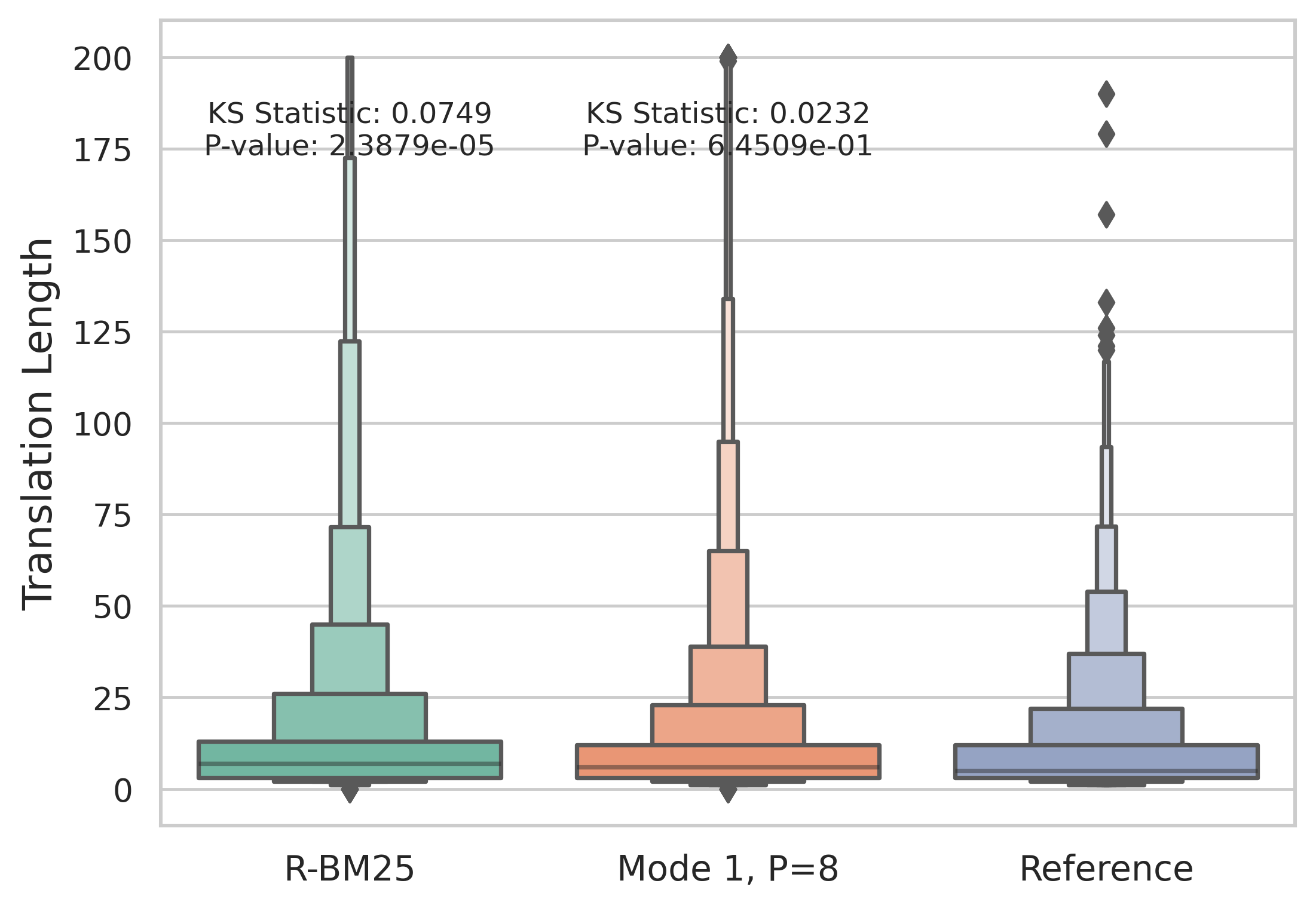} 
    \caption{\textbf{Tokenized Translation Lengths} comparison between R-BM25, our Mode 1, P=8, and the reference. ``KS'' denotes the Kolmogorov-Smirnov test, with the p-value indicating significance.}
    \label{AMTA:fig:length_output}
\end{figure}

\paragraph{Output analysis} Pre-trained \glspl{llm} often exhibit over-generation, i.e., the generation of a larger number of tokens than expected by a human (in comparison to a reference), necessitating extensive post-processing (e.g., post-editing)~% to refine their output for conciseness and relevance~
\cite{bawden-yvon-2023-investigating}. Figure~\ref{AMTA:fig:length_output} shows the tokenized output lengths (translations) for our model (Mode 1, patience 8),\footnote{Our other models in Mode 1 exhibited similar distributions.} alongside the R-BM25 with 16 examples. The analysis shows that the length distributions for both models align with the reference distribution, suggesting that the models do not over-generate. 

To quantitatively compare these distributions to the reference, we employed the Kolmogorov-Smirnov~(KS) test~\cite{kolmogorov1933sulla}.\footnote{The KS test is non-parametric, making it suitable for comparing translation length distributions without assuming normality—important since model outputs may not follow standard distributions.} The results indicate that for R-BM25 versus the reference, the KS statistic is relatively high (\(0.0749\)), reflecting a significant difference between the translation lengths of R-BM25 and the reference distribution. The extremely low p-value (\(2.39 \times 10^{-5}\)) further confirms this significant discrepancy. Conversely, for Mode 1 with P=8 versus the reference, the KS statistic is considerably lower (\(0.0232\)), indicating a much smaller difference in translation lengths. The higher p-value (\(0.6451\)) suggests no significant difference, implying that the distribution of Mode 1, P=8 is similar to the reference distribution. 

These findings suggest that our proposed methodology could yield translations closer in length to the reference, potentially reducing the need for labor-intensive post-processing efforts and enhancing computational efficiency.
%but also streamlines the overall translation process.
\vspace*{-2.00mm}

\paragraph{\gls{ice} number analysis}
The number of selected \glspl{ice} holds a significant importance within the \gls{icl} algorithm, as it directly impacts the token processing time and the capacity of LLMs to handle additional \gls{ice} instances. % (due to their limited capacity~\cite{agrawal-etal-2023-context}). 
We analyzed the number of \glspl{ice} that our algorithm selected %(where a pair of source and target counts as one ICE) 
across all three modes. The results (Table~\ref{AMTA:table:ICE_Count}) show that the minimum number of \glspl{ice} selected is 1, while the maximum is 12 for Mode 1, 16 for Mode 2, and 16 for Mode 3.
The average (mean) number of \glspl{ice} is found to be lowest in Mode 3 and highest in Mode 1. In addition, Mode 2 results in a reduction in the number of \glspl{ice} within our proposed algorithm. The notably lower average number of \gls{ice} instances in Mode 3 can be attributed to its access to the test set, allowing for the selection of optimal \gls{ice} combinations based on test set performance and activating an early stopping condition if the score exceeds 100. Contrarily, while Mode 1 exhibits similarities to Mode 3, its relatively higher average can be linked to inaccuracies in \gls{qe} estimation. Moreover, our analysis shows that \gls{qe} estimations rarely reach a score of 100, thus rendering the early stopping condition inactive. %for that aspect.

\begin{table}[t!]
\centering
\renewcommand{\arraystretch}{1.2}
\begin{adjustbox}{width=0.8\textwidth,center}
\begin{tabular}{l|ccc}
\hline\hline
Mode & Min & Mean & Max \\
\hline\hline
\#1 & [1, 1, 1] & [2.25, 3.76, 4.84] & [12, 16, 16] \\
\#2 & [1, 1, 1] & [2.20, 3.70, 4.74] & [12, 16, 16] \\
\#3 & [1, 1, 1] & [2.15, 3.47, 4.47] & [12, 16, 16] \\
\hline\hline
\end{tabular}
\end{adjustbox}
\caption{\textbf{Number of \glspl{ice} selected for each mode at different patience thresholds.} Labels [x, y, z] correspond to patience values 3, 8, and 16.}
\label{AMTA:table:ICE_Count}
\end{table}

\paragraph{CO\textsubscript{2} emissions}
%Besides evaluating the translation performance of the models, it is crucial to consider their computational costs, notably the associated CO\textsubscript{2} emissions. 
%Considering CO\textsubscript{2} emissions alongside translation performance is crucial. 
Our analysis reveals that using XGLM for translation yields lower CO\textsubscript{2} emissions than fine-tuning mBART-50, making it a more environmentally sustainable choice. In Mode 1 of our proposed methodology, with patience 16, XGLM emitted 0.68 KG of CO\textsubscript{2}, while fine-tuning mBART-50 emitted 1.88 KG. Interestingly, the task-level method with 16 \glspl{ice} emitted the highest amount of CO\textsubscript{2}, totaling 12.80 KG. Our approach produces CO\textsubscript{2} emissions higher than R-BM25.

\paragraph{Illustrative example.}
Table~\ref{AMTA:example of mode1} shows an example from our evaluation, including the selected \glspl{ice}, the model's translation, the reference label, and evaluation scores (BLEU and \gls{qe}).

%ICE example
\begin{table}[h!]
\centering
\renewcommand{\arraystretch}{1.4}
\scalebox{0.67}{
\begin{tabular}{ll}\hline
\addlinespace[2pt]
\multicolumn{2}{l}{\textbf{\underline{ICEs:}}}  \\
\multicolumn{2}{p{17cm}}{Die Sockets, die im except Array aufgelistet sind, werden auf Ausnahmen überwacht. = The sockets listed in the except array will be watched for exceptions. $</s>$ Geben Sie den Namen der Variablen ein, deren Wert überwacht werden soll. = Enter the name of the variable whose value is to be monitored. $</s>$ Nur erlaubt bei Sockets für lokale Displays und den globalen Socket. = Permitted only on sockets of local displays and the global socket. $</s>$ Legt fest, ob Scandaten-Information, die in den MPEG2-Videoströmen enthalten sind, aktualisiert werden sollen. = This controls whether to update the scan data information contained in the MPEG-2 video streams. $</s>$ Die Sockets, die im write Array aufgelistet sind, werden daraufhin überwacht, ob ein Schreibvorgang den Socket blockiert. = } \\ \addlinespace[3pt] \hline
\addlinespace[2pt]
\multicolumn{2}{l}{\textbf{\underline{Translation:}}} \\
\multicolumn{2}{l}{The sockets listed in the write array will be watched for whether a write operation blocks the socket.} \\ \addlinespace[3pt] \hline
\addlinespace[2pt]
\multicolumn{2}{l}{\textbf{\underline{Reference Label:}}} \\
\multicolumn{2}{l}{The sockets listed in the write array will be watched to see if a write will not block.}  \\ \addlinespace[3pt] \hline
\addlinespace[4pt]
\textbf{\gls{qe}} 67.59, \textbf{BLEU score (using reference label):} 52.89 & 
\multicolumn{1}{c}{} \\ \addlinespace[3pt] \hline
\end{tabular}}
\caption{\textbf{An example of selected \glspl{ice} for a source text, its corresponding translation, reference label, and \gls{qe} estimation} compared to the BLEU score computed based on the reference label.}
\label{AMTA:example of mode1}
\end{table}

\section{Related work}\label{AMTA:sec:related_work}
\paragraph{\gls{icl} for \gls{mt}.} \gls{icl}\footnote{Also referred to as the prompt retrieval method} represents a relatively new paradigm in natural language understanding. Unlike traditional fine-tuning approaches, where a PLM undergoes parameter updates using a specific dataset, \gls{icl} typically directly generates the output without any modification to its parameters \cite{Radford2019LanguageMA,NEURIPS2020_1457c0d6}. This is achieved by solely providing the model with a few examples, known as \glspl{ice}, which prime the PLM to enhance its performance for the given task~\cite{10.1162/tacl_a_00324}. %The primary role of ICEs is to prime the PLM, enhancing its performance to generate more favorable results in accordance with the given task \cite{10.1162/tacl_a_00324}.

As shown by~\citeA{vilar-etal-2023-prompting}, the quality of translation is directly proportionate to the quality of \glspl{ice}, where quality refers to \glspl{ice} being relevant, clear, accurate, and domain-specific. However, considering all \glspl{ice} during processing is computationally demanding~\cite{alves-etal-2023-steering}. Hence, it is crucial to selectively choose \glspl{ice} that can enhance \gls{mt} quality. %, particularly considering the limited input capacity of LLMs. Various studies have explored different strategies in MT to address these challenges. 
\citeA{goyal-etal-2022-flores} conducted a study where \glspl{ice} were randomly selected. Despite finding that this random selection of \glspl{ice} resulted in good translation performance, the neglect of their order, which was identified as important~\cite{liu-etal-2022-makes,lu-etal-2022-fantastically}, was a drawback in this approach. To address this, methodologies such as~\cite{agrawal-etal-2023-context} introduced a re-ranking technique (R-BM25). However, their methodology relies solely on n-grams to order examples, which can enhance fluency but may overlook contextual factors. In our approach, we investigated the unigram order of initial \glspl{ice} provided by the BM25 algorithm. We leave the in-depth analysis of \gls{ice} order for future work. Additionally, \citeA{m-etal-2023-ctqscorer} highlighted the advantages of using multiple features in \gls{ice} selection to improve translation quality, while our \gls{qe}-based approach simplifies \gls{ice} selection without the need to generate additional features, ensuring efficiency.

\paragraph{\gls{qe} in \gls{mt} evaluation.} \gls{qe} models offer a quick solution to the assessment of the overall usefulness of translated text. %, typically produced by an MT system.\footnote{Given that a QE system takes as input the source and its translation, it is not strictly required that the translation is generated by an MT system; however this is the standard use-case of QE.} 
These models do not rely on reference translations, thereby reducing the human effort required for quality evaluation~\cite{tamchyna-2021-deploying, murgolo-etal-2022-quality, zerva-etal-2022-findings,blain-etal-2023-findings}.
Similar to \gls{mt} models, previous studies highlight the importance of domain-specific \gls{qe} for accurately estimating translation quality across diverse domains~\cite{lee-2020-two, sharami-etal-2023-tailoring}. This is why, in our work, we employed a domain-specific \gls{qe} model instead of a generic one to enhance the selection of \glspl{ice}. %By doing so, we aimed to improve the quality estimation of domain-specific translations, thereby enhancing overall translation performance.

Integrating \gls{qe} into \gls{icl} offers significant, yet largely unexplored, potential. %QE can quantify the fluency and adequacy of translations without needing reference translations. 
\gls{qe} can also better capture out-of-domain gender and word-sense-disambiguation errors~\cite{dinh-niehues-2023-perturbation}. Additionally, integrating \gls{qe} can mitigate reference bias, a significant challenge in accurately estimating the output quality of LLMs%, particularly in sequence transduction tasks 
~\cite{goyal2023news,raunak-etal-2023-gpts}. The introduction of COMET-\gls{qe} \cite{raunak-etal-2023-dissecting} exemplifies this pursuit, providing a metric tailored to evaluate the quality of perturbed prompts provided to GPT-3 \cite{NEURIPS2020_1457c0d6}, aiming to mitigate reference bias. While in our approach, we employ domain-specific \gls{qe} to guide the selection of \glspl{ice}, this underscores the potential of \gls{qe} in refining LLM inputs (i.e., \glspl{ice}). %Therefore, leveraging QE within ICL has the potential to streamline ICE selection, thereby ensuring improved translation quality and efficiency in LLM-based translation systems.
\vspace*{-2.0mm}
\section{Conclusion}\label{AMTA:sec:conclusion}\vspace*{-1.0mm}
We propose a novel \gls{icl} methodology for enhancing the translation capabilities of large language models (LLMs) while optimizing computational resources. Our approach leverages domain-specific \gls{qe} to guide in-context selection, particularly focusing on determining the suboptimal number and the combinations of \glspl{ice}. This novel strategy moves beyond the conventional reliance solely on translation references from development sets seen in prior methods. %Specifically, our methodology comprises two main phases: (1) employing an unsupervised retriever, namely BM25, to identify relevant examples, and (2) utilizing a search algorithm guided by the QE model to select a combination of examples that enhance translation quality.

We evaluated our approach across different modes and early stopping patience values on the German-to-English IT dataset. Our experiments consistently showed the superior performance of our methodology, surpassing all prior works across both BLEU and COMET metrics. Our method consistently improves BLEU scores, although this comes at the cost of increased computation time. We also investigated the impact of ordering the \glspl{ice} based on their unigram overlap with the source text and found it to be not statistically significant. Furthermore, our experiments highlighted the value of \gls{icl} compared to fine-tuning a \gls{plm}, namely mBART-50. We also highlighted that our method leads to less carbon emissions while achieving better translation performance.
%, gaining insights into its performance across various scenarios.
%, including the selected baseline model, 

In the future, we would like to conduct further research on the impact of our proposed methodology across different datasets and LLMs. Also, we aim to explore alternative metrics beyond BLEU to tailor the selection process, as well as additional features such as bigram, type/token ratio, and length when ordering examples prior to their input into LLMs. %Additionally, a comparison between the fine-tuning of the XGLM (using the parallel dataset) and the proposed ICL methodology could be conducted.  
% \vspace*{4.0mm}
%Several avenues for future research stem from our methodology and findings. Firstly, assessing the generalizability of our methodology by testing it on additional datasets and LLMs could provide valuable insights. Secondly, exploring alternative metrics beyond BLEU to tailor the selection process based on specific quality labels other than BLEU could be beneficial. This also extends to Mode 2, where incorporating additional features (such as bigram, type/token ratio, and length) to prioritize the order of ICEs is feasible. Lastly, a comparison between the fine-tuning of the XGLM (using the parallel dataset) and the proposed methodology could be conducted.

% \newpage
%----------------------–––––––––––––
\chapterseparatorpage
\chapterimage{Assets/separator.pdf}

\chapter[Discussion]{Discussion}
\chaptermark{Conclusion}

Despite the tremendous progress in \gls{mt} and the remarkable evolution of \glspl{llm}, these technologies are not yet capable of meeting the demands of domain-specific translation out of the box. While \glspl{llm} have become more fluent and context-aware, their ability to operate effectively in specialized domains remains limited. This shortcoming primarily stems from their reliance on generic-domain data, which lacks the specificity required to handle the unique terminology, formal registers, and contextual nuances of fields like medicine, law, or information technology. The implications of such domain mismatch are not merely academic; in real-world applications, mistranslations can carry serious consequences, from clinical misunderstandings to contractual misinterpretations. Importantly, this issue is not confined to \glspl{llm} alone: in \gls{mt} more broadly, relying on ever-larger generic corpora often improves overall translation quality but also risks embedding irrelevant or domain-misaligned patterns, thereby limiting reliability in specialized contexts. The challenge, then, is not simply about increasing data volume but ensuring domain relevance and accuracy.

This inaccuracy in \gls{mt} systems is primarily driven by a phenomenon known as domain mismatch. Domain mismatch arises when the data used to train a model comes from a different domain than the one in which it is used. This mismatch often results in misaligned vocabulary, stylistic inconsistencies, and contextually inappropriate translations. The impact extends beyond \gls{mt} output, also affecting the accuracy of \gls{qe}, which depends on consistent reference domains to generate reliable assessments. Despite considerable research efforts, domain mismatch remains a persistent and complex barrier to achieving high-quality, dependable \gls{mt} and \gls{qe} performance in real-world, domain-sensitive applications.

This dissertation set out to explore how \gls{mt} and \gls{qe} systems can be adapted to function more effectively across various domains, particularly in specialized fields such as healthcare, law, and IT where accurate translation is critical. In our work, a domain refers to the source of a dataset. Improved model performance on such data is treated as \gls{da}, reflecting real-world cases where domain-specific and generic content are often intermixed.

The central motivation arose from the observation that general-purpose \gls{mt} systems—especially \glspl{llm} trained on general-domain corpora—struggle to accommodate the specialized terminology, stylistic conventions, and contextual nuances of professional domains. A mistranslated medical instruction or legal clause is not just a linguistic error but a potentially harmful or costly one. This highlights a foundational insight: that carefully selected, domain-specific data can improve translation quality more effectively than adding large amounts of generic text.

At the heart of the challenge is domain mismatch, where the disconnect between training and deployment domains leads to vocabulary mismatches and contextually inappropriate outputs. This problem extends beyond \gls{mt} to the realm of \gls{qe}, where domain-inconsistent references undermine the accuracy of quality assessments. Building robust \gls{mt} and \gls{qe} systems, therefore, requires a comprehensive strategy that accounts for not just the data itself, but how it is selected, represented, tokenized, and evaluated.

To address and better understand these challenges, this dissertation proposes a multi-faceted approach: domain-specific data selection strategies to improve translation quality even in low-data scenarios (Chapter~\ref{chap:CLIN}); a combination of \gls{da} and \gls{dag} techniques to boost \gls{qe} performance in low-resource and multilingual settings (Chapter~\ref{chap:EAMT}); an evaluation of \gls{sw} tokenization and vocabulary configurations, which significantly influence model efficiency and translation quality during fine-tuning––a common method for \gls{da} (Chapter~\ref{chap:BPE}); and a novel \gls{icl} strategy that leverages domain-specific \gls{qe} models to guide \gls{ice} selection, enabling accurate translation with generative \glspl{llm} without fine-tuning, thereby reducing computational costs (Chapter~\ref{chap:AMTA}).

Our work also highlights the interdependence of \gls{mt} and \gls{qe} in addressing domain mismatch, stressing that high-quality translation and accurate evaluation must evolve together. By treating them as mutually reinforcing, it offers an integrated approach to \gls{da}. The result is a set of methods and insights aimed at improving performance while promoting scalable, efficient, and sustainable practices. Computational cost details are included throughout, with a focused breakdown in the appendix.

\section{Summary of findings}

This dissertation addressed four core \glspl{rq}, each targeting a specific aspect of \gls{da} in \gls{mt} and \gls{qe}. To reiterate, the main \gls{rq} guiding this work was:

\textit{How can we design \gls{mt} and \gls{qe} systems that are accurate, adaptable, and efficient across specialized domains?}

To answer this overarching question, four sub-questions were formulated. These are revisited and answered below.

\subsection*{\textbf{RQ1: What is the optimal amount of in-domain data required to achieve state-of-the-art \gls{mt} quality at low computational costs?}}
\begin{itemize}
\item \textbf{RQ1a:} How does the quality of selected in-domain data affect translation performance?
\item \textbf{RQ1b:} What trade-offs arise between translation performance and computational cost when using in-domain versus generic-domain data?
\end{itemize}

\noindent To address RQ1—determining the optimal amount of in-domain data needed to achieve high translation quality with low computational cost—we introduced a comprehensive and adaptable ranking-based data selection method in Chapter~\ref{chap:CLIN}. This approach tackles both RQ1a and RQ1b by enabling the extraction of semantically similar sentence pairs from large general-domain corpora. The method uses a modular pipeline that integrates contextual sentence embeddings, semantic similarity search, and in-domain relevance ranking. It is designed to be scalable and generalizable across languages and domains.

\paragraph{Findings related to RQ1a.} Our findings demonstrate that the quality of selected in-domain data plays a critical role in translation performance. Even relatively small, carefully selected subsets of semantically relevant data significantly outperformed generic-domain baselines. In some cases, models trained on these selected subsets rivaled—or even exceeded—the performance of models fine-tuned on full in-domain corpora (vs. our approach that trained from scratch). This underscores a key insight: data relevance can outweigh data volume.

These findings align with and build upon recent research. Following our work, \citeA{eronen2023improvingpolishenglishneural} showed that using related language data or domain-specific subsets is especially effective in low-resource settings, such as Polish–English translation. Likewise, \citeA{chakrabarty-etal-2020-improving} stressed the importance of evaluating linguistic relevance in feature selection, cautioning against blindly incorporating features without understanding their value. Similarly, \citeA{wu-etal-2024-importance} found that token-level relevance can enhance translation quality more effectively than simply adding more data. While our work shares this emphasis on relevance, it takes a different approach: rather than focusing on individual tokens, we rely on contextual similarity at the sentence level.  Our results extend these insights, showing that domain-relevance-driven selection at the sentence level can match or even surpass in-domain fine-tuning—without relying on \glspl{plm}, unlike the prior work discussed in Chapter~\ref{chap:CLIN}—and with significantly lower computational cost. 

For instance, our method achieved a BLEU score of 31.0 using just 1 million carefully selected sentences, whereas training with 30 times more mixed-domain (in-domain and out-of-domain) data yielded only 26 BLEU. Of course, this result is for a domain-specific test set; for generic-domain or out-of-domain evaluation, such a small specialized model would likely be less effective. Moreover, reaching this level of quality with our approach required approximately 4 hours of training, compared to 10 hours needed to train on 30 million sentences to achieve lower performance. These results highlight that smarter data selection can be more impactful than scale alone, enabling both high-quality and resource-efficient \gls{mt}.

\paragraph{Findings related to RQ1b.} In addressing RQ1b, we demonstrated that our method—selecting in-domain data based on contextual relevance and using it to generate synthetic data—not only enhances translation quality but also does so efficiently. By restricting training to top-ranked, semantically relevant data, the approach significantly reduced computational time and resource consumption without sacrificing performance. Although the optimal data size varied by domain, our results (referred to as the ``Mixing Effect'') consistently revealed a saturation point: beyond this threshold, additional data provided diminishing returns and, in some cases, even led to degraded performance. This degradation in quality stems from the growing inclusion of less relevant data as the selection pool widens—especially when extracting from out-of-domain sources that are not large or diverse. In essence, as quantity increases, contextual similarity—and therefore quality—tends to decline. 

These findings—particularly the insight that translation quality can actually decrease after a certain point—are consistent with earlier work \cite{poncelas-etal-2018-investigating,moslem-etal-2022-domain,mittal-etal-2023-leveraging}, which observed that the benefits of adding more domain-specific data eventually level off. While this pattern has been noted before, it often was not examined in much detail. Our work not only confirms this effect but also offers a clearer explanation: as we continue expanding the dataset—especially from out-of-domain sources that are not very large or diverse—the added data becomes less relevant. This dilution reduces contextual similarity, introduces noise, and ultimately hurts performance. Jointly, these findings highlight an important takeaway: it is not just about having more data, but about having the \emph{right} data. Thoughtful, relevance-based selection can lead to better results, faster training, and more efficient use of resources.

Ultimately, our work in Chapter~\ref{chap:CLIN} presents a clear and effective strategy for improving \gls{mt} in domain-specific, low-resource settings. By prioritizing relevance over volume, we demonstrated that modest, well-selected datasets can yield high-quality translations at significantly lower computational cost. Importantly, our open-source implementation supports reproducibility and flexibility, allowing others to replicate, customize, and extend the method across domains and languages—reinforcing the broader value of targeted data selection in building efficient and accurate \gls{mt} systems.

To support further use and development, we released an open-source toolkit with examples and step-by-step instructions, available at:
\url{https://github.com/JoyeBright/domain-adapt-mt}.

\subsection*{\textbf{RQ2: What are the effects of combining \gls{da} and \gls{dag} on the performance and generalizability of \gls{qe} models across different domains and languages?}}
\begin{itemize}
\item \textbf{RQ2a:} How does \gls{da} impact \gls{qe} performance in low-resource scenarios?
\item \textbf{RQ2b:} What challenges arise when applying \gls{dag} in \gls{qe} across diverse languages?
\end{itemize}

\noindent This dissertation addressed RQ2 by designing and evaluating a \gls{qe} training pipeline aimed at tackling two major challenges in \gls{qe}: data scarcity and domain mismatch. These are common hurdles that often limit the effectiveness and reliability of \gls{qe} systems. Our proposed solution combined \gls{da} with a lightweight \gls{dag} strategy, resulting in a modular and flexible approach that improves both performance and generalizability.

To briefly recap, recent \gls{qe} models—like \gls{mt} models—are neural-based and heavily data-driven. When a \gls{qe} model trained on one dataset is applied to another, mismatches in domain can significantly affect performance. This challenge is further compounded by the fact that \gls{qe} requires not only the source and machine-translated output but also a post-edited reference—a resource-intensive and costly prerequisite. This led us to first confirm a foundational assumption: \gls{qe} models, like \gls{mt} models, must be domain-specific to perform well. From there, we addressed the data scarcity issue by proposing methods to augment or generate synthetic data to bridge the gap.

\paragraph{Findings related to RQ2a.} Our findings show that a customized version of the ``mixed fine-tuning + fine-tuning'' strategy proposed by~\citeA{chu-etal-2017-empirical} significantly improves \gls{qe} performance in low-resource settings. We adapted this strategy for \gls{qe} for the first time by initially fine-tuning a pre-trained XLM-R model on a large out-of-domain (OOD) \gls{qe} dataset to build a strong, generic baseline. In the second stage, we applied a mixing step that combined this OOD data with in-domain (ID) data—both authentic and synthetic—using simple yet effective \gls{dag} techniques. These included multilingual ID concatenation and synthetic data generation via back-translation and \gls{qe} label prediction, as detailed in Section~\ref{EAMT:sec:DAG} of Chapter~\ref{chap:EAMT}. This step helped the model retain general \gls{qe} knowledge while gradually adapting to the target domain. In the final stage, the model was fine-tuned solely on a small, high-quality ID dataset to sharpen its domain-specific performance. This structured training pipeline consistently led to statistically significant gains across all tested language pairs and domains.

These findings align with earlier work that advocated for intermediate fine-tuning steps between pre-training and downstream \gls{qe} tasks \cite{rubino-2020-nict,rubino-sumita-2020-intermediate,lee-2020-two}. However, our pipeline extends this line of research by combining \gls{dag} with domain-specific fine-tuning. Unlike approaches that rely on large amounts of labeled data, our method offers a more efficient path to high performance, especially in real-world scenarios where such data is limited.

\paragraph{Findings related to RQ2b.} In addressing RQ2b, we examined the effects of \gls{dag} across different languages and domains. Overall, augmentation was found to be beneficial, contributing to improved model robustness and generalization. We observed variation in performance across language pairs and settings, with some cases showing limited or no improvement. These outcomes are consistent with prior concerns about potential mismatches between synthetic and real-world \gls{qe} distributions \cite{2022arXiv221210257Q}, particularly when the synthetic data is noisy or poorly curated. Unlike earlier \gls{dag} methods that rely on classifier-based or pseudo-labeled pipelines \cite{baek-etal-2020-patquest,freitag-etal-2021-results}, our approach relied on simpler techniques—such as back-translation and domain-aware sampling—which yielded promising results in multiple settings.

We also found that integrating domain tags during training consistently improved performance. These tags helped the model distinguish between different types of data and reinforced domain-specific cues. This observation is in line with prior work in both \gls{mt}~\cite{sennrich-etal-2016-controlling,Chu2019MultilingualMA} and multi-domain \gls{qe} \cite{fomicheva-etal-2022-mlqe,zerva-etal-2022-findings}, confirming that such techniques are transferable and valuable in \gls{qe} settings.

The generalizability of the proposed approach was further validated through cross-lingual and \gls{zs}—where the model had no exposure to the target language during training—evaluations. The pipeline was tested on a diverse set of language pairs from the WMT21 \gls{qe} shared task, including English--German, --Chinese, --Italian, --Japanese, --Czech, as well as Romanian--English and Russian--English. Even under \gls{zs} conditions, the first stage of the pipeline outperformed most baselines. Notably, the model trained on Romanian--English performed particularly well on its own test set and transferred effectively to English--Japanese. This may be attributed to the model’s exposure to English--Italian data during the first training stage, as Italian and Romanian share typological similarities. These findings suggest that linguistic proximity can support stronger cross-lingual transfer.

In summary, combining \gls{da} and \gls{dag} within a well-structured, staged training process leads to robust \gls{qe} models capable of strong performance across domains and languages. Importantly, the proposed methodology offers a practical and scalable solution for real-world scenarios, where annotated \gls{qe} data is limited or unavailable for many languages and domains.

\subsection*{\textbf{RQ3: What is the optimal combination of \gls{sw} tokenization model source and vocabulary source to maximize translation quality and efficiency during fine-tuning?}}
\begin{itemize}
  \item \textbf{RQ3a:} How do different data sources for the \gls{sw} tokenization model affect translation quality when fine-tuning out-of-domain models on in-domain data?
  \item \textbf{RQ3b:} How do the sources of the \gls{sw} tokenization model and vocabulary impact the trade-off between fine-tuning efficiency (e.g., training time, resource use) and translation accuracy?
\end{itemize}

\noindent While fine-tuning \glspl{plm} for \gls{da} has shown substantial performance gains~\cite{luong-etal-2015-effective,dakw:17fine}, its effectiveness depends heavily on the representation of domain-specific inputs. This not only involves model architecture and data selection, but also the \gls{sw} tokenization strategy and vocabulary. Prior work has identified vocabulary mismatches between base and target domains as a major obstacle to effective \gls{da}~\cite{lim2018exploring,sato-etal-2020-vocabulary,lim2024crossdomainsemanticsegmentationinconsistent}, and proposed embedding-based solutions, such as ``vocabulary adaptation,'' projecting target-domain embeddings into pretrained embedding spaces to improve performance by 3–4 BLEU points~\cite{sato-etal-2020-vocabulary}.

Despite this, the joint effect of \gls{sw} tokenization and vocabulary choice during fine-tuning remains underexplored. While studies have focused on embedding layer adjustments~\cite{sato-etal-2020-vocabulary} or tuning \gls{sw} granularity~\cite{salesky2018optimizingsegmentationgranularityneural}, few have systematically evaluated how tokenization methods and vocabulary sources interact under consistent fine-tuning configurations.

To address RQ3, we empirically evaluated seven consistent configurations that combine different sources for \gls{sw} tokenization models (pre-training, in-domain, or combined) and vocabulary (pre-training, in-domain, or combined) in an English–German \gls{da} setting.

\paragraph{Findings related to RQ3a.}
Our results show that the choice of tokenization model and vocabulary source substantially influences translation quality. Using both BPE and vocabulary from in-domain data yields the highest BLEU, chrF2, and best TER scores and the most effective adaptation to the domain. However, this configuration also introduces the most new tokens and requires the longest training time. In contrast, mismatched setups (e.g., a configuration with in-domain BPE and a vocabulary from the pre-training data) result in significant quality drops, highlighting the need for coherent \gls{sw} and vocabulary alignment. Hybrid configurations, which combine resources from both pre-training and in-domain data, offer practical alternatives that balance adaptation effectiveness and resource demands. These findings reinforce previous work~\cite{sato-etal-2020-vocabulary}, showing that effective vocabulary adaptation must be closely tied to the chosen tokenization model.

\paragraph{Findings related to RQ3b.}
We observed a clear trade-off between translation quality and computational efficiency. Configurations that reused pre-training resources were computationally efficient but lagged in domain accuracy, while purely in-domain setups improved quality at the expense of training speed and resource use. The most balanced results came from hybrid approaches, which combine pre-training and in-domain resources (i.e., using either the BPE model or vocabulary, or both, derived from a mix of pre-trained and in-domain data)––and achieve strong results while reducing computational costs compared to full in-domain setups. 

In summary, these results demonstrate that carefully aligning both \gls{sw} tokenization and vocabulary source with the fine-tuning data substantially enhances adaptation quality, while hybrid configurations can offer a good balance between domain performance and generalization. In practice, one might prefer hybrid setups when the goal is not only to maximize in-domain accuracy but also to retain robustness across out-of-domain or mixed-domain scenarios. These findings highlight the practical importance of consistent \gls{sw}–vocabulary design in effective fine-tuning for \gls{da}. Naturally, our study has certain limitations––such as the exclusive focus on MT, the medical domain, the use of BPE only, and the lack of hyperparameter tuning––which were discussed in detail in Section~\ref{BPE:limitations}.

\subsection*{\textbf{RQ4: How effective are domain-specific \gls{qe} models in determining effective \glspl{ice} for improving \gls{mt} quality in generative \glspl{llm}?}}
\begin{itemize}
\item \textbf{RQ4a:} What criteria should be used to select effective \glspl{ice} for improving \gls{mt} quality?
\item \textbf{RQ4b:} How does the integration of \gls{qe} affect the success of \gls{ice} selection?
\end{itemize}

\noindent \glspl{llm} have demonstrated strong performance across a wide range of \gls{nlp} tasks, including \gls{mt}. Unlike traditional approaches, \glspl{llm} can perform translation without further training by conditioning on a well-crafted input prompt. Prior research has shown that providing additional context—specifically, \glspl{ice}—can substantially enhance translation quality. \glspl{ice} serve to prime the model, enabling it to better focus on and adapt to the target task. However, the effectiveness of \glspl{ice} is influenced by multiple factors, such as the domain of the source text, the number and order of examples, and the prompt format. As there is no universally optimal configuration, selecting effective \glspl{ice} requires a nuanced understanding of how these variables affect translation performance, which is typically evaluated using reference translations or human judgment—both of which are often unavailable or expensive to obtain.

Motivated by the demonstrated advantage of domain-specific \gls{qe} over generic \gls{qe} in Chapter~\ref{chap:EAMT}, and recognizing the absence of a universal strategy for optimal \gls{ice} selection across different \glspl{llm}, we proposed using domain-specific \gls{qe} as a feedback mechanism to guide the selection of \glspl{ice}. To address RQ4, we developed a novel \gls{icl} methodology for \gls{mt} that leverages domain-specific \gls{qe} to inform \gls{ice} selection. This approach was designed to maximize translation quality while minimizing computational cost, without relying on reference translations. 

\paragraph{Findings related to RQ4a.} We demonstrate that the quality, domain relevance, and contextual fit of \glspl{ice} are critical for achieving strong \gls{mt} performance in \glspl{llm}. We show that domain-specific \gls{qe} serves as an effective criterion for selecting \glspl{ice}, consistently outperforming conventional methods such as random selection~\cite{sia-duh-2023-context}, n-gram similarity-based retrieval~\cite{agrawal-etal-2023-context}, and fixed retriever rankings (e.g., R-BM25). Although prior work has found random or n-gram-based selections to be surprisingly competitive, our results suggest that these approaches tend to emphasize surface-level lexical similarity while overlooking deeper semantic and domain-specific signals essential for high-quality translation. Notably, our \gls{qe}-guided approach consistently achieved higher BLEU and COMET scores on German--English IT translation tasks compared to both these \gls{icl} baselines.

Furthermore, our findings suggest that \gls{icl} offers substantial efficiency benefits. Unlike fine-tuning \glspl{plm} such as mBART-50—which requires considerable computational resources—our \gls{qe}-guided \gls{icl} strategy, and \gls{icl} strategies more broadly, achieve competitive results at significantly lower cost, making them practical and scalable solutions for real-world multilingual \gls{mt} applications.

We also examined whether the order of \glspl{ice} influences translation quality by testing reordering based on unigram overlap. While earlier studies emphasized the importance of example ordering~\cite{liu-etal-2022-makes,lu-etal-2022-fantastically}, our results showed that this form of reordering had minimal impact. This suggests that what matters most is not how lexically similar they are to the input, but whether they are semantically relevant and domain-appropriate. In other words, examples that make sense for the task at hand—and match the style or terminology of the input—carry more weight than those chosen purely for their surface similarity. %These findings offer a more grounded understanding of ICL prompt design, emphasizing that thoughtful content selection is more important than ordering heuristics

\paragraph{Findings related to RQ4b.} In addressing RQ4b, we found that integrating \gls{qe} into \gls{ice} selection not only improves translation quality but also enhances efficiency—resulting in faster inference, reduced memory usage, and greater scalability, which are particularly beneficial given token constraints in \glspl{llm}. Although \gls{qe} may occasionally misrank examples, our method consistently selected fewer yet more impactful \glspl{ice}. This stands in contrast to prior approaches that include a larger number of examples without assessing their individual contribution to translation quality~\cite{goyal-etal-2022-flores}. Additionally, our method produced translations with lengths more closely matching the references, which can reduce downstream post-editing effort. Finally, since \gls{qe} models estimate quality without relying on reference translations, they offer a promising direction for mitigating reference bias—a known limitation in \gls{llm} evaluation pipelines~\cite{raunak-etal-2023-dissecting}. Moreover, by modeling quality beyond surface-level features, \gls{qe} may help identify domain-specific issues such as word sense ambiguities or terminology mismatches—though further empirical validation is needed to confirm this potential.

These findings confirm that \gls{qe}-guided \gls{icl} offers a practical and scalable alternative to fine-tuning. In contrast to traditional methods requiring parameter updates or extensive reference data, our approach maintains model generalization while enhancing domain-specific translation quality through selective, high-impact \glspl{ice}.

\section{Limitations and future work}
In this section, we reflect on some general limitations and potential directions for future research that cut across the different parts of this dissertation. We also highlight chapter-specific considerations to provide a more grounded view of where the work could be expanded or refined. While each chapter already includes its own analysis and discussion, what follows is meant to offer a broader perspective and point toward ideas worth exploring further. For more detailed insights, see chapters \ref{chap:BPE}–\ref{chap:AMTA}.

\label{Chap:Conclusion:limit}

\subsection*{General limitations and future directions}
While this dissertation offers a broad and in-depth look at \gls{da}, \gls{sw} modeling, \gls{qe}, and \gls{icl} in \gls{mt}, a few practical constraints inevitably shaped its scope. One notable limitation is the heavy reliance on automatic evaluation metrics such as BLEU, COMET, and TER. These tools are widely used and provide a consistent framework for comparison, but they can only go so far. Human evaluations—though far more insightful when it comes to subtle aspects of translation quality—are costly and time-intensive. Neural-based metrics like COMET, for example, show strong correlation with human judgments~\cite{rei-etal-2020-comet}, but can still carry biases from the data they were trained on, potentially skewing results in cross-domain scenarios~\cite{zouhar-etal-2024-pitfalls}. Therefore, incorporating human assessments would undoubtedly enrich future work by providing a more nuanced understanding of translation quality. That said, such efforts require dedicated resources and were beyond the reach of this dissertation.

Another consideration is the selection of language pairs and domains. While the chosen languages were carefully selected to reflect meaningful linguistic and domain diversity, as well as considerations such as reproducibility and the findability/availability of suitable datasets, the findings would benefit from being tested on an even broader set—especially low-resource or typologically distinct languages. Of course, this is a common limitation in \gls{mt}: it is almost impossible to cover every language and domain, no matter how ambitious the study.

Methodologically, keeping components like tokenization models or retrieval strategies constant was a deliberate choice to ensure clarity and control in experimentation. But future studies for example might try out semantic matchers beyond SBERT to improve generalization. 

Lastly, while hyperparameter tuning was carefully performed, there is still room for refinement. Additional optimization—either automated or task-specific—may further enhance performance across the board.

\subsection*{Chapter-specific limitations and directions}

\textbf{Chapter~\ref{chap:CLIN}}: This chapter introduced a similarity-based data selection method that retrieves a fixed number ($n$) of nearest neighbors (contextually similar examples) for each input example. The method performed well in practice and offered a simple, consistent mechanism for selecting relevant context. However, relying on a fixed number assumes that the top $n$ examples are equally informative across all inputs—which may not always hold true. In some cases, the distribution of similarity scores varies considerably, or, in less ideal scenarios, the generic dataset may not contain enough truly relevant examples. As a result, the model may fall back on less similar, and potentially less useful, instances.

This choice was made deliberately to maintain interpretability and experimental control. That said, there is room for further development. One promising direction would be to experiment with dynamic similarity thresholds that adjust to each input's specific distribution. Naturally, this introduces the challenge of defining an appropriate threshold for each case—a task that becomes an optimization problem in itself—but it is a challenge worth exploring. Future work could also look into alternative similarity metrics or integrate the selection process more tightly with \gls{mt} training, potentially enabling more precise and domain-aware translations.

\textbf{Chapter~\ref{chap:EAMT}}: The proposed \gls{qe} pipeline was validated on several multilingual benchmarks and showed consistent performance gains. Nonetheless, further experiments on additional OOD datasets and with alternative backbone models could help better assess the robustness and generalizability of the approach.

As the field continues to shift away from traditional automatic metrics like TER toward richer annotation frameworks such as Direct Assessment~\cite{graham-etal-2013-continuous} and Multi-dimensional Quality Metrics (MQM)~\cite{lommel-etal-2014-using,freitag-etal-2021-experts}, adapting the pipeline to work with these more expressive and human-centered labels would strengthen its practical relevance. That said, this transition comes with notable challenges: Direct Assessment relies on costly human ratings at the sentence level, and MQM demands detailed linguistic annotation, which is time-intensive and requires domain expertise. As discussed earlier in the general limitations, incorporating human evaluation at scale remains a persistent challenge due to cost and effort.

Another promising direction for future work involves exploring alternative strategies to preserve pre-trained linguistic knowledge, without relying on explicit quality tagging. One such approach is to freeze the \glspl{plm} and leverage its internal representations directly for quality prediction. This method helps maintain the original distribution of information across layers, where each layer typically encodes different linguistic phenomena—such as morphology, syntax, or semantics. In doing so, it can reduce the risk of overwriting useful generalizations during fine-tuning, while also lowering training overhead and making the pipeline more lightweight and scalable.

However, this comes with trade-offs. Without fine-grained, task-specific supervision, the model may prioritize general semantic similarity rather than capturing detailed aspects of translation quality. As a result, important distinctions—such as adequacy or fluency—might be underrepresented in the predictions. Balancing the preservation of pre-trained knowledge with the need for quality-aware modeling remains an open challenge for future \gls{qe} research.

%\noindent 
\textbf{Chapter~\ref{chap:BPE}}: Although the experiments in this chapter were carefully designed and systematically executed, several broader limitations should be acknowledged. First, our study focused solely on \gls{mt} systems and did not extend to \glspl{llm}. While \gls{mt} models and \glspl{llm} share certain sensitivities to data processing and vocabulary choices, \glspl{llm}---particularly multilingual ones---introduce additional challenges, such as managing vocabulary across multiple languages and handling cross-lingual interference. Understanding how \gls{sw} modeling and vocabulary configurations impact \gls{llm} performance is a valuable direction for future research.

Second, the datasets used in our experiments were relatively small and drawn from a single domain (IT). This setup enabled controlled and interpretable comparisons but limits how broadly the results can be applied. We did not test our configurations on other specialized domains like medicine or law, nor did we analyze how much overlap exists between the vocabularies of training and fine-tuning datasets. Both of these factors—domain variation and vocabulary overlap—likely influence how well a model adapts to new data. Exploring these aspects would offer a more complete picture of how domain similarity and token alignment affect translation quality.

Finally, to isolate the effects of \gls{sw} modeling, we intentionally limited the scope of our experiments: we used only BPE and did not tune hyperparameters or compare alternative tokenization methods like SentencePiece~\cite{kudo-richardson-2018-sentencepiece}. While this design choice helped ensure fair comparisons, it also means we did not explore the potential gains from more adaptive approaches. Future work could broaden the experimental setup by testing different tokenizers and by developing ways to quantify how different two datasets are—such as measuring domain relevance or lexical divergence. This could help inform decisions about when existing \gls{sw} and vocabulary settings can be reused and when they need to be adapted for better domain transfer.

\textbf{Chapter~\ref{chap:AMTA}}: Our approach uses XGLM as the base \gls{llm} and applies iterative \gls{qe} scoring to select \glspl{ice}, which introduces additional complexity and increases inference time. While the trained \gls{qe} model can be reused within the same domain, its generalizability to other domains remains uncertain. Future research could explore alternative backbone architectures, develop lightweight or distilled \gls{qe} models to reduce computational overhead, or design more integrated retrieval–scoring pipelines to streamline the selection process.

To ensure a fair comparison with prior work discussed in this chapter, we employed BLEU-based \gls{qe} as the scoring signal. This decision enabled consistent evaluation and direct benchmarking. Nevertheless, future work could incorporate neural evaluation metrics such as COMET, which offer stronger alignment with human judgments and may lead to more accurate \gls{ice} selection, particularly in low-resource or semantically complex settings.

Finally, we did not investigate the effect of \gls{ice} ordering, as evaluating all possible permutations would incur substantial computational costs. However, since the \gls{qe} model provides a quality estimate for each candidate combination, it could be extended to rank and optimize example orderings more efficiently. Exploring this area remains a promising direction, especially in light of recent findings~\cite{xiang-etal-2024-addressing} that show causal \glspl{llm} are sensitive to the order of \glspl{ice}.

\section{Implications of the findings}

In this dissertation, we examined the persistent challenge of domain mismatch in \gls{mt}—a problem that not only affects \gls{mt} and \gls{qe} systems, but also extends across the broader landscape of \gls{nlp}, and even into fields like computer vision. Naturally, domain mismatch manifests differently across modalities: in vision, for example, it may involve shifts in lighting, perspective, or object appearance rather than terminology~\cite{10.1007/978-3-319-16634-6_19}. What unites these scenarios is the shared difficulty models face when deployed on data distributions that diverge from those they were trained on.

At a high level, this dissertation contributes to addressing domain mismatch in both \gls{mt} and \gls{qe} by underscoring the necessity of adaptation. Our findings align with a growing consensus in neural modeling: whether through \gls{icl} in \glspl{llm} or fine-tuning in encoder-decoder architectures (including \glspl{llm} themselves), most models require some degree of adaptation to maintain strong performance across domains~\cite{10820432,sinha2024surveycompositionallearningai}. This need has become especially evident in recent \gls{llm}-based workflows, where domain-specific prompts and carefully selected \glspl{ice} play a central role in achieving reliable results. Although our focus has been on \gls{mt} and \gls{qe}, this broader principle likely extends to many other \gls{nlp} tasks—and even further, to fields such as computer vision—where models are increasingly deployed in specialized, real-world settings.

Our results reinforce the idea that general-purpose models should not be expected to perform optimally out of the box. Instead, they must be aligned with their target domain and task context to ensure both accuracy and utility. In Chapter~\ref{chap:EAMT}, we extended this observation to \gls{qe} itself, showing that evaluation models—like the systems they assess—also require domain-specific adaptation. A \gls{qe} model trained on general data may fail to detect quality variations in specialized domains, leading to unreliable assessments. This issue also affects reference-based metrics such as COMET, which may inherit biases from their training data~\cite{zouhar-etal-2024-pitfalls}. As such, data-driven evaluation models must be applied with awareness of their domain limitations to avoid misinforming downstream decisions.

These findings contribute to a broader theoretical shift in machine learning: from model-centric to data-centric thinking. While scaling models has been a dominant strategy, our results challenge the assumption that bigger always means better. Instead, we emphasize the need to understand the nature of the data models see—and how this shapes their behavior. This is especially crucial in domain-sensitive tasks like \gls{mt} and \gls{qe}, where even small mismatches in terminology, structure, or tokenization can lead to significant drops in performance.

Of course, adaptation is not without cost. It typically requires additional training, hyperparameter tuning, and sometimes re-tokenization, all of which come with computational and environmental overhead. While some argue that scaling models and training data may eliminate the need for \gls{da}, our findings suggest that more data alone is not always the answer. In fact, we show that carefully curated, domain-specific data—whether used in prompting or fine-tuning—can yield better results than simply increasing volume. This is central to the data-centric approach we advocate throughout the dissertation.

In Chapter~\ref{chap:BPE}, we also demonstrated that the influence of data extends to upstream components like tokenization. Often considered a low-level detail, tokenization was shown to significantly impact translation quality when mismatched with the domain. Our results revealed that tokenizers trained on domain-specific data can improve output quality, especially for specialized terminology. This underscores that \gls{da} must consider the entire pipeline—from tokenization to evaluation—rather than focusing solely on model architecture.

Further, in Chapter~\ref{chap:AMTA}, we explored how \gls{qe} can be used not just for evaluation, but as an external signal to guide and refine \gls{mt} outputs. This reflects a broader shift toward feedback-driven, interactive systems where multiple components collaborate to improve performance~\cite{brivaiglesias2025aiagentsnewmachine}. Our work aligns with emerging multi-agent workflows—such as TransAgents~\cite{wu-etal-2024-transagents}—in which, for example, one agent generates a translation and others provide feedback to iteratively enhance it. These agent-based systems, often powered by \glspl{llm}, point to the growing potential of integrating tools like \gls{qe} models or glossaries into real-time translation workflows. While our implementation is research-oriented, it signals a promising direction for scalable, adaptive, and collaborative \gls{mt} systems.

A key methodological implication of this work is the value of building modular, reusable tools that support data-centric workflows. One such example is the domain-specific data selection tool introduced in Chapter~\ref{chap:CLIN}. Originally designed to retrieve the top-$k$ most similar examples for \gls{icl}, the tool has since evolved to support more flexible configurations—including the ability to surface dissimilar examples, which makes it useful for contrastive learning setups such as triplet loss.

What makes this tool particularly impactful is its generalizability. It is designed to be multilingual, language-agnostic, and easily adaptable, making it relevant well beyond the \gls{mt} use case. Whether for selecting examples in \gls{icl}, generating negatives for contrastive training, or supporting tasks like \gls{dag}, retrieval, or representation learning, the tool serves as a practical building block for a range of domain-aware applications. More broadly, it reflects a shift toward developing adaptable, context-sensitive components that help researchers and practitioners work more effectively across domains—without having to start from scratch each time.

Finally, it is important to acknowledge the very real costs—computational, financial, and environmental—associated with building domain-specific systems. Our work shows that while \gls{da} is often necessary for achieving reliable performance, it adds to the already significant demands of pretraining large models. Given the substantial investments of time, data, and compute that go into developing massive, general-purpose \glspl{llm}, this raises a natural question: why not design these models to be more adaptable or better aligned with domain-specific needs from the start? Of course, building truly one-size-fits-all models may still be out of reach. However, our findings suggest that some of the effort currently devoted to making models broadly generalizable could be more usefully redirected toward making them flexible by design. This shift could reduce the need for repeated adaptation, particularly as model sizes—and their associated costs—continue to grow. Encouragingly, we are already seeing signs of this in today’s commercial \glspl{llm}, which often perform reasonably well in specialized domains with minimal additional effort, especially compared to what was required just a few years ago.

To support transparency and reproducibility, we provide a detailed, chapter-by-chapter breakdown of the computational costs involved in our experiments in the appendix.

\section{Wrapping up}

This dissertation addressed the core challenges of domain mismatch. Despite the remarkable advancements achieved by large-scale, \glspl{plm}, our findings illustrate that these systems frequently fall short when applied directly to specialized or out-of-domain contexts—particularly in tasks like \gls{mt}, where terminology and structure can vary significantly across domains. Through extensive experimentation and analysis, we demonstrated that reliable performance in real-world, domain-sensitive applications demands deliberate adaptation—whether through fine-tuning, targeted data selection, or informed use of \gls{icl} guided by domain-specific \gls{qe}.

Beyond technical outcomes, this dissertation contributes to a broader theoretical shift in the field, moving away from purely model-centric solutions towards a more data-driven perspective. We have shown repeatedly that simply scaling up models or datasets does not automatically guarantee improved performance. Instead, the careful selection and preparation of domain-specific data—ranging from the choice of examples used in prompts to the nuances of vocabulary and tokenization strategies—play a crucial role in model accuracy and efficiency. This shift underscores the importance of data quality and domain awareness in achieving high performance.

In addition to these theoretical insights, we developed practical methodologies and publicly available tools—such as the domain-specific data selection system introduced in Chapter~\ref{chap:CLIN}. These resources are designed to help researchers and practitioners efficiently implement \gls{da} strategies across languages, domains, and tasks. By emphasizing transparency and reproducibility—highlighted by the detailed computational analyses provided in the appendix—we aim to facilitate future work that is both effective and environmentally sustainable.

Looking ahead, the tension between broad generalization and specialized adaptability in language technologies is likely to intensify. Although modern commercial \glspl{llm} offer impressive performance with minimal adaptation, our research indicates that specialized domains still pose significant challenges. Therefore, future progress will likely depend on building systems that combine general-purpose capabilities with inherent adaptability. This balanced approach promises not only better domain-specific solutions but also more responsible and efficient use of resources in a rapidly evolving technological landscape.

%----------------------–––––––––––––

\cleardoublepage
\phantomsection
\addcontentsline{toc}{chapter}{Summary}
\chapter*{Summary}
\noindent Machine Translation (\gls{mt}) and Quality Estimation (\gls{qe}) have advanced rapidly, fuelled by neural architectures and the growing influence of large language models (LLMs). Yet, despite these breakthroughs, modern systems still struggle when used outside the general-domain settings they were trained for. In specialized domains such as healthcare, law, or information technology,\footnote{In our studies, the domain is treated as the origin of a dataset; topical labels here are used only to help readers intuitively understand the concept.} the data often come from very different sources, follow different conventions, and contain terminology that generic models are not equipped to handle. When models encounter a mismatch between training data and real-world deployment—a problem widely known as \textit{domain mismatch}—translation accuracy drops, terminology becomes unreliable, and automated quality predictions lose consistency. Simply scaling up model size or training data volume does not resolve this issue; on the contrary, it introduces substantial additional computational requirements without improving domain-specific performance. What matters is whether systems are adapted to the domain they need to serve, especially with respect to the source and nature of the data used during training.

This dissertation investigates how \gls{mt} and \gls{qe} systems can be made more accurate, more adaptable, and more computationally efficient in domain-specific scenarios. Because there is no single solution to domain mismatch, the work approaches the problem from several angles: selecting the right training data, designing adaptation pipelines for \gls{qe}, understanding the role of subword tokenization and vocabulary choices during fine-tuning, and finally, using \gls{qe} to guide in-context learning (ICL) for LLM-based translation. Jointly, these investigations are organized around four core research questions:

\begin{itemize}
    \item RQ1: What is the optimal amount of in-domain data required to achieve high MT quality at low computational cost?
    \item RQ2: What are the effects of combining domain adaptation and data augmentation on the performance and generalizability of QE models across different domains and languages?
    \item RQ3: What is the optimal combination of subword tokenization model source and vocabulary source to maximize translation quality and adaptation efficiency during fine-tuning?
    \item RQ4: How effective are domain-specific QE models in determining effective in-context examples (ICEs) that improve MT quality in generative LLMs?
\end{itemize}

The first contribution of this dissertation addresses RQ1 by introducing a modular, similarity-based method for selecting high-quality in-domain sentence pairs from large generic corpora. Experiments show that small, carefully selected subsets of contextually relevant data consistently outperform far larger mixed-domain datasets. Translation quality saturates quickly, and expanding the dataset beyond a certain point introduces noise that degrades performance. These findings highlight that \textit{carefully selected data} can substitute for expensive large-scale training, enabling high-quality MT with significantly reduced computational cost compared to training on the entire available dataset or scaling model size.

The second contribution addresses RQ2 by proposing a staged \gls{qe} training pipeline that integrates domain adaptation with lightweight data augmentation. This three-step process—generic fine-tuning, mixed-domain training, and final in-domain specialization—produces \gls{qe} models that are robust across languages, domains, and resource levels. Evaluation across diverse multilingual benchmarks, including zero-shot and cross-lingual settings, confirms that \gls{qe} models, like \gls{mt} systems, require domain-specific adaptation for reliable quality prediction. Domain tags, synthetic data generation, and multilingual mixing collectively improve generalization while keeping training efficient and scalable.

The third contribution answers RQ3 by conducting the first systematic analysis of how subword tokenization and vocabulary source jointly affect fine-tuning outcomes in MT. Results show that mismatched tokenization–vocabulary configurations lead to unstable training and poorer translation quality, while coherent configurations—especially those derived from in-domain data—yield the most effective domain adaptation. Fully in-domain tokenization offers the highest accuracy but at greater computational cost; hybrid configurations provide a balance by combining strong performance with better efficiency. These findings show that tokenization is a core component of domain adaptation, not a preprocessing detail.

Answering RQ4, the fourth contribution introduces a novel \gls{qe}-guided ICL framework for generative LLMs. Domain-specific \gls{qe} models are used to score and select ICEs that maximize translation quality without requiring reference translations or parameter updates. This method outperforms random, n-gram–based, and BM25-based retrieval approaches,\footnote{BM25~\cite{10.1561/1500000019} ranks documents—in our case, sentences—based on their relevance to a search query.} showing that the quality, relevance, and domain fit of ICEs determine their effectiveness. QE-guided ICL achieves strong improvements in BLEU and COMET while requiring far less computation than full fine-tuning, establishing it as a practical alternative for domain-specific MT with LLMs.

These four contributions offer a unified, data-centric view of domain adaptation in MT and QE. The results underscore three central themes:
(1) \textit{Domain specificity is essential for reliable performance}: neither MT nor QE systems generalize well across domains without targeted adaptation, as demonstrated in Chapters~\ref{chap:CLIN} and~\ref{chap:EAMT}, and further reinforced in the tokenization experiments of Chapter~\ref{chap:BPE};
(2) \textit{Better data beats more data}: high-quality, domain-relevant examples consistently outperform large volumes of generic data, as shown in Chapter~\ref{chap:CLIN} and later examined from a vocabulary perspective in Chapter~\ref{chap:BPE}; and
(3) \textit{Adaptation must be efficient}: from targeted data selection to hybrid tokenization and QE-guided ICL, the methods proposed in this dissertation provide practical guidance for reducing computational and environmental costs while preserving—or improving—task performance, as explored across Chapters~\ref{chap:EAMT}, \ref{chap:BPE}, and \ref{chap:AMTA}.

Overall, this dissertation advances both the theory and practice of domain adaptation by investigating the full MT pipeline—from data selection and tokenization to inference for MT systems and prompting for LLMs, as well as quality evaluation and estimation—and by providing reproducible tools and detailed computational analyses that enable sustainable, domain-aware MT workflows. The findings reflect a broader shift in the field away from purely model-centric scaling and toward approaches that prioritize data relevance, domain specificity, and computational efficiency. As MT and QE systems are increasingly deployed in real-world professional settings, the methods developed here offer a solid foundation for building solutions that are more robust, more adaptable, and better aligned with the demands of specialized domains, including those faced by language service providers and other industry practitioners.

\cleardoublepage
\phantomsection
\addcontentsline{toc}{chapter}{Samenvatting (Dutch Summary)}
\chapter*{Samenvatting (Dutch Summary)}
\noindent Machinevertaling (\gls{mt}) en kwaliteitsinschatting (\gls{qe}) hebben zich de afgelopen jaren snel ontwikkeld, aangedreven door neurale netwerken en de groeiende invloed van grote taalmodellen (LLM’s). Toch blijven moderne systemen slecht presteren zodra zij worden ingezet buiten de generieke domeinen waarop zij zijn getraind. In gespecialiseerde domeinen zoals de gezondheidszorg, het recht, of informatietechnologie,\footnote{In onze studies wordt het domein beschouwd als de herkomst van een dataset; thematische labels worden hier uitsluitend gebruikt om het concept voor lezers intuïtiever te maken.} is de data vaak afkomstig uit verschillende bronnen, volgt zij andere conventies, en bevat zij terminologie waar generieke modellen niet over beschikken. Wanneer modellen worden geconfronteerd met een mismatch tussen trainingsdata en daadwerkelijke toepassingen--een probleem dat algemeen bekendstaat als \textit{domein-mismatch}--neemt de vertaalnauwkeurigheid af, wordt terminologie onbetrouwbaar en verliezen automatische kwaliteitsinschattingen hun consistentie. Het opschalen van modelgrootte of de hoeveelheid trainingsdata lost dit probleem niet op; sterker nog, het verhoogt de computationele kosten aanzienlijk zonder de prestaties binnen het gewenste domein te verbeteren. Wat werkelijk telt, is of systemen zijn aangepast aan het domein waarin zij moeten functioneren, met name wat betreft de herkomst en aard van de gebruikte data.

Dit proefschrift onderzoekt hoe \gls{mt}- en \gls{qe}-systemen nauwkeuriger, beter aanpasbaar en computationeel efficiënter kunnen worden gemaakt bij domeinspecifieke scenario’s. Omdat er geen eenduidige oplossing bestaat voor domein-mismatch, wordt dit vraagstuk benaderd vanuit meerdere invalshoeken: het selecteren van geschikte trainingsdata, het ontwerpen van adaptatiepijplijnen voor \gls{qe}, het begrijpen van de rol van subwoordtokenisatie en woordenschatkeuze tijdens fine-tuning, en het gebruik van \gls{qe} om in-context learning (ICL) voor LLM-gebaseerde vertaling te sturen. Deze onderzoeken zijn gestructureerd rond vier centrale onderzoeksvragen:

\begin{itemize}
    \item RQ1: Wat is de optimale hoeveelheid domeinspecifieke data die nodig is om hoge MT-kwaliteit te behalen tegen lage computationele kosten?
    \item RQ2: Wat zijn de effecten van het combineren van domeinadapatie en data-augmentatie op de prestaties en generaliseerbaarheid van QE-modellen voor verschillende domeinen en talen?
    \item RQ3: Wat is de optimale combinatie van de bron van het subwoordtokenisatiemodel en de bron van de woordenschat om vertaalkwaliteit en adaptatie-efficiëntie tijdens fine-tuning te maximaliseren?
    \item RQ4: Hoe effectief zijn domeinspecifieke QE-modellen in het bepalen van effectieve in-context voorbeelden (ICE’s) die de MT-kwaliteit verbeteren in generatieve LLM’s?
\end{itemize}

De eerste bijdrage van dit proefschrift beantwoordt RQ1 en introduceert een modulaire, op similariteit gebaseerde methode voor het selecteren van hoogwaardige domeinspecifieke zinnen-paren uit grote generieke corpora. Experimenten tonen aan dat kleine, zorgvuldig geselecteerde subsets van contextueel relevante data consequent beter presteren dan veel grotere datasets met gemengde domeinen. De vertaalprestatie verzadigt snel, en het uitbreiden van de dataset voorbij dit punt introduceert ruis die de kwaliteit juist vermindert. Deze bevindingen tonen aan dat \textit{zorgvuldig geselecteerde data} dure grootschalige training kunnen vervangen en hoogwaardige MT mogelijk maken tegen aanzienlijk lagere computationele kosten.

De tweede bijdrage betreft RQ2 en stelt een gefaseerde \gls{qe}-training-pijplijn voor die domeinadapatie combineert met beperkte augmentatie van data. Dit driestapsproces--generieke fine-tuning, training op gemengde domeinen, en tenslotte specifieke domeinaanpassing--levert \gls{qe}-modellen op die robuust zijn met betrekking tot talen, domeinen, en bronnen. Evaluaties op diverse meertalige benchmarks, waaronder zero-shot en cross-linguale scenario’s, bevestigen dat \gls{qe}-modellen, net als \gls{mt}-systemen, domeinspecifieke adaptatie nodig hebben voor betrouwbare kwaliteitsvoorspellingen. Domeintags, synthetische data, en meertalige mengstrategieën verbeteren de generalisatie terwijl de training efficiënt en schaalbaar blijft.

De derde bijdrage beantwoordt RQ3 door de eerste systematische analyse uit te voeren van hoe subwoordtokenisatie en woordenschatbron gezamenlijk de fine-tuning-prestaties van MT beïnvloeden. De resultaten laten zien dat slecht op elkaar afgestemde combinaties van tokenisatiemodel en woordenschat leiden tot instabiele training en lagere vertaalnauwkeurigheid, terwijl coherente configuraties-—met name die gebaseerd op domeinspecifieke data—-de meest effectieve domeinadapatie opleveren. Volledig domeinspecifieke tokenisatie levert de hoogste nauwkeurigheid op, maar tegen hogere computationele kosten; hybride configuraties bieden een balans tussen prestaties en efficiëntie. Deze bevindingen tonen aan dat tokenisatie een kernonderdeel is van domeinadapatie, en niet slechts een detail van pre-processing.

De vierde bijdrage, die RQ4 beantwoordt, introduceert een nieuw \gls{qe}-gestuurd ICL-raamwerk voor generatieve LLM’s. Domeinspecifieke \gls{qe}-modellen worden gebruikt om ICE’s te scoren en zodanig te selecteren dat vertaalprestaties maximaliseren zonder referentievertalingen of parameterupdates. Deze methode presteert beter dan willekeurige selectie, n-grams, en BM25-gebaseerde retrieval,\footnote{BM25~\cite{10.1561/1500000019} rangschikt documenten—hier: zinnen—op basis van hun relevantie voor een zoekopdracht.} en laat zien dat de kwaliteit, relevantie en domeinfit van ICE’s bepalend zijn voor hun effectiviteit. QE-gestuurde ICL levert sterke verbeteringen op in BLEU en COMET, en vereist aanzienlijk minder computationele middelen dan volledige fine-tuning, waardoor het een praktisch alternatief vormt voor domeinspecifieke MT met LLM’s.

Gezamenlijk bieden deze vier bijdragen een samenhangende, datagedreven visie op domeinadapatie in MT en QE. De resultaten onderstrepen drie centrale thema’s:
(1) \textit{Domeinspecificiteit is essentieel voor betrouwbare prestaties}: zowel MT- als QE-systemen generaliseren niet goed over domeinen zonder gerichte adaptatie, zoals aangetoond in Hoofdstukken~\ref{chap:CLIN} en~\ref{chap:EAMT}, en verder bevestigd in tokenisatie-experimenten in Hoofdstuk~\ref{chap:BPE};
(2) \textit{Betere data verslaat meer data}: hoogwaardige, domeinrelevante voorbeelden presteren consequent beter dan grote hoeveelheden generieke data, zoals aangetoond in Hoofdstuk~\ref{chap:CLIN} en opnieuw onderzocht vanuit een woordenschat-perspectief in Hoofdstuk~\ref{chap:BPE}; en
(3) \textit{Adaptatie moet efficiënt zijn}: van gerichte dataselectie tot hybride tokenisatie en QE-gestuurde ICL—-de in dit proefschrift voorgestelde methoden bieden praktische handvatten om computationele en ecologische kosten te verlagen terwijl de taakprestatie behouden blijft of verbetert, zoals onderzocht in Hoofdstukken~\ref{chap:EAMT}, \ref{chap:BPE} en \ref{chap:AMTA}.

Samenvattend draagt dit proefschrift bij aan zowel theorie als praktijk van domeinadapatie door de volledige MT-keten te onderzoeken-—van dataselectie en tokenisatie tot inferentie voor MT-systemen en prompting voor LLM’s, evenals kwaliteitsbeoordeling en -inschatting. Bovendien levert het reproduceerbare tools en gedetailleerde computationele analyses die duurzame, domeinbewuste MT-workflows mogelijk maken. De bevindingen weerspiegelen een bredere verschuiving in het vakgebied, weg van puur modelcentrische schaalvergroting, richting benaderingen die datakwaliteit, domeinspecificiteit en computationele efficiëntie centraal stellen. Nu MT- en QE-systemen steeds vaker worden ingezet in professionele praktijken, bieden de ontwikkelde methoden een solide basis voor het bouwen van systemen die robuuster, beter aanpasbaar, en beter afgestemd zijn op de eisen van gespecialiseerde domeinen, waaronder die van taaldienstverleners en andere industriële gebruikers.

\cleardoublepage
\phantomsection
\addcontentsline{toc}{chapter}{Acknowledgments}
\chapter*{Acknowledgments}
A PhD, in many ways, can be viewed as an extreme form of domain adaptation. You start as a generally trained person with a few default settings, and then go through years of fine-tuning—hoping that by the end, you become something stable and useful, without too many exploding gradients along the way. Whether that metaphor truly matches reality is debatable. In any case, this dissertation is the output of that long training process. It would not have been possible without the many people who guided me, supported me, and sometimes rescued me. I group them here using the domain adaptation metaphor—not because people fit neatly into categories, but because it adds a bit of structure (and humor) to the gratitude I owe them.
\vspace{-0.5em}
\paragraph{General-domain pretraining folks}
These are the people who helped build the base model I started with and gently pulled me away from becoming too task-specific—those who shaped my values, kept me grounded, and made it possible for me to grow.

I begin with \textit{Mahsa}. Your support has been steady and essential. Moving to a new country, far from our families, was not easy, but your courage, trust, and patience made even the hard moments manageable. Your belief in me gave me the strength to do a PhD with a calmer mind and a hopeful heart. Nothing in this work would have been possible without your constant support. Love you.

My brother \textit{Mohammad} has been a strong source of support. Even from far away, your humor and belief in me always reached me at the right time. The PhD may have pushed me down a few levels in FIFA, but I promise to restore that soon—along with our banter.

My thanks also go to my \textit{mom and dad}, and to my \textit{parents-in-law}. Your support reached us in many ways—big and small—and made life abroad feel much lighter. Thank you for cheering us on from afar.

\paragraph{Fine-tuning folks}
These are the people who guided the task-specific part of my training—my PhD supervisors.

To \textit{Dimitar}: you taught me not only how to do research, but also how to grow in academia. From teaching to supervision, thank you for giving me the opportunity to learn and improve. Your support went far beyond papers and experiments; you helped me through both the good moments and the challenging ones. You were always there, from beginning to end, willing to help. Thank you for the role you played in my growth.

And to \textit{Pieter}: I could not have asked for a better promoter. You treated me as a colleague—not just a student—and I appreciated how you valued my ideas and gave me the space to express what mattered to me. It meant more than I can say and shaped my PhD experience in the best possible way. And of course, our board game gatherings were perfect breaks at exactly the right moments. Thank you.

I also want to highlight an important point: both of you were always present, available, and engaged throughout my PhD. This may sound normal, but I know it is not always the case. Your presence and support made the journey much smoother, and I am thankful for that.

\paragraph{Parallel corpus folks}
These are the people who kept me grounded, positive, and supported throughout the ups and downs of PhD life. I met many wonderful individuals along the way, and I am grateful to each of them. This list could easily be longer, and if you expected to find your name here but do not—you still made a meaningful impact, and I appreciate you.

I begin with \textit{Chris Emmery}, whose positive energy and generosity in sharing his experiences meant a great deal to me—thank you, \textit{Chris}. I also thank \textit{Stijn Rotman} (a.k.a.\ ``True Dutch'') for our post-submission beer sessions, and \textit{Samaneh Khoshrou} for her positivity and our food discussions that always made me hungry.

I am grateful to \textit{Marijn van Wingerden} for his guidance and for giving me room to grow; to \textit{Marie Postma} for her openness and support when I stepped in as a lecturer; and to \textit{Drew Hendrickson} for his approachability—especially the random corridor chats that helped me reset between breaks.

My thanks also go to colleagues from our machine translation group who shaped my journey in their own ways: \textit{Mirella De Sisto}, \textit{Eva~Vanmassenhove}, and \textit{Fred Blain}.

And to two of the best—\textit{Karin Berkhout} and \textit{Eva Verschoor-Suitela}. Thanks for your patience in indulging my attempts to learn Dutch, and for helping me navigate the many requests that came my way.

And finally, thanks to my great friends who have helped in various ways to make my moments more fun: \textit{Marcel van Groenendaal}, \textit{Reza Bakhtiari}, \textit{Mahyar Ghazanfari}, \textit{Alireza Farzad}, and \textit{Kambiz Darvishnezhad}. You guys really did me a solid.

\backmatter

% --- Appendices (AFTER references) ---
\cleardoublepage
\makeatletter
\@appendixtrue
\makeatother
\begin{appendices}
\chapterseparatorpage
\chapterimage{Assets/separator.pdf}

\chapter*{Appendix A: Efficiency and Sustainability Overview}
\addcontentsline{toc}{chapter}{Appendix A: Efficiency and Sustainability Overview}
\renewcommand{\thechapter}{A}
\setcounter{chapter}{1}
\markboth{Appendix A: Efficiency and Sustainability Overview}{}

\label{Appendix:ComCosts}
This appendix provides an overview of the computational costs and efficiency considerations associated with the methodologies and experiments in this dissertation. As the scale and complexity of \gls{mt} and \gls{qe} systems continue to grow—particularly in settings that involve training—understanding the trade-offs between performance, resource consumption, and sustainability becomes increasingly important. The following sections summarize key findings from each chapter, focusing on metrics such as training time, inference speed, GPU utilization, energy consumption, and CO\textsubscript{2} emissions. These analyses demonstrate how the methodological choices adopted in this dissertation—ranging from data selection and \gls{da} to \gls{sw} tokenization and \gls{icl}—shape the efficiency and environmental footprint of \gls{mt} and \gls{qe} workflows.

\section{Chapter \ref{chap:CLIN}}
Training large-scale \gls{mt} models is computationally expensive, particularly when relying on generic-domain corpora containing millions of sentences. The data selection method proposed in Chapter~\ref{chap:CLIN} substantially reduces computational costs by retaining only the most relevant in-domain sentences from a large corpus, thereby improving training efficiency and reducing resource consumption. Compared to training on the full 31-million-sentence generic corpus, this approach reduces training time by over 90\%—from 2 days and 5 hours on three NVIDIA Tesla V100 GPUs to just 8 hours for the best-performing model—reduces the data footprint by 97\% (i.e., the volume of data processed during training), and lowers GPU energy consumption and associated carbon emissions by over 85\%, estimated proportionally from the reduction in training time on identical hardware.

% \begin{itemize}
%     \item Reduces training time by over 90\%––from 2 days and 5 hours on 3 NVIDIA Tesla V100 GPUs to just 8 hours for the best-performing model.
%     \item Cuts the data footprint by 97\%, requiring only 1 million sentences instead of 31 million, without sacrificing translation quality.
%     \item Reduces GPU energy consumption and carbon emissions by over 85\%, as fewer training iterations are needed to achieve optimal performance.
% \end{itemize}  

\begin{table}[h]
\centering
\renewcommand{\arraystretch}{1.3} % increases row height
\setlength{\tabcolsep}{8pt}       % increases column padding

\begin{tabular}{lcc}
\hline \hline
\textbf{Aspect} & \textbf{Baseline} & \textbf{Selected Data} \\
\hline \hline
Training time & 2d 5h (3×V100) & 8h \\
Data required & 31M sentences & 1M sentences \\
Energy / CO\textsubscript{2} & 100\% & $\sim$15\% \\
\hline \hline
\end{tabular}

\caption{Resource usage and efficiency gains for the data selection approach introduced in Chapter~\ref{chap:CLIN}.}
\end{table}

These results show that our method enables high-quality, domain-specific \gls{mt} while substantially reducing computational overhead, making it a more sustainable and cost-effective alternative to large-scale, general-domain training. That said, a fair comparison should also consider the computational cost of the data selection process itself. In Chapter~\ref{chap:CLIN}, we did not record the runtime of this step, and its cost can vary depending on factors such as the size of the out-of-domain dataset, the amount of in-domain data available, and how many in-domain samples are selected per query.

\section{Chapter \ref{chap:EAMT}}
The proposed \gls{da} pipeline introduced for \gls{qe} in Chapter~\ref{chap:EAMT} adds an extra training step, which increases total training time by 77.8\% (from 3.6h to 6.4h). While this raises computational cost, it yields substantial gains in generalization and cross-lingual robustness compared to the baseline fine-tuning approach, as summarized in Table~\ref{tab:eamt-results}; further details are provided in Chapter~\ref{chap:EAMT}.

\begin{table}[h]
\centering
\renewcommand{\arraystretch}{1.15}
\setlength{\tabcolsep}{6pt}

\begin{tabular}{l c}
\hline\hline
\textbf{Metric} & \textbf{Outcome (vs.\ baseline)} \\
\hline\hline
Training time & 3.6h → 6.4h \, (\,+77.8\%\,) \\
Avg.\ Pearson correlation & ↑ 14.36\% \\
Zero-shot EN-ZH & ↑ 25.51\% \\
Cross-lingual inference & ↑ 11.60 Pearson pts \\
OOD fine-tuning (per language) & ↓ 45\% time \\
Step~2 vs.\ training from scratch & ↓ 50\% time \\
\hline\hline
\end{tabular}

\caption{Efficiency and performance outcomes of the \gls{da} pipeline presented in Chapter~\ref{chap:EAMT}. 
Improvements (↑) indicate performance gains, while reductions (↓) indicate lower computational cost. 
``Avg.\ Pearson correlation'' reflects overall QE accuracy; ``Zero-shot EN-ZH'' measures model performance without EN-ZH training examples; 
``Cross-lingual inference'' captures transfer performance across language pairs; 
``OOD fine-tuning'' refers to initializing a new language pair by fine-tuning a pre-trained out-of-domain model; 
and ``Step~2 vs.\ training from scratch'' compares the cost of the second adaptation stage with full model retraining.}

\label{tab:eamt-results}
\end{table}

\section{Chapter \ref{chap:BPE}}

Chapter~\ref{chap:BPE} evaluated how different BPE and vocabulary configurations affect the efficiency of fine-tuning for in-domain translation. The results show clear differences in computational cost: hybrid configurations that reuse the original tokenizer and vocabulary (C4, C5, C7) are the most efficient, completing training in roughly three hours and producing substantially lower CO\textsubscript{2} emissions. Among these, C5 offers the best balance between runtime and translation quality.

In contrast, the configuration achieving the highest in-domain BLEU score (C6) is also the most resource-intensive, requiring nearly three times longer to train and emitting more than double the CO\textsubscript{2} of the efficient hybrid methods. Even the standard baseline (C1) remains costly. These efficiency patterns are summarized in Table~\ref{tab:bpe-efficiency}.

\begin{table}[h]
\centering
\renewcommand{\arraystretch}{1.2}
\setlength{\tabcolsep}{10pt}

\begin{tabular}{lccc}
\hline\hline
\textbf{Configuration} & \textbf{BLEU} & \textbf{Time (h)} & \textbf{CO\textsubscript{2} (g)} \\
\hline\hline
C6 (Best in-domain)   & 54.8\,↑ & 9.5\,↑ & 1587\,↑ \xmark \\
C1 (Baseline)         & 53.6    & 7.75\,↑ & 1658\,↑ \xmark \\
C5 (Fastest hybrid)   & 53.1    & 3.25\,↓ & 729\,↓ \cmark \\
C7 (Minimal compute)  & 52.9\,↓ & 3.0\,↓ & 543\,↓ \cmark \\
C4 (Minimal compute)  & 53.0    & 3.1\,↓ & 723\,↓ \cmark \\
\hline\hline
\end{tabular}

\caption{Efficiency comparison of BPE/vocabulary configurations (Chapter~\ref{chap:BPE}). 
Arrows denote relative differences: ↑ = higher value (worse for time/CO\textsubscript{2}, better for BLEU), ↓ = lower value (more efficient). 
\cmark\ marks efficient configurations; \xmark\ marks costly ones.}
\label{tab:bpe-efficiency}
\end{table}

Overall, the results show that small gains in BLEU can come with disproportionately large increases in computation and emissions. Hybrid approaches offer a more sustainable alternative, cutting training time and CO\textsubscript{2} by more than half while maintaining competitive translation quality.

\section{Chapter \ref{chap:AMTA}}

Chapter~\ref{chap:AMTA} evaluated the computational efficiency of our \gls{icl} approach relative to retrieval-based methods, task-level prompting, and full model fine-tuning. The focus here is on how much time, energy, and CO\textsubscript{2} can be saved while still achieving strong translation performance. Table~\ref{tab:amta-eff} provides a compact overview of the main efficiency indicators.

\begin{table}[h]
\centering
\renewcommand{\arraystretch}{1.05}
\setlength{\tabcolsep}{6pt}

\begin{tabular}{lccc}
\hline\hline
Method & Time (h) & CO\textsubscript{2} (kg) & Energy (kWh) \\
\hline\hline
BM25 (q=16)               & 0.97 & 0.07 & 0.19 \\
R-BM25 (q=16)             & 1.07 & 0.07 & 0.21 \\
ICL (patience 3)          & 1.82 & 0.22 & 0.67 \\
ICL (patience 8)          & 3.80 & 0.50 & 1.51 \\
ICL (patience 16)         & 5.11 & 0.68 & 2.05 \\
Fine-tuned mBART--50      & 11.20 & 1.88 & 4.82 \\
Task-level prompting (16) & 78.5 & 12.80 & 35.91 \\
\hline\hline
\end{tabular}

\caption{Runtime, CO\textsubscript{2} emissions, and energy usage for methods evaluated in Chapter~\ref{chap:AMTA}. ICL introduces additional search cost compared to retrieval-based baselines, but remains substantially more efficient than full fine-tuning and task-level prompting.}
\label{tab:amta-eff}
\end{table}

Our best-performing ICL configuration (patience 16) achieves a 3.5 BLEU point improvement over R-BM25 while requiring 5.11 hours of computation. Although this introduces a runtime increase relative to R-BM25, it remains far more efficient than fine-tuning mBART--50. \gls{icl} reduces time-to-prediction by approximately 54 percent compared to fine-tuning and emits only 36 percent of its CO\textsubscript{2}, corresponding to a 64 percent reduction in carbon footprint.

Energy consumption follows the same trend: \gls{icl} uses about 42.5 percent less GPU power than fine-tuning. Early stopping also provides significant gains. Using patience 3 instead of 16 reduces runtime by around 71 percent while retaining competitive BLEU performance, making it an attractive option when rapid inference is needed.

Task-level prompting is the most computationally expensive strategy, requiring 78 hours for 16 examples—over fifteen times the cost of our most expensive \gls{icl} setting. In comparison, \gls{icl} offers a practical middle ground: it substantially improves translation accuracy while keeping runtime, energy, and emissions far below those of full fine-tuning or task-level prompting.

\vspace{3pt}
\noindent Across all four chapters, the results show that high-quality MT and QE inevitably involve trade-offs between performance and computational cost, but thoughtful design choices can make these processes far more sustainable. Data selection (Chapter~\ref{chap:CLIN}) demonstrates that small, domain-focused subsets can replace large generic corpora while sharply reducing training time and emissions. The QE adaptation pipeline (Chapter~\ref{chap:EAMT}) shows that modest extra training yields strong gains in generalization. Tokenization and vocabulary strategies (Chapter~\ref{chap:BPE}) indicate that hybrid or partially reused vocabularies retain most benefits of full in-domain adaptation at much lower energy cost. Finally, the \gls{icl} approach (Chapter~\ref{chap:AMTA}) offers a lightweight alternative to fine-tuning, reducing inference time, energy usage, and CO\textsubscript{2} while maintaining strong translation performance. Overall, these findings highlight the value of efficient, sustainable design choices and the importance of being transparent about their computational implications.

\end{appendices}

\clearpage
\includepdf[
  pages=1,
  pagecommand={\thispagestyle{empty}},
  fitpaper=true
]{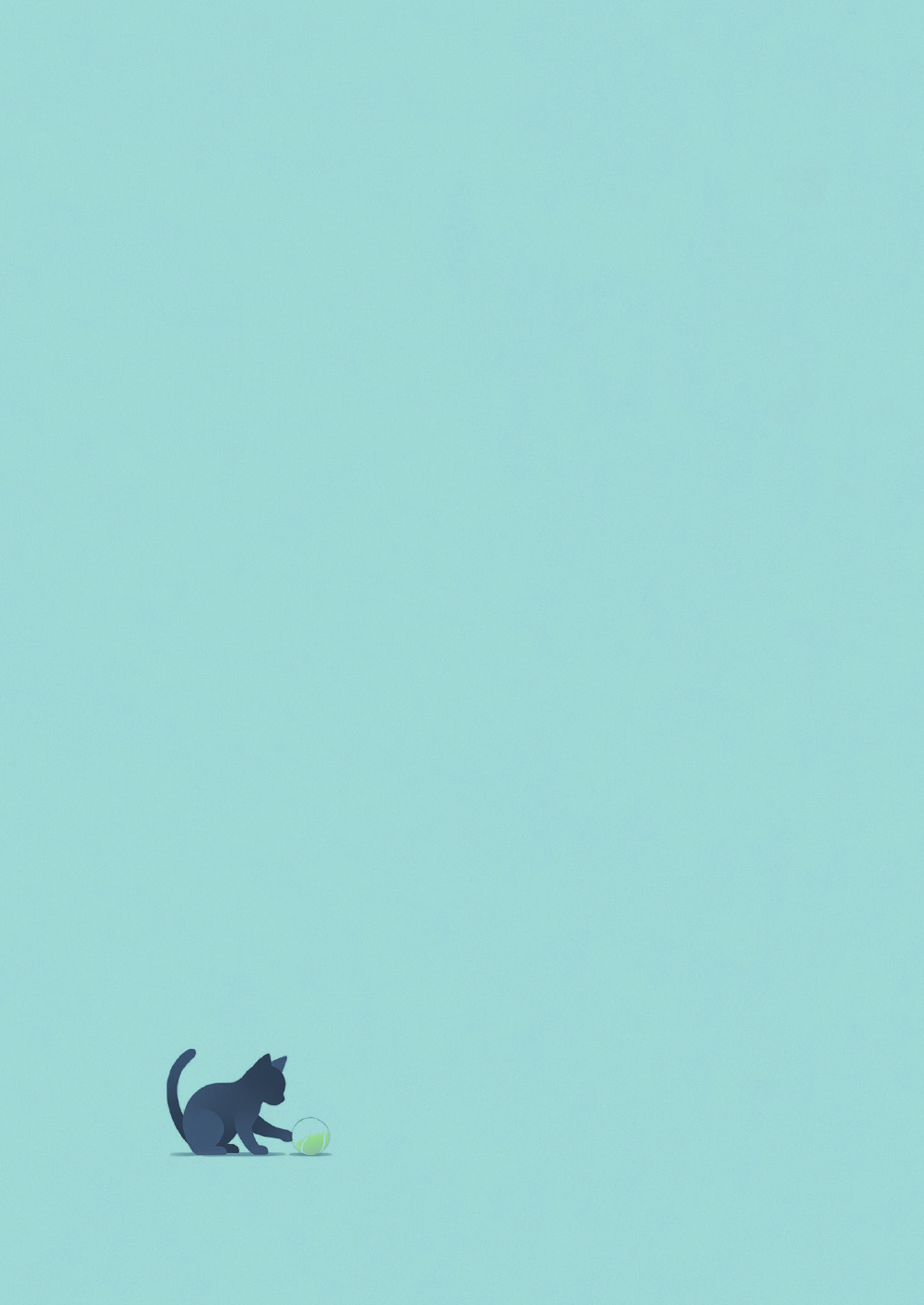}
\clearpage

\end{document}